\documentclass[lettersize,journal]{IEEEtran}
\usepackage{amsmath,amsfonts,amssymb}
\usepackage{array}
\usepackage[caption=false,font=normalsize,labelfont=sf,textfont=sf]{subfig}
\usepackage{textcomp}
\usepackage{stfloats}
\usepackage{url}
\usepackage{verbatim}
\usepackage{graphicx}
\usepackage{cite}
\usepackage{booktabs}
\usepackage{xcolor}
\usepackage{hyperref}
\usepackage{multirow}
\usepackage[capitalize]{cleveref}
\usepackage{tikz}
\usetikzlibrary{arrows.meta, positioning, calc}

\begin{document}

\title{The Gaussian Is Enough: Flow-Matching Priors Do Not Help When Fine-Tuning Large Behavior Models}

\author{%
  Chen~Xu\textsuperscript{1},
  Rishi~Shah\textsuperscript{2},
  Hadas~Kress-Gazit\textsuperscript{3},
  Haruki~Nishimura\textsuperscript{1},
  and~Masha~Itkina\textsuperscript{1}%
  \\[2pt]
  \normalfont\small
  \textsuperscript{1}Toyota Research Institute \quad
  \textsuperscript{2}Woven by Toyota \quad
  \textsuperscript{3}Cornell University%
  \thanks{Correspondence to: \texttt{chen.xu@tri.global}}%
}

\markboth{}{Xu \MakeLowercase{\textit{et al.}}: The Gaussian Is Enough: Flow-Matching Priors in LBM Fine-Tuning}

\maketitle

\begin{abstract}
Modern robot imitation learning increasingly relies on generative policies based on diffusion or flow-matching models, which generate actions by transforming samples from a prior distribution. A key question is whether the choice of prior matters. Replacing the standard Gaussian with a closer-to-target, non-Gaussian prior has been shown to substantially improve performance when \emph{training from scratch}. A natural next step is to ask whether these gains transfer to \emph{fine-tuning} pretrained Large Behavior Models (LBMs) such as LBM 1.0, $\pi_{0.5}$, and GR00T~N1.5, where one might expect even larger gains. Surprisingly, we find that this is not the case, except possibly at very low fine-tuning data fractions. Across over 100K simulation rollouts spanning all three aforementioned LBMs on 40+ tasks in two simulation platforms, and 1250 hardware rollouts on five bimanual manipulation tasks, non-Gaussian priors that are demonstrably closer to the target yield statistically indistinguishable or worse fine-tuning performance than a standard Gaussian prior. Diagnostic analyses suggest why: fine-tuned imitation learning policies converge to similar action predictions across priors, despite their fine-tuned \emph{encoder embeddings} diverging substantially from the pretrained embeddings and each other. A learning-rate ablation further confirms that encoder training is the dominant factor in fine-tuning performance, substantially outweighing the effect of prior choice. We conclude with concrete directions for future research on when and why learned priors might still matter in fine-tuning. Project page: \url{https://cxu-tri.github.io/non_gaussian_FT/}.
\end{abstract}

\begin{IEEEkeywords}
Imitation learning, flow-matching \& diffusion policies, action priors, large behavior models, fine-tuning.
\end{IEEEkeywords}

\section{Introduction}
\label{sec:intro}

\IEEEPARstart{I}{mitation} learning for robot manipulation increasingly employs generative models, such as diffusion and flow-matching policies~\cite{chi2023diffusion, reuss2023beso}, which transport samples from a prior distribution~$p_0$ to a target action distribution conditioned on observations. Large Behavior Models (LBMs) scale this approach: pretrain once on diverse vision, language, and robot action data, then fine-tune on a modest set of task-specific demonstrations. Recent examples include LBM 1.0~\cite{tri2026lbm}, $\pi_{0.5}$~\cite{black2025pi05}, and GR00T~N1.5~\cite{gr00tn1_2025}. By default, these LBMs use an isotropic Gaussian $\mathcal{Z}\sim\mathcal{N}(0,I)$ as the prior, though flow matching~\cite{lipman2023flow} is agnostic to $p_0$ by construction. In other words, the Gaussian is a convention, not a requirement.
A line of recent work~\cite{jia2026a2a, chen2024bridger, dong2026conditioning, chang2026efficientflow} has exploited this flexibility when \emph{training from scratch}: learning single-task policies without loading weights pretrained on large-scale data. Replacing the Gaussian with a closer-to-target prior substantially improves success rates on common benchmarks such as LIBERO~\cite{liu2023libero}, MetaWorld~\cite{yu2020metaworld}, and Adroit~\cite{rajeswaran2018adroit}. It remains unknown whether this benefit carries over to \emph{fine-tuning} pretrained LBMs, 
where policies are initialized from pretrained weights rather than learned from scratch.

\begin{figure}[!t]
    \centering
    \begin{tabular}{@{}c@{\hspace{2pt}}c@{}}
    \includegraphics[width=0.49\linewidth]{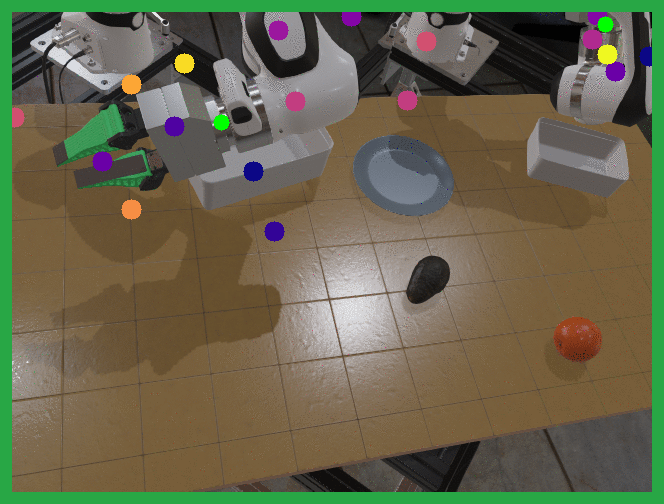} &
    \includegraphics[width=0.49\linewidth]{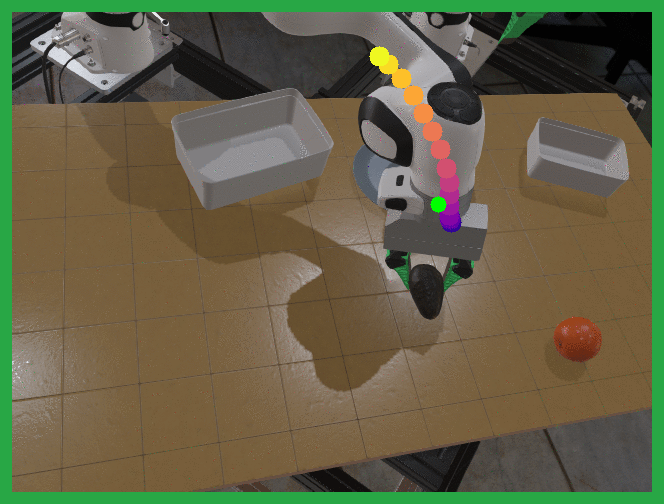} \\[1pt]
    \small $a_0\sim\mathcal{Z}$ (Gaussian) &
    \small $a_0\sim A_{\text{pre}}$ (pretrained prior)
    \end{tabular}
    \includegraphics[width=0.95\linewidth]{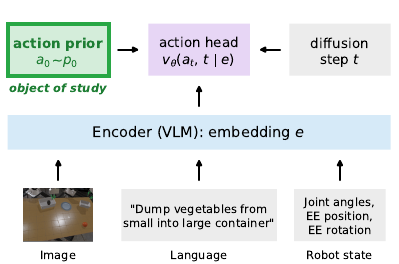}\\
    \caption{\textbf{Prior definition and visualization.} \textbf{Bottom:} schematic of an example flow-matching policy. The action head $v_\theta$ is conditioned on the encoder embedding $e$ and takes in diffusion step $t$ as well as the action prior $p_0$. Only the prior varies across our experiments. \textbf{Top:} samples drawn \emph{directly from two examples of the prior $p_0$} 
    on the simulated DumpVegetablesIntoContainer task from LBM Eval~\cite{tri2026lbm}, with overlaid colored dots showing both arms' predicted end-effector positions over the action chunk horizon. Color gradient indicates temporal ordering within the chunk, with lighter dots representing predictions further into the future. The Gaussian $\mathcal{Z}$ (left) produces incoherent, random motion, while $A_{\text{pre}}$ (right) already makes partial task progress by picking up the avocado and moving it to the bin, reflecting a prior close to the target action. We hypothesize that the close-to-target prior $A_{\text{pre}}$ would outperform $\mathcal{Z}$ after policy fine-tuning, but we empirically find that this is not the case.
    }
    \label{fig:prior_illustration}
\end{figure}

A natural question arises: \emph{Can we fine-tune more capable policies by starting from an informed action prior rather than a standard Gaussian?} Swapping the Gaussian for another distribution, such as actions sampled from the pretrained LBM itself, requires no change to the training pipeline.
The intuition is appealing: if the prior is already close to the target, the policy only needs to traverse a short transport distance in action space, and doing so should facilitate fine-tuning and improve policy performance (\cref{fig:prior_illustration}).

\begin{figure*}[t]
    \centering
    \includegraphics[width=\textwidth]{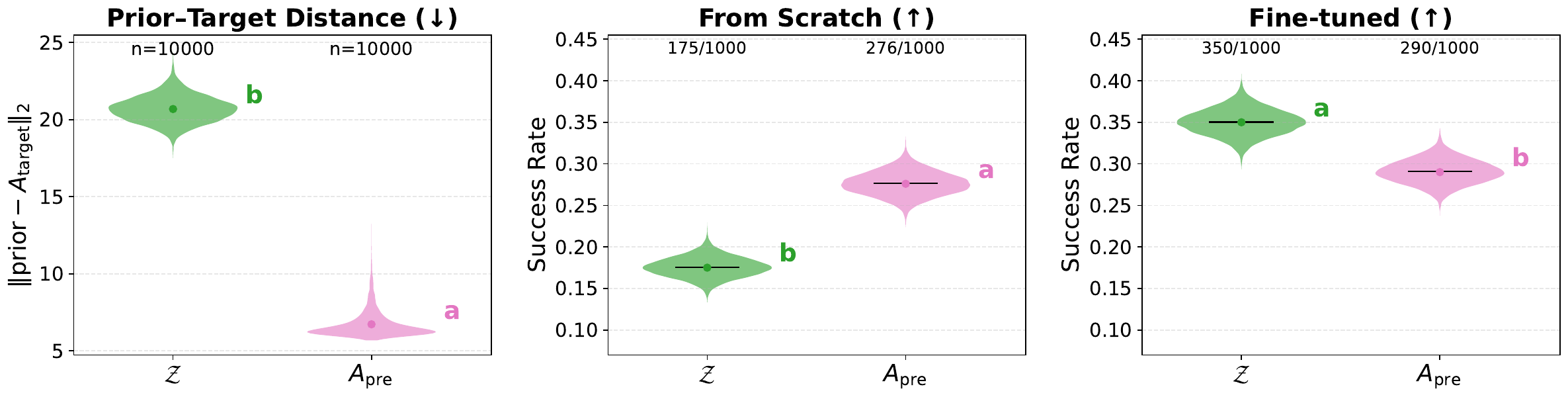}
    \caption{Action priors help when training from scratch but not when fine-tuning (Setting A, 5 LBM Eval kitchen tasks~\cite{tri2026lbm}, 1000 rollouts per prior; see \cref{sec:setting_a} for details). All fine-tuning prior variants start from the same pretrained LBM 1.0 checkpoint. $A_{\text{pre}}$ denotes the action prior sampled from the pretrained policy.
    \textbf{Left:} $\ell_2$ distance between prior samples and ground-truth actions, in un-normalized action coordinates; $A_{\text{pre}}$ is $3\times$ closer to the target data than the Gaussian.
    \textbf{Middle:} from-scratch training; $A_{\text{pre}}$ significantly outperforms the Gaussian.
    \textbf{Right:} after fine-tuning, the Gaussian prior matches or, in this case, exceeds $A_{\text{pre}}$ in success rate.
     The full comparison across all seven prior variants (\cref{fig:setting_a_full}) confirms that most priors we consider yield statistically indistinguishable performance after fine-tuning, and, thus, provide no benefit to the policy.}
    \label{fig:opener}
\end{figure*}

We therefore evaluate whether the from-scratch-training benefits of non-Gaussian priors persist after LBM pretraining.
If so, learned priors would compound the gains from pretraining with no change to the deployment pipeline.
Surprisingly, we consistently obtain negative results to this question based on average success rates and average task progress across tasks (\cref{fig:opener}). The priors that help when training from scratch almost always do not improve fine-tuning performance, and in several regimes they do worse than the default Gaussian prior.

To our knowledge, this paper is the first large-scale study of non-Gaussian action priors for LBM fine-tuning. We make three contributions:
\begin{itemize}
    \item \emph{A negative finding at scale:} Non-Gaussian priors that are demonstrably closer to the target yield fine-tuning performance that is statistically indistinguishable from, or worse than, a standard Gaussian. This holds both on average and for the majority of individual tasks.
    \item \emph{Consistency across the fine-tuning design space:} The finding is consistent across seven prior constructions, most data fractions (25\%--100\% of fine-tuning data), three LBM architectures (LBM 1.0, $\pi_{0.5}$, and GR00T~N1.5), as well as both simulation and robot hardware settings. One exception is the very-low-data regime (5\% fraction), which we discuss in \cref{sec:discussion}.
    \item \emph{An introspective explanation:} 
    Across priors, three diagnostics reveal that fine-tuned encoder embeddings diverge from pretrained embeddings and each other, yet yield indistinguishable conditional action likelihoods, and the resulting policy head outputs comparable action predictions.
    A learning-rate ablation confirms that encoder training matters far more than prior choice for policy performance, pointing to the pretrained encoder, not the prior, as the dominant factor. In short, action priors shape fine-tuning training dynamics, particularly in the observation embedding space, but do not affect policy~performance.
\end{itemize}

\section{Related Work}
\label{sec:related}

\subsection{Flow-Matching and Diffusion Policies for Robotics}

Diffusion and flow-matching policies have become prominent approaches for visuomotor imitation learning. At their core, both families define time-dependent transport between a prior and target distribution in action space. Single-task diffusion policy~\cite{chi2023diffusion} introduced the formulation, which was then extended in BESO~\cite{reuss2023beso} to goal-conditioned settings, and scaled in LBMs~\cite{black2024pi0, black2025pi05, tri2026lbm, gr00tn1_2025}. Compared to DDPM-based diffusion models~\cite{ho2020denoising}, flow-matching variants~\cite{lipman2023flow} take a more direct path from the prior to the target distribution and make no distributional assumption on the prior. These desirable properties are central to recent flow policies~\cite{gr00tn1_2025,black2025pi05} and to this paper.
Within the generative policy paradigm, LBMs (e.g., LBM 1.0~\cite{tri2026lbm}, $\pi_{0.5}$~\cite{black2025pi05}, and GR00T~N1.5~\cite{gr00tn1_2025}) pretrain on large-scale, diverse manipulation data before fine-tuning on downstream tasks, but by default use the Gaussian prior distribution in both pretraining and fine-tuning.
Parallel empirical studies of fine-tuning design choices (action-space parameterization~\cite{feng2026actionspace}, data scaling~\cite{lin2025data}, co-training~\cite{lin2026cotraining}) have examined many axes of the recipe but hold the prior fixed. 
Our paper investigates the effect of action prior choice on policy fine-tuning performance.
\vspace{-0.5cm}
\subsection{Pretraining, Transfer Learning, and Introspection}

Fine-tuning adapts a pretrained model's learned representations to new tasks, a form of transfer learning that has proven highly effective.
Prior works have shown that pretrained features improve performance in both computer vision~\cite{yosinski2014transferable, he2019rethinking} and robotics: frozen pretrained encoders~\cite{Radosavovic2022,nair2022r3m}, shared trunks~\cite{wang2024hpt}, and full pretrained policies~\cite{tri2026lbm,gr00tn1_2025,black2025pi05} all outperform training from scratch. Other works examine \emph{what explains} whether a pretrained model transfers well to a new task. Training-free probes such as LogME~\cite{you2021logme} estimate the predictive power of pretrained features. Scaling-law studies~\cite{kaplan2020scaling, schaeffer2023emergence} and likelihood-based assessments~\cite{theis2016note} examine when pretraining metrics can be trusted as proxies for downstream policy performance. Our work applies these introspective tools to the choice of prior: we compare priors using downstream performance (\cref{sec:results}) and training-free metrics, such as LogME, per-observation $\ell_2$ distance, and cosine similarity in action and embedding spaces (\cref{sec:analysis}).

\subsection{Non-Gaussian Prior Distributions}

A growing body of work has proposed structured non-Gaussian priors for policy learning.
BRIDGER~\cite{chen2024bridger} trains a conditional variational autoencoder~(CVAE) to produce informative action priors. Cocos~\cite{dong2026conditioning} uses a condition-dependent Gaussian whose parameters are learned from observations to prevent loss collapse. Retrieval~\cite{pari2022surprising, he2026demystifying, kai2026retrieval} constructs the action prior from nearest-neighbor actions, with distances measured in the observation embedding space. Action-to-action flow matching~\cite{jia2026a2a} maps demonstration actions directly to target actions for single-step inference.
Other variants include equivariance guarantees for non-Gaussian priors~\cite{chang2026efficientflow}, cluster-based and residual-action priors~\cite{chiang2026cotfm, su2026rfs}, and Schr\"odinger-bridge formulations~\cite{liu2025bridgepolicy, nguyen2025xflowmp}.
In each case, however, the proposed prior is evaluated by training a policy from scratch; none of these works investigates whether the benefit transfers to fine-tuning a pretrained~LBM.

A complementary line of research modifies the prior without changing the policy's weights.
Frozen-policy steering methods operate on the latent noise space of a trained policy: DSRL~\cite{wagenmaker2025dsrl} defines a reward function over the noise space to denoise high-reward noise samples, Golden Ticket~\cite{patil2026goldenticket} searches for optimal constant noise vectors, and GoldenStart~\cite{zhang2026goldenstart} trains a Q-guided CVAE noise prior. Related inference-time warm-starting methods~\cite{li2026step, jiang2025streaming, li2025adpro, odonchimed2025retrieve} adjust the prior at rollout time. These methods change sampling, not training, and apply to any pretrained policy. Since we focus on the effects of training-time priors in fine-tuned policies, these approaches are outside the scope of this work.

\section{Problem Setup}
\label{sec:problem}

\textbf{Setting.} We fine-tune a pretrained LBM $\pi_{\text{pre}}$ on task-specific demonstration datasets and consider whether replacing the default Gaussian prior with a non-Gaussian prior improves the fine-tuned policy performance. The pretrained policy $\pi_{\text{pre}}$ can be any flow-matching- or diffusion-based LBM (in our experiments: LBM 1.0~\cite{tri2026lbm}, $\pi_{0.5}$~\cite{black2025pi05}, or GR00T~N1.5~\cite{gr00tn1_2025}).

\textbf{Notation.} We follow the diffusion policy notation~\cite{chi2023diffusion}. The target $a_1$ is an action chunk ($H$ steps of future actions) drawn from the demonstration dataset $\mathcal{D}$. The raw observation~$o$ (camera images, language instructions, proprioception) is passed through the LBM's vision-language-action encoder to produce the observation embedding $e$, which conditions the action-head network $v_\theta$. The action head is trained to transport prior samples $a_0\sim p_0$ to targets $a_1$. We write $a_0 \sim p_0$ for a generic prior sample; the specific prior constructions (e.g., $\mathcal{Z}$, $A_{\text{pre}}$, etc.) defined in \cref{sec:prior_variants} each instantiate $p_0$. We use ``prior'' throughout the paper to refer to $p_0$.

\textbf{Why flow matching is the right testbed.} Flow matching~(FM)~\cite{lipman2023flow} regresses a velocity field directly between prior and target samples. The conditional FM objective
\begin{equation}
\begin{aligned}
    \mathcal{L}_{\text{FM}} &= \mathbb{E}_{t, a_0, a_1} \left\| v_\theta(a_t, t \mid e) - (a_1 - a_0) \right\|^2, \\
    a_t &= (1-t)\, a_0 + t\, a_1,
\end{aligned}
\end{equation}
only requires paired samples $(a_0, a_1)$ from the prior and target, so swapping $p_0$ from $\mathcal{N}(0, I)$ to any other distribution requires no change to the training or inference procedure. DDPM-style diffusion models~\cite{ho2020denoising}, by contrast, are designed around a forward noising process whose stationary distribution is Gaussian. Replacing this distribution would require redesigning the noise schedule, objective, and sampling procedure, which are non-trivial changes.

\section{Prior Variants}
\label{sec:prior_variants}
\begin{table}[t]
\centering
\caption{The seven prior variants compared in this work. $\pi_{\text{pre}}$ is the pretrained LBM (LBM 1.0, $\pi_{0.5}$, or GR00T~N1.5). \cref{sec:appendix_priors} gives full sampling procedures and auxiliary-training details.}
\label{tab:priors}
\renewcommand{\arraystretch}{1.2}
\begin{tabular}{@{}lp{6.7cm}@{}}
\toprule
\textbf{Prior} & \textbf{Description} \\
\midrule
$\mathcal{Z}$ & Isotropic Gaussian $\mathcal{N}(0, I)$ \\
\midrule
\multicolumn{2}{@{}l}{\textit{Proposed priors (constructed from $\pi_{\text{pre}}$)}} \\[2pt]
$A_{\text{pre}}$ & Actions generated by $\pi_{\text{pre}}$ \\
$A_{\text{pre}} {+} \sigma_A\mathcal{Z}$ & Actions from $\pi_{\text{pre}}$ with added Gaussian noise (scale $\sigma_A$) \\
$E_{\text{pre}} {+} \sigma_E\mathcal{Z}$ & Encoder embeddings with added Gaussian noise (scale $\sigma_E$) \\
\midrule
\multicolumn{2}{@{}l}{\textit{Baselines from the literature}} \\[2pt]
Cocos & Condition-dependent Gaussian prior~\cite{dong2026conditioning} \\
BRIDGER & CVAE-based action prior~\cite{chen2024bridger} \\
Retrieval & $k$-NN-based action prior~\cite{pari2022surprising, he2026demystifying, kai2026retrieval} \\
\bottomrule
\end{tabular}
\end{table}

Our study compares seven types of priors, including the default Gaussian $\mathcal{Z}$, \emph{three we propose} (built directly from the pretrained LBM with no auxiliary training), and \emph{three baselines} from the literature. \cref{tab:priors} summarizes the priors; \cref{sec:appendix_priors} gives the exact sampling procedures.

\textbf{Proposed priors.} 
Our three proposed priors are constructed on the fly from $\pi_{\text{pre}}$, inheriting the structure the pretrained policy has already learned. 
\begin{itemize}
    \item $A_{\text{pre}}$ (\textbf{pretrained actions}) is 
    a partially denoised action chunk from a frozen copy of $\pi_{\text{pre}}$ run for half of its usual denoising budget (``half'' splits the denoising budget evenly between $\pi_{\text{pre}}$ and the fine-tuned action head to avoid extra computational costs).    
    The pretrained model already roughly outputs plausible actions, so the velocity field only needs to cover a shorter transport distance to reach the target (see \cref{fig:opener} left against $\mathcal{Z}$).
    \item $A_{\text{pre}}{+}\sigma_A\mathcal{Z}$ (\textbf{noise-perturbed actions}) is $A_{\text{pre}}$ with added isotropic Gaussian jitter at action-space scale $\sigma_A$. Unperturbed $A_{\text{pre}}$ can collapse the prior too tightly around certain modes of the pretrained policy's output; adding noise enriches the distribution of $A_{\text{pre}}$.
    \item $E_{\text{pre}}{+}\sigma_E\mathcal{Z}$ (\textbf{noise-perturbed embedding}) is the pretrained encoder output embedding $e$, injected with Gaussian noise at embedding-space scale $\sigma_E$ \emph{before} inputting into $\pi_{\text{pre}}$ to produce $a_0$. This perturbs the prior in embedding space rather than action space, exploring a different axis of prior variation.
\end{itemize}
\textbf{Baselines from the literature.} We include three representative external priors effective for policy learning.
\begin{itemize}
    \item \textbf{Cocos}~\cite{dong2026conditioning} is a condition-dependent Gaussian whose mean is produced by passing the observation embedding $e$ through a small encoder into the action space. The encoder is jointly trained with a decoder that maps back to the embedding space of $e$.
    \item \textbf{BRIDGER}~\cite{chen2024bridger} is a CVAE-based action prior that maps a Gaussian latent to an action chunk, conditioned on the observation embedding $e$.
    \item \textbf{Retrieval ($k$-NN)}~\cite{pari2022surprising, he2026demystifying, kai2026retrieval} is a non-parametric prior that averages the $k$ nearest-neighbor action chunks from a frozen bank indexed by the observation embedding $e$. The three cited works used $k$-NN actions directly as policies; we repurpose this approach as a prior for fine-tuning.
\end{itemize}

Some existing non-Gaussian prior methods, such as action-to-action FM~\cite{jia2026a2a} and VITA~\cite{gao2026vita}, replace the policy network with a custom architecture, conflating the effect of the prior with architectural changes. These methods are thus incompatible with fine-tuning from pretrained weights. The seven priors above all preserve the pretrained encoder and action-head weights, so \emph{only the prior differs across experiments}.
\vspace{-5pt}
\section{Experimental Setup}
\label{sec:experiments}

In this section, we detail our experimental setup for evaluating whether and under what conditions non-Gaussian priors benefit policy fine-tuning. We cover: how we control for confounds, test for statistical significance, evaluate in simulation and on robot hardware, and define our plotting conventions.

\vspace{-5pt}
\subsection{Experimental Controls} 
Throughout the paper, we hold the task, data, architecture, optimizer, and number of fine-tuning steps identical across all prior variants within each experimental setting (see~\cref{sec:appendix_implementation} for hyperparameter choices). This protocol applies uniformly to all settings. For prior noise levels, we select $\sigma_A$ and $\sigma_E$ via grid search (\cref{fig:noise_ablation}; values in \cref{sec:appendix_priors}).
\vspace{-5pt}
\subsection{Statistical Testing} 
We treat policy comparison as a formal hypothesis test rather than only reporting empirical average success rates, following prior work~\cite{tri2026lbm, lin2026cotraining, snyder2025step}. For binary success/failure we use the STEP statistical framework~\cite{snyder2025step}; for continuous metrics (per-observation $\ell_2$ distance and cosine similarity) we use Welch's $t$-test~\cite{welch1947ttest}. All tests use a global significance level $\alpha{=}0.05$ (95\% confidence level). When more than two priors are compared we apply Bonferroni correction: the family-wise false-positive probability of $m$ independent pairwise tests at level $\alpha$ is $1-(1-\alpha)^m$. Bonferroni controls this by testing each pair at $\alpha/m$ such that the overall false-positive rate is upper-bounded by $\alpha$. A compact letter display (CLD)~\cite{piepho2004cld} concisely summarizes each method comparison: two methods share at least one alphabetical letter \emph{if and only if} their pairwise difference is not significant at the corrected level. Implementation details (STEP's anytime-valid guarantee, Lai's~\cite{lai1988sequential} fallback at per-method sample sizes above ${\sim}600$, and CLD ordering conventions) are in \cref{sec:appendix_stats}.

\vspace{-5pt}
\subsection{Hardware A/B testing} 
Following best practices for robot policy evaluation~\cite{kress2024robot}, hardware rollouts use blind A/B testing: for each fixed initial condition (IC), all priors under evaluation are run in a randomized order before moving to the next IC,
ensuring the human operator does not know which prior is being tested at any given trial. This design eliminates operator bias and, because priors are interleaved rather than grouped by prior, distributes any ambient drift (e.g., lighting, wear, operator attention) approximately equally across all priors.

\vspace{-5pt}
\subsection{Benchmarks} 
We use two simulators and one hardware setup.
The \emph{LBM Eval}~\cite{tri2026lbm} Drake-based~\cite{drake} simulation benchmark serves as our main testbed. It consists of eight bimanual manipulation tasks of varying difficulty and sample sizes (approximately 196--392 per task at 100\%) that were unseen during the pretraining phase for LBM 1.0~(see~\cite{tri2026lbm} for pretraining data selection).
We also evaluate on the \emph{RoboCasa365}~\cite{nasiriany2026robocasa} MuJoCo-based~\cite{todorov2012mujoco} simulation suite. This benchmark is split into 18 \emph{atomic} tasks (single-step manipulations, e.g., opening a cabinet) and 16 \emph{composite} tasks (multi-step sequences, e.g., preparing coffee), totaling 34 tasks.
Hardware is a tabletop bimanual setup with two Franka FR3 arms and parallel grippers. Each rollout is scored by a human with a binary success label and a task-specific \emph{partial task progress} rubric. The latter decomposes the task into ordered milestone predicates (e.g., \emph{rotor picked up}, \emph{rotor seated}, \emph{lockring tightened}) and reports the fraction of milestones achieved. Hardware details and per-task demonstration counts are in \cref{sec:appendix_implementation}.

\subsection{Plotting Conventions} 
In all figures, an ($\uparrow$) arrow indicates higher is better and a ($\downarrow$) arrow lower is better. Violin plots for binary success rates show Bayesian posterior distributions; a black horizontal tick marks the posterior mean and a colored dot marks the empirical mean. Violin plots for continuous metrics (e.g., $\ell_2$ distance, cosine similarity, task progress) show empirical distributions with a colored dot at the empirical mean. All plots are annotated with CLD letters: methods sharing a letter are statistically indistinguishable; methods with different letters are statistically separable.

\begin{table*}[t]
\centering
\caption{Index of all experimental settings, ablations, and analyses.
Each row links to its corresponding figure(s).
$\dagger$\,Reuses the rollouts of its corresponding main setting (no additional evaluations).}
\label{tab:overview}
\renewcommand{\arraystretch}{1.3}
\setlength{\tabcolsep}{5pt}
\begin{tabular}{@{}lp{8.8cm}llr@{}}
\toprule
\textbf{Setting} & \textbf{Description} & \textbf{Model} & \textbf{Figures} & \textbf{\# Rollouts} \\
\midrule
\multicolumn{5}{@{}l}{\textit{Main experiments}} \\[2pt]
A & Do closer-to-target priors improve fine-tuning performance? & LBM 1.0~\cite{tri2026lbm} & \ref{fig:opener}, \ref{fig:setting_a_full} & 14000 \\
B\textsubscript{sim} & Do priors matter across varying data fractions? & LBM 1.0~\cite{tri2026lbm} & \ref{fig:setting_b} & 56000 \\
\quad Analysis & How similar are fine-tuned models across priors? & & \ref{fig:logme}--\ref{fig:convergence_cosine} & $\dagger$ \\
\quad Encoder LR & Does the encoder matter more than the prior? & & \ref{fig:encoder_lr} & 9600 \\
B\textsubscript{real} & Do priors matter on robot hardware? & LBM 1.0~\cite{tri2026lbm} & \ref{fig:setting_b_real} & 1250 \\
\quad Encoder LR & Does the encoder-LR effect carry over to hardware? & & \ref{fig:encoder_lr} & 500 \\
\multirow{2}{*}{C} & \multirow{2}{=}{Do priors matter across LBM architectures?} & $\pi_{0.5}$~\cite{black2025pi05} & \multirow{2}{*}{\ref{fig:setting_c}} & 27200 \\
 & & GR00T~N1.5~\cite{gr00tn1_2025} &  & 27200 \\
\midrule
\multicolumn{5}{@{}l}{\textit{Appendix controls \& ablations}} \\[2pt]
B\textsubscript{sim}: easy/hard & Does task difficulty affect prior behavior? & LBM 1.0~\cite{tri2026lbm} & \ref{fig:setting_b_seen_unseen} & $\dagger$ \\
B\textsubscript{FP} & Do priors depend on the pretraining objective? & LBM 1.0~\cite{tri2026lbm} & \ref{fig:setting_bfp} & 7000 \\
\quad Analysis & Do the \cref{sec:analysis} diagnostics hold for flow-matching-only training? & & \ref{fig:bfp_logme}--\ref{fig:bfp_repr_change_cosine} & $\dagger$ \\
\quad Noise $\sigma$ & How do noise-perturbed priors scale with $\sigma$? & & \ref{fig:noise_ablation} & 6000 \\
B\textsubscript{real}: pilot & Which learned prior to carry forward to the full hardware evaluation (small real-robot pilot)? & LBM 1.0~\cite{tri2026lbm} & \ref{fig:hw_pilot} & 225 \\
\bottomrule
\end{tabular}
\end{table*}
\vspace{-0.2cm}
\section{Research Questions and Results}
\label{sec:results}
Our experiments are guided by three research questions:
\begin{itemize}
    \item[(Q1)] Do non-Gaussian priors that are measurably closer to the target improve policy fine-tuning performance?
    \item[(Q2)] Do informed priors have an advantage across data fractions, particularly in the low-data regime where the added inductive bias may be most beneficial?
    \item[(Q3)] Is the effect of prior choice consistent across model architectures as well as simulation and hardware experiments?
\end{itemize}
We design four experimental settings, summarized in \cref{tab:overview}, to carefully study these three questions. Setting~A isolates Q1 using a contrived prior that is by construction closer in $\ell_2$ distance to the target action distribution than the Gaussian.
Setting~B\textsubscript{sim} addresses both Q1 and Q2 by varying the amount of fine-tuning data across five fractions on eight tasks.
Setting~B\textsubscript{real} tests Q3 on whether the result transfers to hardware.
Setting~C further studies Q3 on other LBMs by repeating the experiments on $\pi_{0.5}$ and GR00T~N1.5.
We consistently find that non-Gaussian priors do not improve fine-tuning, with a possible exception in very low-data regimes.
Alongside each main setting we run targeted ablations (encoder learning rate, training objective, noise scale), each isolating a single confounding variable.

\vspace{-5pt}
\subsection{Setting A: Fraction-Pretrained Prior}
\label{sec:setting_a}

The key challenge in studying non-Gaussian priors for fine-tuning is controlling prior quality: a pretrained LBM has never seen the fine-tuning task, so its zero-shot actions may be out-of-distribution and form a weak prior. A null result would then be hard to tell apart from a poor-quality prior. Setting~A therefore isolates this question by constructing an artificially strong prior that is measurably close to the target in $\ell_2$ distance. We use a two-stage pipeline: first, the pretrained LBM 1.0 is fine-tuned using the Gaussian prior on 50\% of the demonstrations to produce a task-aware checkpoint whose actions are much closer to the target than the Gaussian (\cref{fig:opener}, left); second, this checkpoint serves as the ``pretrained model'' for another round of fine-tuning, in which we vary only the prior. We evaluate on five kitchen-scenario tasks from LBM Eval (see \cref{tab:appendix_tasks_sim} for full names) and compare all seven priors from \cref{tab:priors}, with 200 rollouts per task (1000 total per prior).

\begin{figure}[t]
    \centering
    \includegraphics[width=\linewidth]{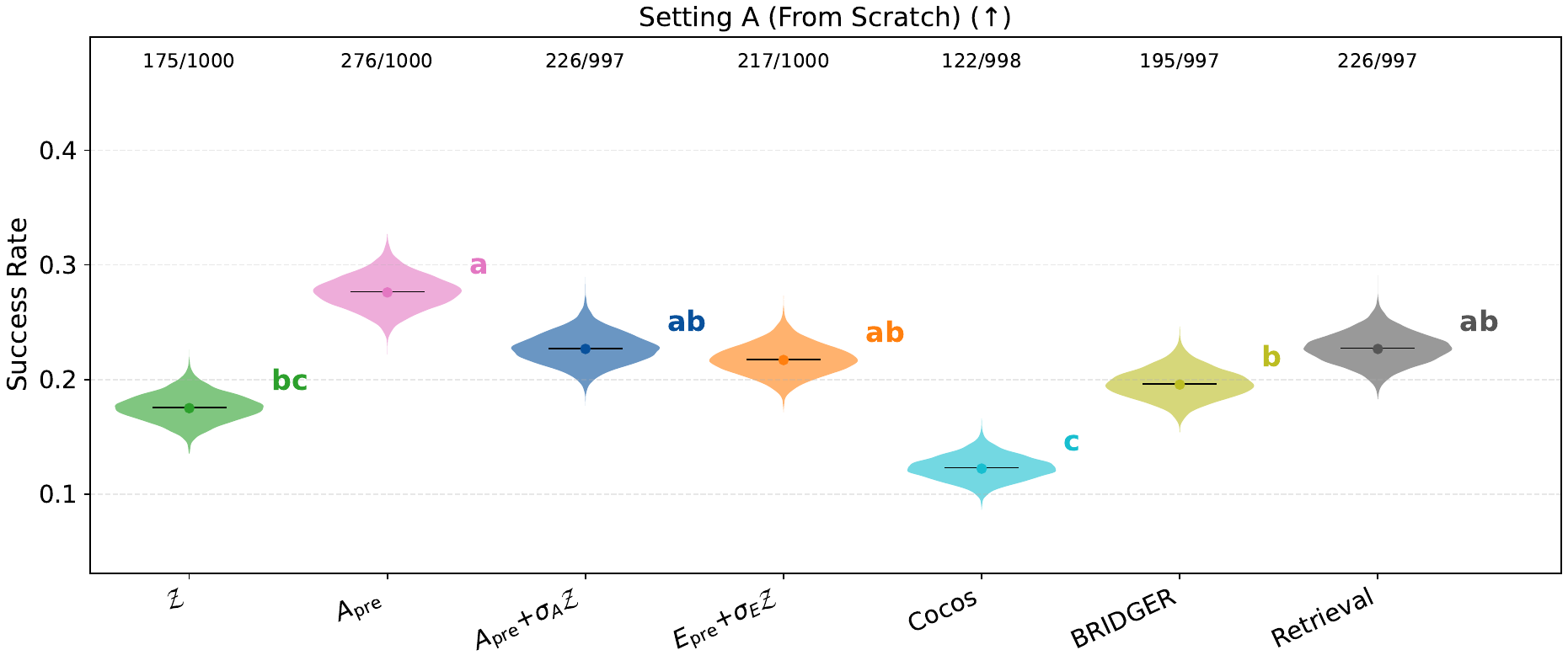}\\[4pt]
    \includegraphics[width=\linewidth]{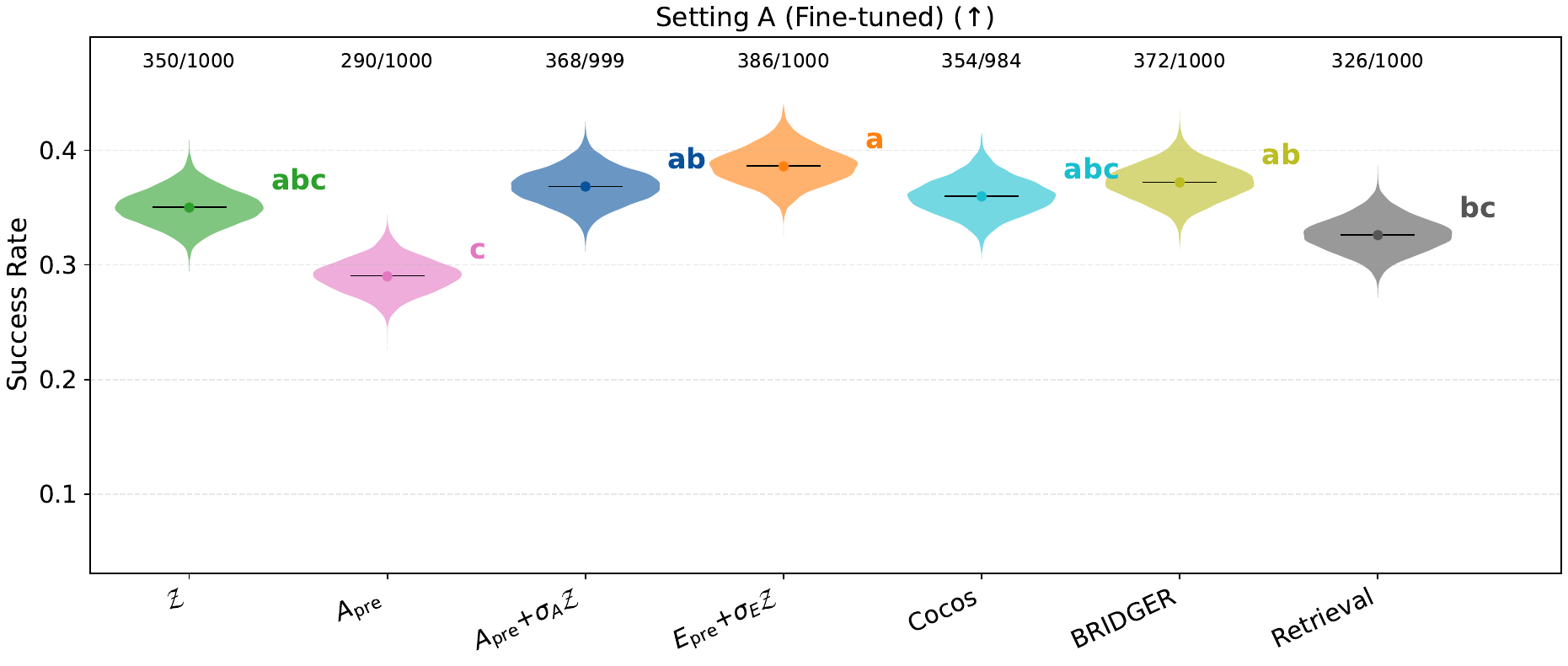}
    \caption{Setting A: full seven-prior comparison; each violin corresponds to 1000 rollouts.
    \textbf{Top:} from scratch; $A_{\text{pre}}$ statistically outperforms $\mathcal{Z}$, which is second lowest in empirical success rate among all priors.
    \textbf{Bottom:} fine-tuned; all priors perform comparably to $\mathcal{Z}$ with six of seven priors sharing CLD group ``a'' or ``b''.}
    \label{fig:setting_a_full}
\end{figure}

\textbf{Results.} \cref{fig:setting_a_full} summarizes the key finding across all seven prior variants.
When training from scratch (i.e., training the same LBM~1.0 architecture from random weight initialization), all non-Gaussian priors (except for Cocos; see \cref{sec:appendix_priors})
outperform the Gaussian baseline $\mathcal{Z}$ based on empirical success rates, though they are largely statistically indistinguishable. The simplest of our proposed priors, $A_{\text{pre}}$, significantly outperforms $\mathcal{Z}$ (CLD ``a'' vs.\ ``bc'') with an average success rate of roughly 28\% vs.\ 18\%. This is consistent with prior work on non-Gaussian priors in the from-scratch regime~\cite{chen2024bridger, dong2026conditioning, jia2026a2a}.
Nevertheless, this empirical advantage disappears entirely in the fine-tuning regime: $A_{\text{pre}}$ becomes the worst-performing method based on success rate and no prior is statistically separable from the Gaussian. Additionally, the posterior distributions overlap more substantially in the fine-tuning case than in the from-scratch case, reflecting the lesser impact of action priors on policy performance during fine-tuning.

\subsection{Setting B\texorpdfstring{$_{\text{sim}}$}{sim}: LBM Fine-Tuning in Simulation}
\label{sec:setting_b}

Setting~A tests Q1 and finds no prior advantage under a controlled good prior at small scale. Two questions remain for the practical fine-tuning regime. First, do priors matter at full LBM scale without the artificial construction in Setting A
(Q1)? Second, how do learned priors perform across fine-tuning data fractions, especially in the low-data regime that may benefit from inductive biases of learned priors (Q2)?
To address both questions, we fine-tune the LBM 1.0 model on the eight LBM Eval unseen tasks and sweep over multiple data fractions (5\%, 25\%, 50\%, 75\%, 100\% of available demonstrations). This mirrors the standard LBM deployment pipeline~\cite{tri2026lbm, lin2026cotraining}.
Of the eight tasks, three are comparatively easier (higher success rates under the standard Gaussian) and five are harder; we report both the overall average and the per-task breakdown (\cref{sec:appendix_results}). Each of the seven priors is evaluated with 200 rollouts per prior--task--fraction combination, totaling 56,000 simulation rollouts for this setting.

\begin{figure}[t]
    \centering
    \includegraphics[width=\linewidth]{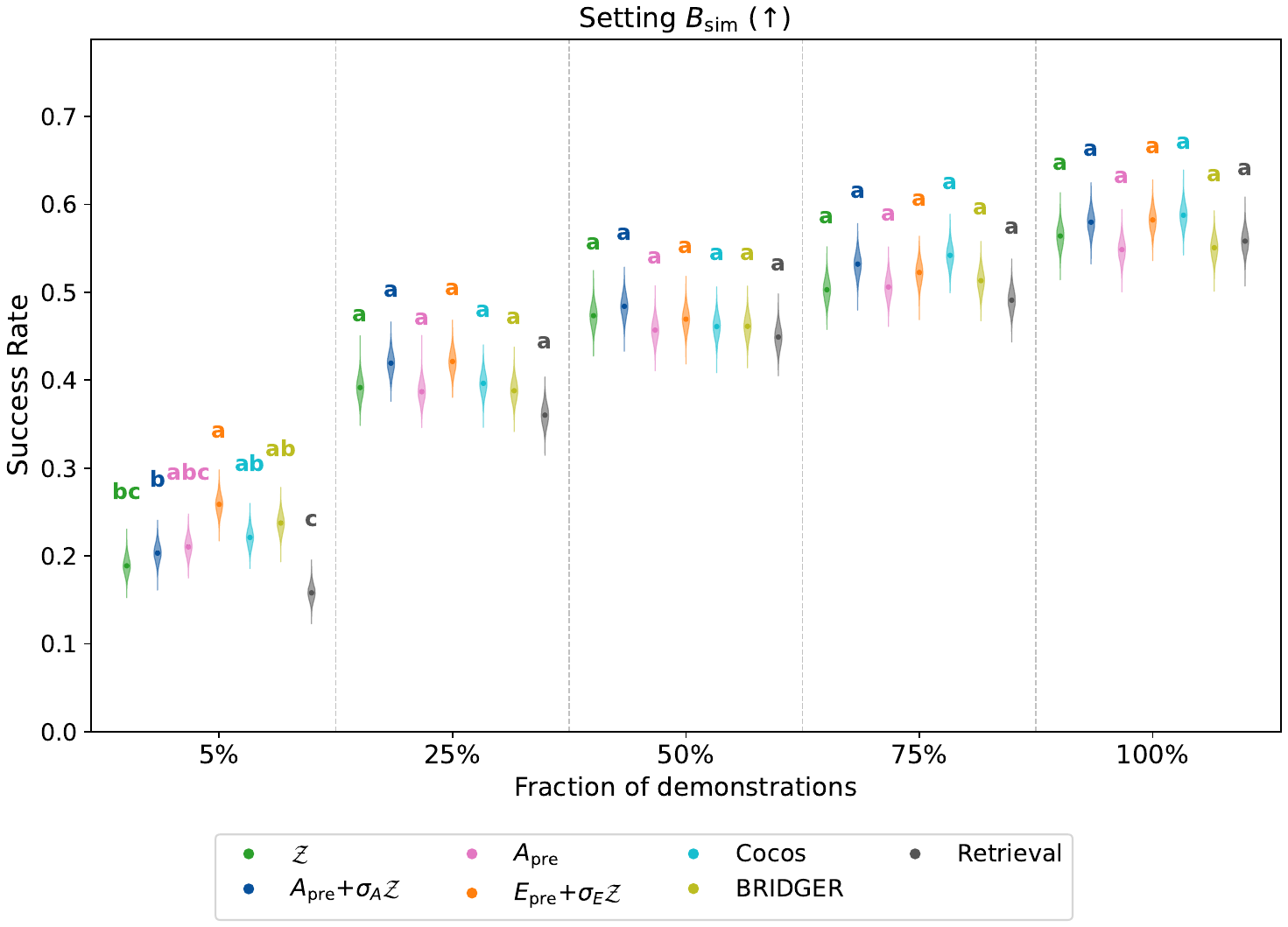}
    \caption{Fine-tuning performance across data fractions (Setting B\textsubscript{sim}, eight tasks).
    CLD letters are computed within each data fraction independently. All priors are statistically indistinguishable or worse than $\mathcal{Z}$ across fractions. The one exception is $E_{\text{pre}}{+}\sigma_E\mathcal{Z}$ at 5\% data, which significantly outperforms $\mathcal{Z}$.}
    \label{fig:setting_b}
\end{figure}

\textbf{Results.} \cref{fig:setting_b} shows the success rates averaged across the eight simulation tasks as a function of data fraction.
At every data fraction from 25\% upward, most prior variants cluster closely and all share the CLD letter ``a''.
The one exception is at 5\% data, where $E_{\text{pre}}{+}\sigma_E\mathcal{Z}$ outperforms $\mathcal{Z}$ (CLD ``a'' vs.\ ``bc,'' 25.9\% vs.\ 18.9\% success rate).
Intuitively, at such a low data fraction the demonstrations cover only a small portion of the target action space, and the embedding-space jitter from $E_{\text{pre}}{+}\sigma_E\mathcal{Z}$ appears to act as an informative inductive bias that the Gaussian cannot provide. This very low-data regime (approximately 10--20 demonstrations per task) is the one exception in our study where a learned prior helps.

\subsection{Setting B\texorpdfstring{$_{\text{real}}$}{real}: LBM Fine-Tuning on Robot Hardware}
\label{sec:setting_b_real}

Settings A and B\textsubscript{sim} both rely on simulation, which could in principle mask simulation-specific artifacts. The hardware setting tests whether the finding holds under real-world conditions that cannot be replicated in simulation.
We fine-tune LBM 1.0 on five bimanual FR3 hardware tasks, spanning varying difficulty levels: \textit{FoodBank}, \textit{ClearKitchenCounter}~\cite{tri2026lbm}, \textit{GatheringIngredientsForCookies}, \textit{BusBin}, and \textit{BikeRotorInstall}~\cite{tri2026lbm}. Per-task descriptions and demonstration counts are in \cref{sec:appendix_implementation}. 
We compare five of the seven prior variants: $\mathcal{Z}$ (the standard Gaussian baseline), $E_{\text{pre}}{+}\sigma_E\mathcal{Z}$, Cocos, BRIDGER\footnote{Its CVAE training was unstable on both simulation and hardware datasets (NaN losses, requiring restarts with different random seeds until stable); despite this instability, stable runs of BRIDGER were included in the evaluation.}, and Retrieval, with 50 A/B-test rollouts per prior per task (250 per prior, 1250 total; protocol in \cref{sec:experiments}). The remaining two priors, $A_{\text{pre}}$ and $A_{\text{pre}}{+}\sigma_A\mathcal{Z}$, were dropped after a pilot study (\cref{sec:appendix_hw_pilot}) in which all three action-space priors were statistically tied and $E_{\text{pre}}{+}\sigma_E\mathcal{Z}$ was selected as the representative learned prior for the full evaluation.

\begin{figure}[t]
    \centering
    \includegraphics[width=\linewidth]{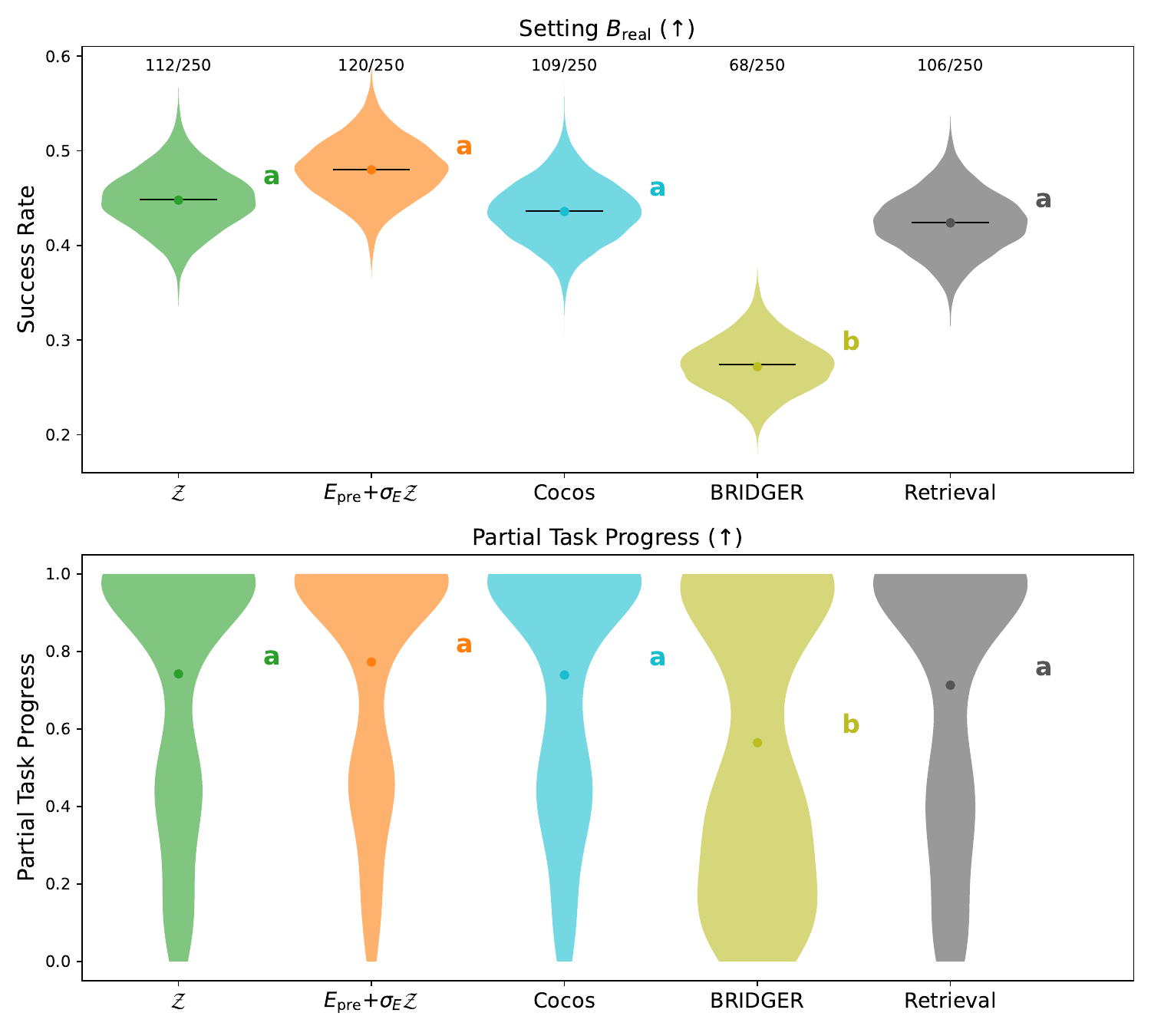}
    \caption{Setting B\textsubscript{real}: hardware A/B testing (five bimanual FR3 tasks, 50 rollouts per prior per task, 1250 total).
    \textbf{Top:} Success rate violins; four of five priors share CLD group ``a'' and BRIDGER (group ``b'') underperforms.
    \textbf{Bottom:} Partial-task-progress violins; the same grouping holds.}
    \label{fig:setting_b_real}
\end{figure}

\textbf{Results.} \cref{fig:setting_b_real} shows full results across the five bimanual FR3 hardware tasks.
Out of 250 A/B-test rollouts per prior, four of the five priors (i.e., $\mathcal{Z}$, $E_{\text{pre}}{+}\sigma_E\mathcal{Z}$, Cocos, and Retrieval) fall in the same CLD group ``a,'' with partial-progress scores in the 71--77\% range.
BRIDGER is the sole outlier (group ``b,'' partial progress ${\sim}56\%$), consistent with the training instability noted earlier and documented in \cref{sec:appendix_priors}.
Excluding BRIDGER, the pattern mirrors our simulation findings: the choice of prior has no statistically significant effect on fine-tuning performance.

\subsection{Setting C: Validation on Other LBM Architectures}
\label{sec:setting_c}

Settings A, B\textsubscript{sim}, and B\textsubscript{real} all use LBM 1.0 as the backbone, which uses a CLIP encoder and DiT action head. Setting C tests whether the results generalize across model architectures and simulation environments.
We repeat the comparison on the RoboCasa simulation benchmark using two additional LBMs: $\pi_{0.5}$~\cite{black2025pi05} (fused SigLIP + Gemma 2B with interleaved cross-attention) and GR00T~N1.5~\cite{gr00tn1_2025} (Eagle-2 VLM + diffusion transformer action expert). 

Each model is fine-tuned on each of the 34 RoboCasa tasks and evaluated with 200 rollouts per task (see \cref{sec:appendix_per_task} for per-task breakdowns). The fused architecture constrains which priors can be ported from the LBM 1.0 experiments. Of the seven priors from Setting~B\textsubscript{sim}, four have natural analogues here: $\mathcal{Z}$ (Gaussian), $A_{\text{pre}}$ (pretrained actions), $A_{\text{pre}}{+}\sigma_A\mathcal{Z}$ (action~+~noise), and Retrieval ($k$-NN in the LBM's vision-encoder feature space, $k{=}3$). We construct $A_{\text{pre}}$ for each model from a frozen copy of itself (e.g., $\pi_{0.5}$'s $A_{\text{pre}}$ uses frozen $\pi_{0.5}$). The embedding-space priors $E_{\text{pre}}{+}\sigma_E\mathcal{Z}$, Cocos, and BRIDGER are not immediately portable because both LBMs lack a single bottleneck embedding: visual and action tokens are fused across many transformer layers with per-layer cross-attention, leaving no natural injection point. They are thus omitted in this setting.

\begin{figure*}[t]
    \centering
    \includegraphics[width=\textwidth]{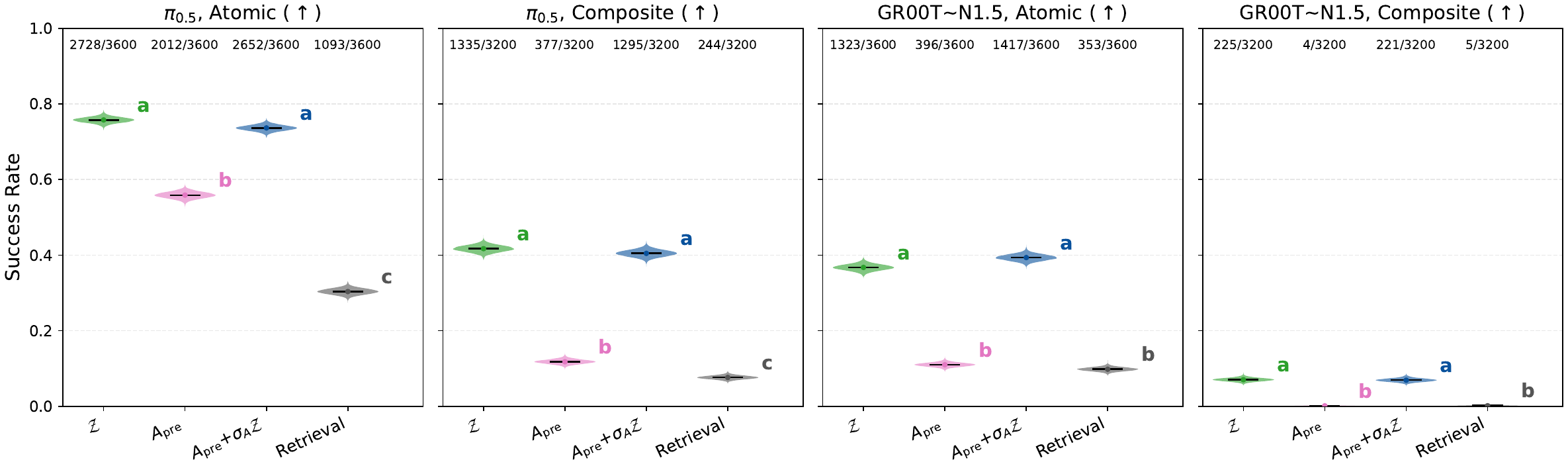}
    \caption{Setting C: LBM fine-tuning on RoboCasa, with shared y-axis (Success Rate, 0--1) across all four panels.
    \textbf{Left two panels:} $\pi_{0.5}$ on 18 atomic (3600 rollouts per prior) and 16 composite (3200 rollouts per prior) tasks.
    \textbf{Right two panels:} GR00T~N1.5 on the same splits. Both architectures share the same qualitative ordering; $\mathcal{Z}$ and $A_{\text{pre}}{+}\sigma_A\mathcal{Z}$ tie at the top on both task types, while $A_{\text{pre}}$ and Retrieval priors underperform.}
    \label{fig:setting_c}
\end{figure*}

\textbf{Results.} \cref{fig:setting_c} reports success rates for atomic and composite tasks separately for each of the two LBMs. Both architectures exhibit the same qualitative ordering across task types: $\mathcal{Z}$ and $A_{\text{pre}}{+}\sigma_A\mathcal{Z}$ tie at the top in CLD group ``a,'' while the pure learned prior $A_{\text{pre}}$ falls into group ``b''; Retrieval is either ``b'' or in a distinct group ``c.''
Despite the observed performance degradation of $A_{\text{pre}}$ and Retrieval, $A_{\text{pre}}$ remains ${\sim}3\times$ closer to the target than $\mathcal{Z}$ in $\ell_2$ on RoboCasa (\cref{fig:setting_c_prior_distance}), the same factor as in Setting~A on LBM Eval (\cref{fig:opener}, left). Yet $A_{\text{pre}}$ degrades only here in Setting~C, indicating that geometric proximity to the target is not what drives the gap. A key difference in this setting compared to the earlier ones is that LBM 1.0's pretraining includes simulation data from the same simulator, whereas $\pi_{0.5}$ and GR00T~N1.5 were pretrained predominantly on real-world data. This domain gap specifically hurts $A_{\text{pre}}$ and Retrieval because both rely solely on a frozen copy of the pretrained LBM (for action generation and for building the retrieval bank, respectively). We hypothesize that the other two priors, $\mathcal{Z}$ and $A_{\text{pre}}{+}\sigma_A\mathcal{Z}$, are regularized by random noise and thus are less affected by this domain gap.

\section{Analysis: Why Does the Prior Not Matter Under Fine-Tuning?}
\label{sec:analysis}

Summarizing the negative-result experiments against our research questions, we found that Q1 (closer priors should help) yields a negative answer, Q2 (advantage across fine-tuning data fractions) is only partly supported at 5\% data, and Q3 (generalization across policies and hardware) also yields a negative answer. The experimental results thus raise a central question: if a non-Gaussian prior is measurably closer to the target distribution (in Setting A by construction, and in Settings~B and~C empirically), why does this proximity not translate to better fine-tuning performance?

We approach this question from two complementary perspectives.
First, we ask whether fine-tuned models \emph{converge toward the target} despite starting from different priors. We examine convergence at two levels: whether different priors produce encoder features with comparable predictive power for the target actions (\cref{sec:logme}), and whether the resulting action predictions are similarly close to ground truth (\cref{sec:action_convergence}).
Second, we ask whether fine-tuned models \emph{diverge from a shared starting point}: that is, whether different priors lead to structurally different optimization trajectories, as measured by $\ell_2$ distance and cosine similarity in the action or embedding space (\cref{sec:convergence}). The observed embedding-space divergence motivates a natural hypothesis: fine-tuning performance is governed primarily by the observation encoder rather than by the prior itself. We test this hypothesis directly through a learning-rate ablation that deliberately undertrains the encoder (\cref{sec:encoder_lr}).

All analyses in this section are carried out in Setting~B\textsubscript{sim} (LBM 1.0 simulation), where we have the most comprehensive results: seven prior variants $\times$ five data fractions $\times$ eight tasks, enabling controlled comparisons across the prior--data landscape. We focus on four representative priors: $\mathcal{Z}$ (Gaussian), $A_{\text{pre}}$ (pure learned prior, closest to target), $A_{\text{pre}}{+}\sigma_A\mathcal{Z}$ (action~+~noise; defined in \cref{sec:prior_variants}), and $E_{\text{pre}}{+}\sigma_E\mathcal{Z}$ (embedding~+~noise). 
The baselines (Cocos, BRIDGER, Retrieval) are omitted from these introspective ablations as their performance was found to be worse than or comparable to these four priors.
Action-space metrics are computed on the flattened action chunk.
Embedding-space metrics are computed on the observation embedding $e$ defined in \cref{sec:problem}.
For notation, we reuse $A_{\text{pre}}$ from \cref{sec:prior_variants} (actions drawn from the frozen pretrained policy) and introduce $A_{\text{post}}$ for the corresponding actions drawn from the fine-tuned policy. The observation embedding $e$ (\cref{sec:problem}) has two instances depending on the encoder checkpoint: $E_{\text{pre}} = e_{\pi_{\text{pre}}}(o)$ is the embedding from the pretrained encoder, and $E_{\text{post}} = e_{\pi_{\text{post}}}(o)$ is the embedding from the fine-tuned encoder, evaluated on the same $o$.

\subsection{Feature Quality: LogME Analysis}
\label{sec:logme}

\begin{figure}[t]
    \centering
    \includegraphics[width=\linewidth]{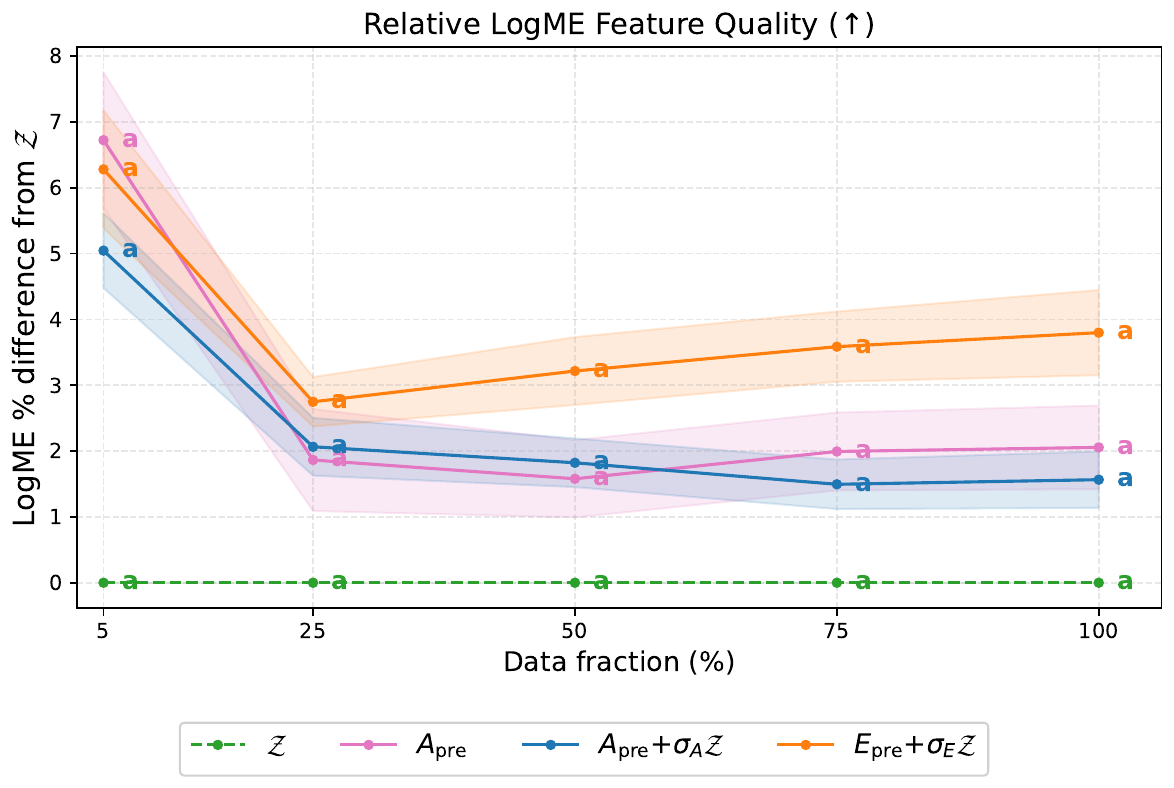}
    \caption{LogME feature quality as percentage difference from $\mathcal{Z}$ at each fraction (Setting B\textsubscript{sim}, eight tasks). A value of five means LogME for that prior is 5\% higher than $\mathcal{Z}$ (dashed green line at zero) at the same fraction. Ribbons show $\pm 1$ standard error across the eight tasks. We normalize because raw LogME drops sharply from 5\% to 25\% data due to a small-sample artifact of the LogME computation (see raw values in \cref{fig:logme_raw}); normalizing to $\mathcal{Z}$ removes this artifact since we only compare across priors at the same sample size. CLD letters (computed on raw values) confirm all priors share group ``a'' at every fraction.}
    \label{fig:logme}
\end{figure}

We begin at the feature layer, asking whether fine-tuned observation encoders produce features of equal predictive power for the target actions, regardless of the prior used during training.
We measure this with LogME~\cite{you2021logme}, a training-free proxy for feature quality that estimates the log marginal evidence of a Bayesian linear regression from encoder features $E_{\text{post}}$ to ground-truth actions $A_{\text{target}}$.
LogME produces a single aggregate score per prior--task--fraction combination by fitting a weight matrix to all (observation-embedding, action) pairs.
Higher LogME indicates more predictive encoder features for action prediction.
As shown in \cref{fig:logme}, LogME scores are inseparable across all priors. The prior thus governs how much the encoder restructures, but not the quality of the final result: all priors arrive at features that are equally predictive of actions (\cref{sec:logme}).

\subsection{Action-Space Similarity After Fine-Tuning}
\label{sec:action_convergence}

\begin{figure}[t]
    \centering
    \includegraphics[width=\linewidth]{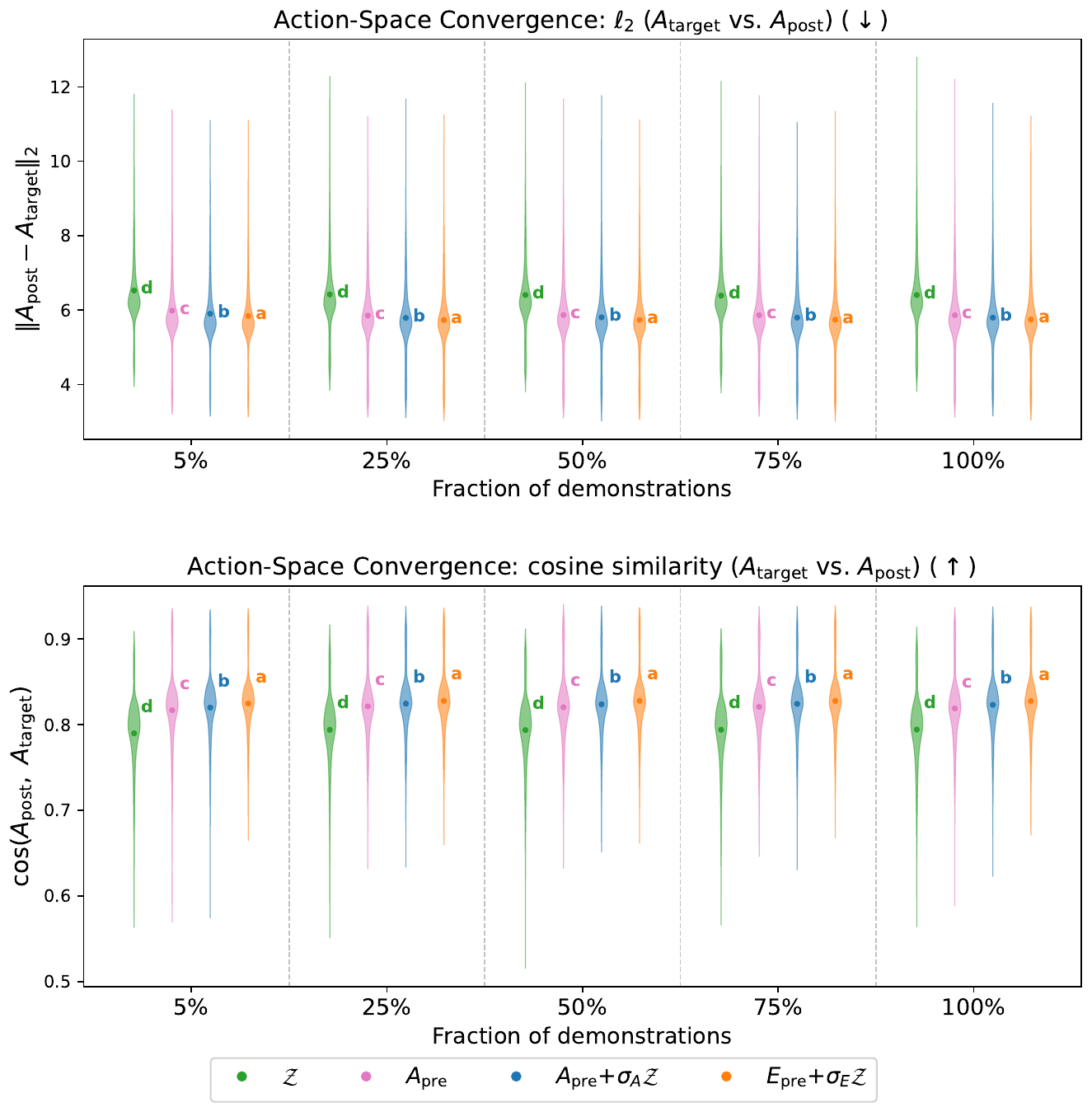}
    \caption{Action-space similarity after fine-tuning (Setting B\textsubscript{sim}, eight tasks). Violins show per-observation distances between predicted and ground-truth actions at each fraction, with CLD letters from Welch's $t$-test.
    \textbf{Top:} $\ell_2$ distance ($\downarrow$). All four priors cluster within a narrow range (5.7--6.4), with the Gaussian consistently ${\sim}0.5$ higher than the non-Gaussian priors.
    \textbf{Bottom:} cosine similarity ($\uparrow$). All four priors largely overlap at every fraction (means within ${\sim}0.79$--$0.83$).
    Note: per-prior sample sizes are very large ($n{\approx}16000$), so CLD letters here flag any detectable mean difference and may not reflect practical separation.}
    \label{fig:action_convergence}
\end{figure}

\cref{sec:logme} established convergence at the feature layer: fine-tuned observation encoders are equally predictive of actions across priors. We now examine convergence at the output layer, asking whether the post-fine-tuning predicted actions $A_{\text{post}}$ converge to the ground-truth actions $A_{\text{target}}$ in the target demonstration data. 
%
Quantitatively, we measure the per-observation $\ell_2$ distance (computed over random samples in the demonstration data; \cref{fig:action_convergence}) and the per-observation cosine similarity (\cref{fig:action_convergence}).
The two metrics probe complementary aspects of the action discrepancy: $\ell_2$ captures both magnitude and directional errors, while cosine similarity is scale-invariant and isolates directional agreement.
We also note that the per-prior sample size in these distributions is very large ($n{\approx}16000$ per prior per fraction), so Welch's $t$-test with Bonferroni correction can detect extremely small mean differences. CLD letters here should therefore be interpreted as indicators of \emph{any} detectable mean difference, not as indicators of practically significant separation.

The post-fine-tuning action predictions are broadly similar across priors under both metrics, with violin distributions that overlap substantially at every fraction.
Under $\ell_2$ (\cref{fig:action_convergence}), the Gaussian prior ($\mathcal{Z}$) consistently shows a slightly higher (worse) mean distance (${\sim}6.4$) than the non-Gaussian priors (${\sim}5.7$--$5.9$), with the absolute gap remaining below 0.8 at every data fraction.
Under cosine similarity (\cref{fig:action_convergence}), the four priors still have largely overlapping violins, with means within ${\sim}0.79$--$0.83$ at every fraction; again, $\mathcal{Z}$ is consistently slightly worse (lower cosine similarity).
Note that this small open-loop gap in both metrics does not translate into closed-loop success rate differences (\cref{fig:setting_b}, except against $E_{\text{pre}}{+}\sigma_E\mathcal{Z}$ at 5\%), suggesting it is within the noise tolerance of the downstream rollout evaluation.
These action-space results confirm that post-fine-tuning predictions are similar across priors, consistent with the equal feature quality reported in \cref{sec:logme}.

\subsection{Representation Divergence Despite Similar Outputs}
\label{sec:convergence}

\begin{figure}[t]
    \centering
    \includegraphics[width=\linewidth]{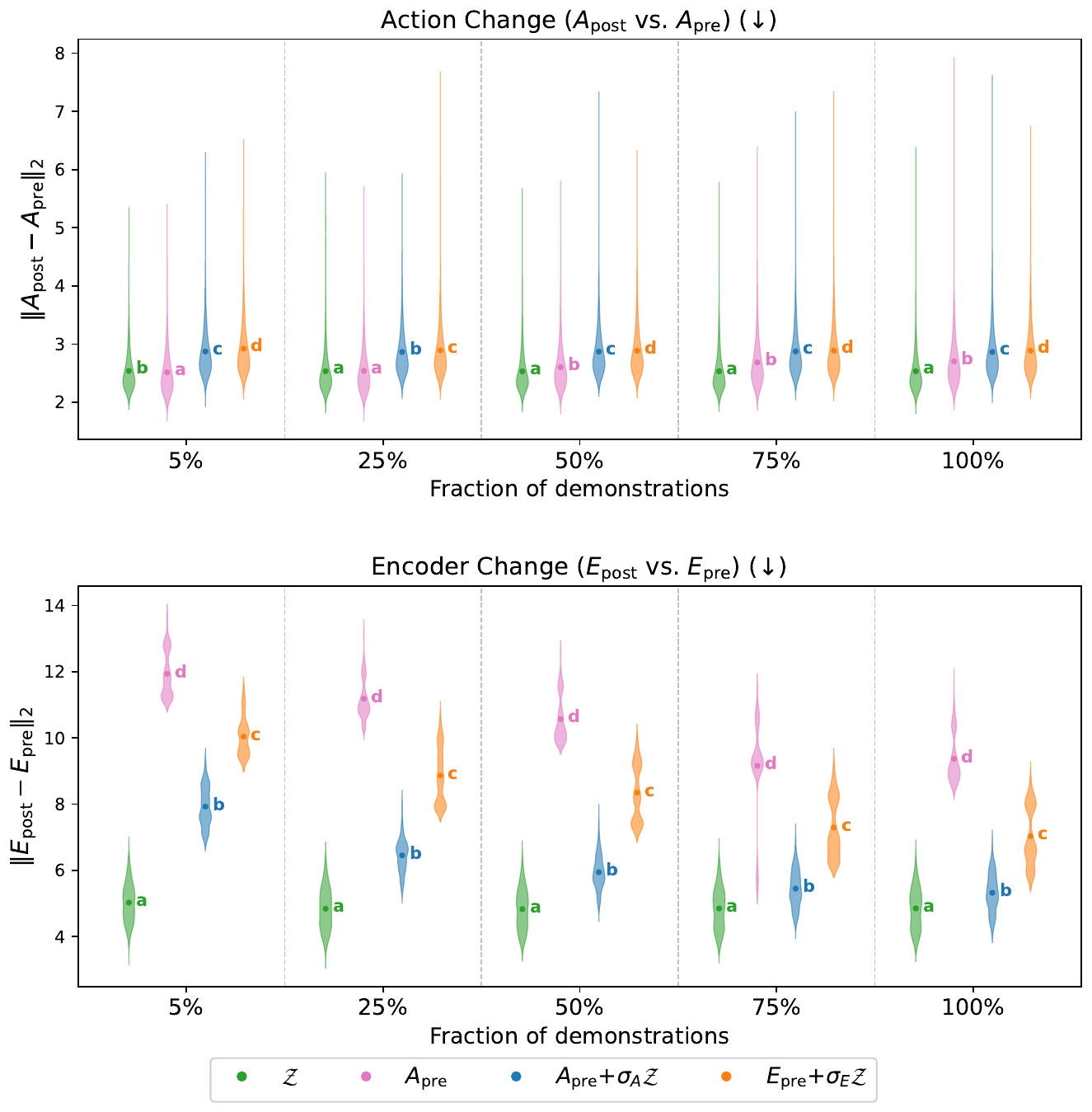}
    \caption{How much does fine-tuning change the model's internal representations? (Setting B\textsubscript{sim}, eight tasks.)
    Violins show per-observation $\ell_2$ distances with CLD letters from Welch's $t$-test at each fraction.
    \textbf{Top:} action change ($A_{\text{post}}$ vs.\ $A_{\text{pre}}$); all priors cluster within a narrow band (2.5--2.9).
    \textbf{Bottom:} encoder change ($E_{\text{post}}$ vs.\ $E_{\text{pre}}$); the four priors show large, visually apparent separation between violin distributions at every fraction (unlike the action change above, where the four CLD groups arise from small mean differences within an otherwise largely overlapping band), with $A_{\text{pre}}$ restructuring the encoder substantially (mean $\ell_2$ roughly 9--12 across fractions) while $\mathcal{Z}$ moves much less ($\ell_2 \approx 5$).}
    \label{fig:convergence}
    \vspace{-15pt}
\end{figure}

\begin{figure}[t]
    \centering
    \includegraphics[width=\linewidth]{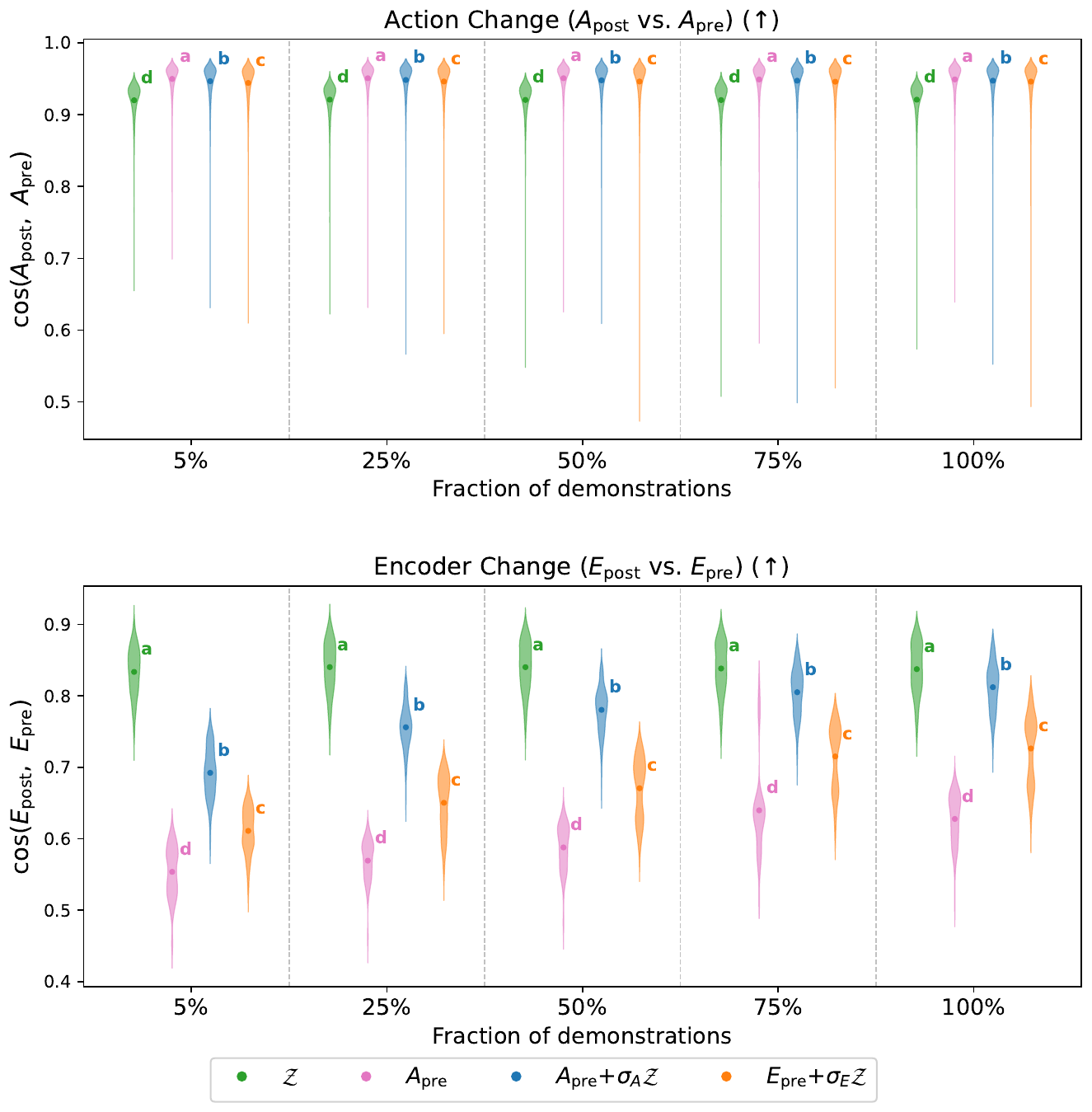}
    \caption{Representation change, cosine-similarity view (Setting B\textsubscript{sim}, eight tasks); companion to \cref{fig:convergence}.
    Higher values indicate less directional rotation from the pretrained representation.
    \textbf{Top:} action change; all priors largely overlap ($\cos \approx 0.92$--$0.95$).
    \textbf{Bottom:} encoder change; $\mathcal{Z}$ retains the highest similarity ($\cos \approx 0.84$) while $A_{\text{pre}}$ rotates the most ($\cos \approx 0.55$--$0.64$), with the mixed priors in between.}
    \label{fig:convergence_cosine}
\end{figure}

The preceding two analyses together characterize convergence \emph{toward the target}: fine-tuned observation encoders produce features of indistinguishable quality for action prediction (\cref{sec:logme}), and their action predictions reach the ground truth with broadly comparable accuracy (\cref{sec:action_convergence}).
We now shift perspective and measure \emph{divergence from the pretrained starting point}: how much do the model's representations change during fine-tuning, and does the amount of change depend on the prior? If different priors induce different amounts of change, this would indicate that prior choices affect the optimization landscape yet admit multiple equally good solutions as judged by closed-loop performance.
We quantify this also at two levels (action-space and embedding-space) using two complementary metrics: the per-observation $\ell_2$ distance between pre- and post-fine-tuning outputs (\cref{fig:convergence}) and the corresponding per-observation cosine similarity (\cref{fig:convergence_cosine}). Note that absolute $\ell_2$ values are not directly comparable across the two panels because the action and embedding spaces have different dimensionalities; cosine similarity is scale-free and enables direct comparison.

\textbf{Action-space change.}
Across all priors, the post-fine-tuning actions remain close to their pretrained counterparts: both $\ell_2$ metrics ($\sim$2.5--2.9) and cosine similarity metrics ($\sim$0.92--0.95) span a narrow band (top figures of \cref{fig:convergence} and \cref{fig:convergence_cosine}).
The takeaway is that fine-tuning makes comparable adjustments to the action output across priors, though $A_{\text{pre}}$ shifts slightly less (smaller $\ell_2$ metrics and higher cosine similarity) from the pretrained starting point than the other priors.

\textbf{Embedding-space change.}
In the observation embedding space ($E_{\text{post}}$ vs.\ $E_{\text{pre}}$), however, the priors diverge sharply, and the CLD analysis reveals statistically significant separation at every data fraction under both $\ell_2$ (\cref{fig:convergence} bottom) and cosine similarity (\cref{fig:convergence_cosine} bottom).
The Gaussian prior shows the smallest change (CLD group ``a''), meaning the encoder barely moves from its pretrained state. The learned prior $A_{\text{pre}}$, however, induces the largest restructuring (group ``d''), roughly 2 to 2.5 times the Gaussian's $\ell_2$ change and the largest drop in cosine similarity. The mixed priors $A_{\text{pre}}{+}\sigma_A\mathcal{Z}$ and $E_{\text{pre}}{+}\sigma_E\mathcal{Z}$ fall between these extremes, each occupying a distinct CLD group under both metrics.

\textbf{Practical significance.}
The action-space distributions (\cref{fig:convergence} top) also show statistically distinct CLD groups, but their empirical distributions overlap almost entirely---the detected differences are too small to be practically meaningful.
In contrast, the embedding-space separability corresponds to large, visually apparent gaps between the violin distributions, particularly at lower data fractions, and the same prior ranking holds under both $\ell_2$ and cosine metrics.
We speculate one possible explanation for the large embedding-space changes: because the Gaussian was used during pretraining, fine-tuning with $\mathcal{Z}$ only shifts the target action distribution toward the fine-tuning task data while the prior remains unchanged. Fine-tuning with a pretrained action prior $A_{\text{pre}}$, in contrast, shifts both the prior and the target relative to the pretrained policy, forcing the encoder to compensate more. Mixed priors that combine noise with pretrained actions dilute this shift and produce intermediate encoder changes.

The prior thus governs how much the encoder restructures, but not the quality of the final result: all priors arrive at features that are equally predictive of actions (\cref{sec:logme}). This view resonates with previous loss landscape analyses~\cite{yosinski2014transferable, he2019rethinking} on vision transfer learning, where different encoder configurations could lead to equivalent downstream performance. In our case, it is likely that the prior determines which local minimum the optimizer settles into, but these local minima are comparably effective downstream. Whether this divergence in encoder representations has other implications for transfer learning, beyond the downstream success rate we measure here, remains to be explored.

\subsection{Encoder Training Matters More Than Prior Choice}
\label{sec:encoder_lr}

The previous subsection shows that with respect to the pretrained policy, action-space changes after fine-tuning are relatively small (significant overlap in the violins), whereas the observation encoder restructures substantially across priors.
This naturally raises the question: how much does encoder training quality matter relative to prior choice?
If the encoder dominates and the prior is secondary, then degrading the encoder should hurt performance far more than changing the prior.
We test this by reducing the observation encoder's learning rate to $1/10$ of the action head's, deliberately undertraining it while keeping all other hyperparameters identical.

\begin{figure}[t]
    \centering
    \includegraphics[width=\linewidth]{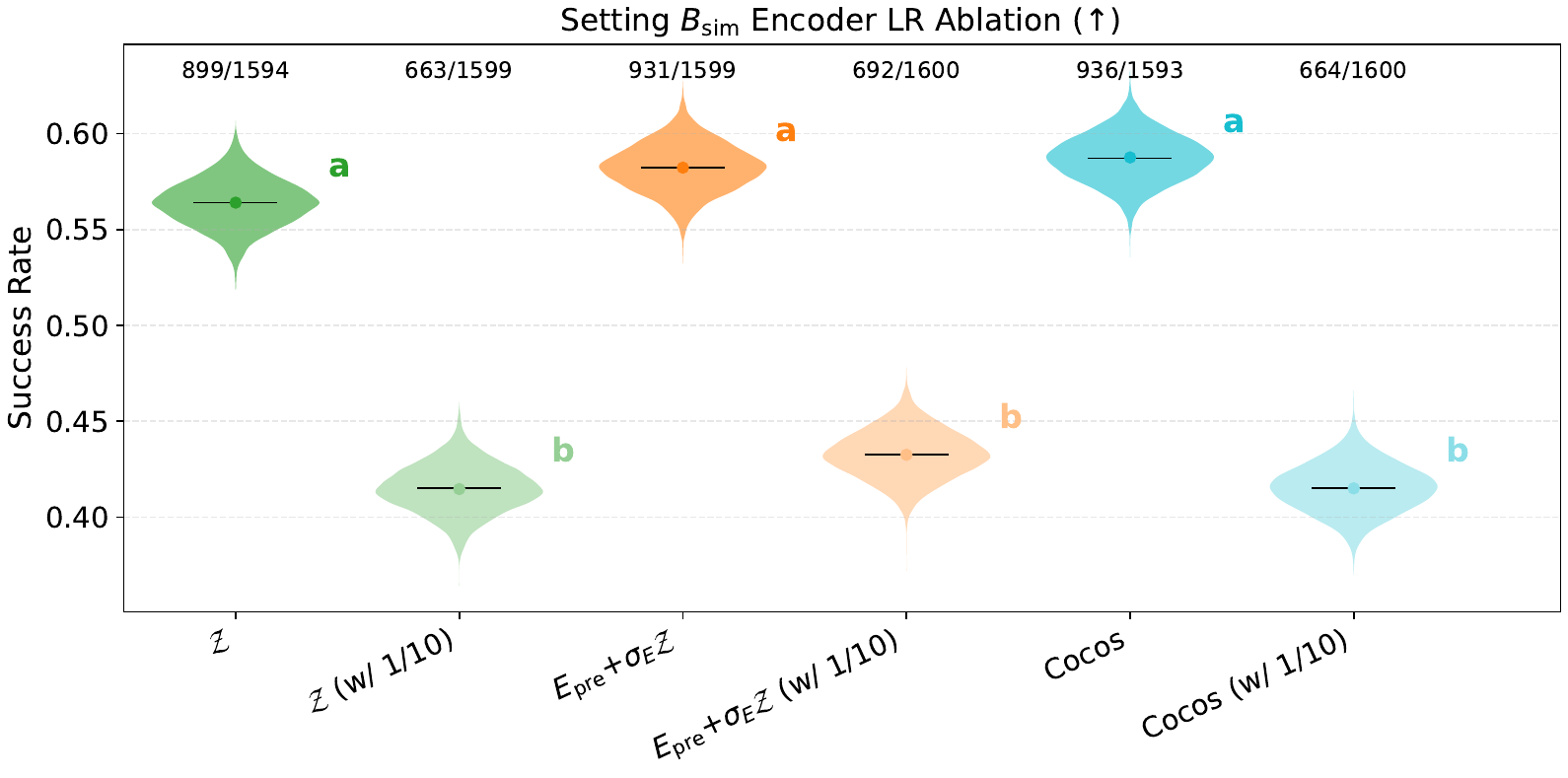}\\[6pt]
    \includegraphics[width=\linewidth]{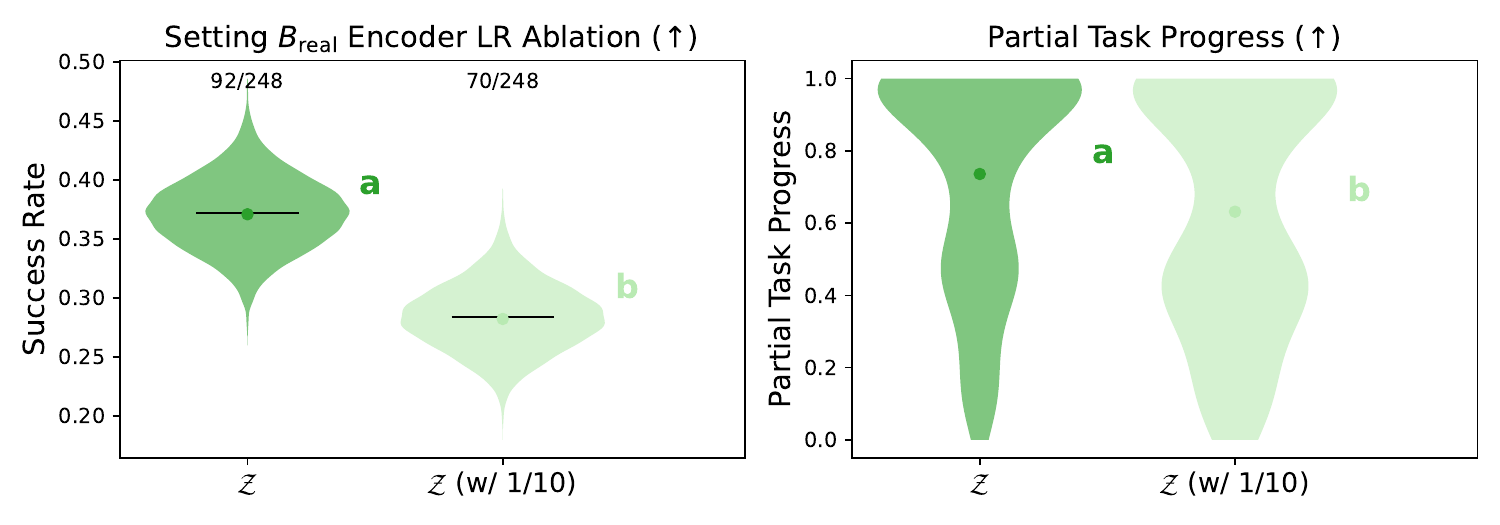}
    \caption{Encoder learning rate ablation.
    \textbf{Top:} Setting~$B_{\mathrm{sim}}$ (eight tasks, 100\% data, three representative priors). Reducing the encoder LR to $1/10$ of the action head's (lighter violins, ``b'') degrades performance by 15--17\% across all three priors, while the choice of prior has minimal effect within each LR group (both groups share a single CLD letter).
    \textbf{Bottom:} Setting~B\textsubscript{real} (Gaussian prior, five tasks). Reducing the encoder LR to $1/10$ drops both success rate and partial task progress (different CLD groups), a larger effect than any prior-related difference in this study.}
    \label{fig:encoder_lr}
\end{figure}

In simulation, we run this ablation on the LBM 1.0 model with three representative priors---$\mathcal{Z}$ (Gaussian baseline), $E_{\text{pre}}{+}\sigma_E\mathcal{Z}$ (best-performing among our proposed priors), and Cocos (most recent external baseline)---across all eight Setting~B\textsubscript{sim} tasks at 100\% data (\cref{fig:encoder_lr}).
Reducing the encoder learning rate consistently and significantly degrades performance across all three priors: success rates drop from the 56--59\% range to the 41--43\% range.
Statistical tests confirm significant degradation for all three priors (reduced-LR variants share CLD group ``b''; standard-LR variants share group ``a''). Moreover, the magnitude of degradation is comparable across priors, which is exactly the expected result if the encoder, not the prior, is the dominant factor.

On hardware (Setting~B\textsubscript{real}), we validate this finding with the Gaussian prior across five tasks (\cref{fig:encoder_lr}).
The standard encoder LR achieves 37\% success rate, while the reduced $1/10$ LR drops to 28\%, a statistically significant decrease (different CLD letters).
This hardware result confirms the simulation finding: undertraining the encoder has a far larger impact on policy quality than any prior-related change we study.

Overall, observation-encoder training quality is the dominant factor in fine-tuning performance: within each encoder-LR regime, the choice of action-space prior makes no measurable difference, while changing the encoder's learning rate shifts performance by an amount that exceeds any prior-related effect. This resonates with recent work on run-time monitoring~\cite{XuC2-RSS-25,gu2025safe}, which found that policy encoder embeddings carry rich information predictive of downstream task success. Our encoder-LR ablation supports the view that the quality of these embeddings is a critical lever for policy performance.

\section{Discussion}
\label{sec:discussion}



\subsection{Implications for Practitioners}

In practice, the standard Gaussian prior is sufficient for fine-tuning LBMs with a flow-matching action head.
Implementing learned priors adds engineering complexity with no measurable benefit.
Our encoder-LR ablation (\cref{sec:encoder_lr}) suggests that the key factor appears to be not the prior itself but the quality of the pretrained representations: once the encoder is sufficiently well-pretrained, the encoder dominates the transport regardless of where the flow starts.
On the other hand, in the from-scratch regime without pretrained models, learned priors may still be valuable, improving success rates even though differences could be statistically indistinguishable, as supported by both our results and prior work~\cite{chen2024bridger, dong2026conditioning, jia2026a2a, chang2026efficientflow}.

\subsection{Limitations}

Our findings are established on three LBMs (LBM 1.0, $\pi_{0.5}$, and GR00T~N1.5).
The prior-independence finding holds consistently across all three, suggesting it is not an artifact of a particular model capacity or architecture.
Nevertheless, we cannot rule out that much larger or smaller models, or models trained on special pretraining distributions, might behave differently. Similarly, our fine-tuning datasets are relatively small (20--400 demonstrations per task); whether the prior-independence finding persists under substantially larger or more diverse fine-tuning data remains open. Our proposed explanation, that the pretrained encoder dominates the action prior choices, is plausible at other scales but remains unverified. Although the metrics we use (LogME, per-observation $\ell_2$, and cosine similarity) are widely adopted in the transfer learning literature~\cite{you2021logme, yosinski2014transferable}, alternative metrics such as LEEP~\cite{nguyen2020leep} or H-Score~\cite{bao2019hscore} could reveal additional aspects of how priors affect LBM fine-tuning that we have not explored. 

\pagebreak

\subsection{Future Work}

We point out three concrete research directions:
\begin{enumerate}
    \item \textbf{Where is the transition?} At 5\% data in Setting~B\textsubscript{sim} (approximately 20 demonstrations per task), $E_{\text{pre}}{+}\sigma_E\mathcal{Z}$ shows significant improvements over $\mathcal{Z}$. This hints at a regime where the prior begins to matter, but the boundary is not clearly characterized. Mapping this transition as a function of dataset size, task complexity, and model capacity would let practitioners decide when learned priors are worth the engineering cost.
    \item \textbf{Why do different priors yield divergent encoder representations of equal quality?} Our analysis shows that they do, but not why the loss landscape admits multiple equally good encoder configurations. Loss-landscape geometry, plasticity~\cite{lyle2023plasticity}, and linear mode connectivity~\cite{frankle2020linear} are all plausible directions worth investigating.
    \item \textbf{Does encoder divergence have downstream consequences?} Different priors lead to structurally different encoders with indistinguishable nominal policy performance. Whether these solutions differ in robustness to distribution shift, continual-learning stability, or catastrophic forgetting is an open question with practical~implications.
\end{enumerate}

\section{Conclusion}
\label{sec:conclusion}

We presented a large-scale empirical study of non-Gaussian priors under LBM fine-tuning for visuomotor imitation learning.
Across 100K+ cross-architecture simulation rollouts on 40+ tasks in two simulation platforms, and 1250 hardware rollouts, the evidence consistently points in the same direction: learned priors that help when training from scratch do not improve fine-tuning performance.
Diagnostic analyses reveal that regardless of prior choice, fine-tuned encoders yield indistinguishable conditional action likelihoods, and the resulting policy outputs similar actions. However, their encoder representations diverge from the pretrained model and each other. An encoder-LR ablation further confirms that the observation encoder, not the prior, is the dominant factor.
These findings corroborate that the standard Gaussian prior used in current LBM fine-tuning pipelines is a well-supported choice. 
More broadly, given these negative results, we hope to inspire the community to explore the underlying learning dynamics of large-scale visuomotor architectures, focusing on why certain components (such as the observation encoder) matter more than others and how that understanding can improve different aspects of downstream robot policy~performance.


\section*{Acknowledgment}
The authors thank Paarth Shah, Eric Cousineau, Aykut Onol, and Katherine Liu for helpful discussions and feedback, and Patrick Tree Miller and all the robot teachers (Rudy Bravo, Emma Dixon, Fredy Navas, Christopher Rodriguez, Derick Seale) for assistance with the hardware experiments.


\bibliographystyle{IEEEtran}
\bibliography{references}


\clearpage
\appendices
\crefalias{section}{appendix}
\crefalias{subsection}{appendix}
\crefname{appendix}{Appendix}{Appendices}
\Crefname{appendix}{Appendix}{Appendices}

\section{Prior Formulations}
\label{sec:appendix_priors}

This appendix specifies, for each of the seven prior variants in \cref{tab:priors}, the exact sampling procedure used to draw the prior sample $a_0$ that the flow-matching velocity field transports to the target action during training and inference. A common convention applies to denoising budgets: if invoking $\pi_{\text{pre}}$ to denoise from a Gaussian to a clean action would take $N$ steps under its pretraining objective (DDPM for LBM 1.0, flow matching for $\pi_{0.5}$ and GR00T~N1.5), then priors that invoke $\pi_{\text{pre}}$ to construct $a_0$ ($A_{\text{pre}}$, $A_{\text{pre}}{+}\sigma_A\mathcal{Z}$, $E_{\text{pre}}{+}\sigma_E\mathcal{Z}$) spend $N/2$ steps on prior construction and $N/2$ on the subsequent flow-matching transport. Priors that do not invoke $\pi_{\text{pre}}$ during sampling ($\mathcal{Z}$, Cocos, BRIDGER, Retrieval) use the full $N$ steps for transport. Concretely, $N{=}8$ for LBM 1.0, $N{=}10$ for $\pi_{0.5}$, and $N{=}4$ for GR00T~N1.5. We select $\sigma_A{=}1$ and $\sigma_E{=}0.25$ for the main experiments; see \cref{sec:appendix_noise_ablation} for the supporting grid search.

Following \cref{sec:problem}, $o$ denotes the conditioning observation and $e \in \mathbb{R}^{d_e}$ the observation embedding passed to the pretrained action head. For LBM 1.0, $e$ is the concatenated output of the CLIP ViT-B/16 vision encoder applied to each camera over the two-frame history ($d_e{=}6732$). For $\pi_{0.5}$, $e$ is the concatenation of the SigLIP visual tokens across cameras. For GR00T~N1.5, $e$ is the 1536-dimensional EAGLE-VLM embedding (Qwen3-1.7B LLM + SigLIP-2 vision encoder with a learned projection) conditioning the DiT action head. $H$ denotes the action-horizon length and $d_a$ the per-timestep action dimensionality, so each action sample lies in $\mathbb{R}^{H\times d_a}$: $H{=}16$, $d_a{=}20$ for LBM 1.0; $H{=}50$, $d_a{=}12$ for $\pi_{0.5}$ on RoboCasa; and $H{=}16$, $d_a{=}12$ (effective; padded to a 32-dimensional action slot) for GR00T~N1.5 on RoboCasa.

\paragraph{Gaussian, $\mathcal{Z}$.}
\begin{equation}
a_0 \sim \mathcal{N}(0, I), \quad a_0 \in \mathbb{R}^{H\times d_a}.
\end{equation}
No auxiliary training.

\paragraph{Pretrained-model actions, $A_{\text{pre}}$.}
$a_0$ is produced by denoising a Gaussian sample through a frozen copy of the pretrained policy for $N/2$ steps,
\begin{equation}
a_0 \;=\; \pi_{\text{pre}}\bigl(z \mid e;\, N/2\bigr), \qquad z\sim\mathcal{N}(0,I),
\end{equation}
where $\pi_{\text{pre}}(\cdot \mid e; K)$ denotes running the pretrained denoiser from $z$ for $K$ steps conditioned on $e$, under its pretraining objective (DDPM for LBM 1.0, flow matching for $\pi_{0.5}$ and GR00T~N1.5). No auxiliary training.

\paragraph{Perturbed pretrained-model actions, $A_{\text{pre}}{+}\sigma_A\mathcal{Z}$.}
\begin{equation}
a_0 \;=\; \pi_{\text{pre}}\bigl(z \mid e;\, N/2\bigr) + \sigma_A\,\tilde z, \qquad z, \tilde z \sim \mathcal{N}(0,I),
\end{equation}
with $\sigma_A{=}1$ (selected from the noise-scale ablation of \cref{fig:noise_ablation}). No auxiliary training.

\paragraph{Perturbed-embedding prior, $E_{\text{pre}}{+}\sigma_E\mathcal{Z}$.}
Noise is injected into the observation embedding before $\pi_{\text{pre}}$ denoises,
\begin{equation}
\tilde e = e + \sigma_E\,\xi,\ \ \xi\sim\mathcal{N}(0,I); \quad
a_0 = \pi_{\text{pre}}\bigl(z \mid \tilde e;\, N/2\bigr),
\end{equation}
with $\sigma_E{=}0.25$ (selected from \cref{fig:noise_ablation}). No auxiliary training.

\paragraph{Cocos~\cite{dong2026conditioning}.}
An auxiliary encoder--decoder $F_\phi$ is trained to reconstruct $e$ through an action-space bottleneck of dimension $H\cdot d_a$; its bottleneck activation is then used as the mean of a Gaussian prior in the action space,
\begin{equation}
a_0 \sim \mathcal{N}\!\bigl(\alpha\, F_\phi(e),\ \beta^2 I\bigr), \quad \alpha{=}1, \ \beta{=}1.
\end{equation}
$F_\phi$ is a single-layer Transformer autoencoder (hidden dimension 384, 6 attention heads) whose decoder output matches $e$. Before policy fine-tuning, we collect 1000 steps of $(e, a)$ pairs under the frozen pretrained policy (2000 for Setting~B\textsubscript{real}), then train $F_\phi$ for 100 epochs with a negative cosine-similarity reconstruction loss (Adam, learning rate $3{\times}10^{-4}$, batch size 256). $F_\phi$ is frozen thereafter. The original Cocos paper uses $\beta{=}0.2$; that value collapsed the prior too tightly around its mean to remain useful under policy fine-tuning, so we report $\beta{=}1$, which was reliable in preliminary sweeps. Additionally, we noted that the original paper kept the policy encoder frozen during training, but the frozen encoder led to ${\sim}0\%$ success rate on the LBM Eval benchmark. We hence keep the policy encoder trainable when training the policy action head, and note that this is a usage difference from the original paper.

\paragraph{BRIDGER~\cite{chen2024bridger}.}
A CVAE maps a Gaussian latent to an action prior, conditioned on the observation embedding:
\begin{equation}
z \sim \mathcal{N}(0, I), \quad a_0 \;=\; D_\psi(z, e) + \sigma\,\eta,\ \ \eta\sim\mathcal{N}(0,I),
\end{equation}
with $\sigma{=}0.5$ and latent dimension 64. Encoder and decoder are 3-layer MLPs (hidden dimension 512, ELU activations). Auxiliary training mirrors Cocos: 1000 frozen-policy collection steps (2000 for Setting~B\textsubscript{real}), followed by 200 epochs of CVAE training (Adam, learning rate $10^{-3}$, batch size 256, ELBO with reconstruction MSE and KL weight annealed linearly from 0 to $\beta_{\text{KL}}{=}0.5$). The CVAE is frozen thereafter. The original BRIDGER paper reports $\sigma{=}0$ and $\beta_{\text{KL}}{=}0.01$; in our fine-tuning regime those values collapsed the decoder onto a narrow slice of the demonstration manifold, so we report the values above, identified in preliminary sweeps. BRIDGER's CVAE training was also unstable on the Setting~B\textsubscript{real} datasets: NaN losses required restarting with a different random seed to obtain a usable prior.

\paragraph{Retrieval, $k$-NN.}
Before policy fine-tuning, we collect 1000 steps of $(e_i, a_i)$ pairs under the frozen pretrained policy (2000 for Setting~B\textsubscript{real}) and store them as a retrieval bank. At every training and inference step, given a query embedding $e$, we compute
\begin{equation}
\mathcal{I}_k(e) \;=\; \operatorname*{arg\,top}\text{-}k\ \bigl\{-\lVert e - e_i\rVert_2\bigr\}, \quad
a_0 \;=\; \frac{1}{k}\sum_{i\in\mathcal{I}_k(e)} a_i + \sigma\,\eta,
\end{equation}
with $k{=}3$, $\sigma{=}0.5$, and $\eta\sim\mathcal{N}(0,I)$. The bank is built once at the end of warm-up and then frozen; distances are in the same embedding space $e$ that conditions the pretrained action head.

\section{Statistical Testing Details}
\label{sec:appendix_stats}

This appendix expands on the condensed stats description in \cref{sec:experiments}.

\textbf{STEP for binary outcomes.} Sequential Testing for Efficient Policy Comparison (STEP)~\cite{snyder2025step} provides strict error control with adaptive stopping for success/failure comparisons. Because STEP's false-positive rate is guaranteed to stay below a specific value (e.g., 0.05) at \emph{any} stopping time (``anytime-valid'' error control), we can monitor the test as rollouts accumulate and stop as soon as the result is decisive, without inflating the false-positive rate.

\textbf{Lai fallback at large $n$.} At per-method sample sizes above ${\sim}600$, STEP's per-comparison computation becomes heavy; we fall back to the Lai sequential test~\cite{lai1988sequential}, which is near-optimal asymptotically for the same hypothesis class. The fallback is used for the per-task CLD plots in \cref{sec:appendix_per_task} where $n$ accumulates across tasks.

\textbf{Welch's $t$-test for continuous outcomes.} For continuous metrics (per-observation $\ell_2$ distance and cosine similarity in \cref{sec:analysis}) we use Welch's $t$-test~\cite{welch1947ttest}. The per-method sample sizes in our experiments (typically ${\geq}10000$) allow tight central-limit-theorem approximation for the sample mean.

\textbf{Compact letter display (CLD)~\cite{piepho2004cld}.} CLD assigns one or more letters to each method such that two methods \emph{share at least one letter if and only if} their pairwise difference is not significant at the corrected level. Letters are typically ordered by mean performance (earlier letters = better) unless false negatives occur between some pairs. For example, a four-method CLD of ``a,'' ``ab,'' ``b,'' ``c'' means the first and third differ significantly, the second is indistinguishable from the first and third, and the fourth is strictly worse than all others.

\section{Implementation Details}
\label{sec:appendix_implementation}

All fine-tuning runs use a single AWS \texttt{p5.48xlarge} node with up to 8${\times}$NVIDIA H100 80\,GB GPUs.

\subsection{Fine-Tuning Hyperparameters: LBM 1.0 Backbone}
\label{sec:appendix_hyperparams_lbm}

Settings A, B\textsubscript{sim}, B\textsubscript{real}, and their ablations (B\textsubscript{FP}, A$'$, noise $\sigma$, encoder LR) all fine-tune the same pretrained LBM 1.0 policy~\cite{tri2026lbm} (${\sim}560$M parameters: CLIP ViT-B/16 vision encoder, DiT action head with depth 10, embed dim 768, 12 heads). \cref{tab:appendix_hparams_lbm} lists the hyperparameters held identical across all prior variants; per the controlled comparison protocol of \cref{sec:experiments}, only the prior differs.

\begin{table*}[!htbp]
\centering
\caption{Fine-tuning hyperparameters for the LBM 1.0 backbone (Settings A, B\textsubscript{sim}, B\textsubscript{real}). Values in parentheses apply to the corresponding ablations. The fine-tuning step budget in Setting~B\textsubscript{sim} follows a discrete lookup by data fraction.}
\label{tab:appendix_hparams_lbm}
\renewcommand{\arraystretch}{1.15}
\begin{tabular}{@{}lp{11cm}@{}}
\toprule
\textbf{Hyperparameter} & \textbf{Value} \\
\midrule
Optimizer & AdamW ($\beta_1{=}0.95$, $\beta_2{=}0.999$, $\epsilon{=}10^{-8}$) \\
Weight decay & $10^{-6}$ \\
Peak learning rate & $2\times 10^{-5}$ \\
LR schedule & Cosine decay with 50-step linear warmup \\
Trainable parameters & Full model (CLIP ViT-B/16 encoder + DiT action head), with encoder LR multiplier $1.0$ ($0.1$ in the encoder-LR ablation) \\
Global batch size & 304 (38 per GPU $\times$ 8 GPUs) \\
Gradient accumulation & 1 \\
EMA & enabled (inverse-gamma $1.0$, power $0.75$, max decay $0.9999$) \\
Action horizon $H$ & 16 (first 8 executed before replanning) \\
Observation history & 2 frames \\
Fine-tuning steps (Setting B\textsubscript{sim}, 100\%) & 3000 \\
Fine-tuning steps (Setting B\textsubscript{sim}, 5/25/50/75\%) & 500 / 1000 / 1500 / 2500 \\
Fine-tuning steps (Setting A, stage 1 + 2) & 1500 (on 50\% data) + 3000 (on 100\% data) \\
Fine-tuning steps (Setting B\textsubscript{real}) & 30000 \\
Denoising steps at inference & 8 total (4 via frozen copy for $A_{\text{pre}}$ construction, 4 via the fine-tuned action expert) \\
Aux-model data collection (Cocos, BRIDGER, Retrieval) & 1000 steps of frozen-policy $(e, a)$ collection (2000 for Setting B\textsubscript{real}) \\
\bottomrule
\end{tabular}
\end{table*}

The Gaussian baseline $\mathcal{Z}$ keeps LBM 1.0's DDPM pretraining objective; all non-Gaussian priors use flow matching. Setting B\textsubscript{FP} (\cref{fig:setting_bfp}) re-runs the Gaussian under flow matching to confirm that this objective choice does not confound the comparison.

\subsection{Fine-Tuning Hyperparameters:
  \texorpdfstring{$\pi_{0.5}$}{pi\_0.5} Backbone}
\label{sec:appendix_hyperparams_pi05}

Setting C fine-tunes the public $\pi_{0.5}$ checkpoint~\cite{black2025pi05} (${\sim}3$B parameters: SigLIP vision encoder, Gemma 2B VLM, Gemma 300M action expert). Hyperparameters follow the official \texttt{openpi} release and are reproduced in \cref{tab:appendix_hparams_pi05} for completeness.

\begin{table*}[!htbp]
\centering
\caption{Fine-tuning hyperparameters for the $\pi_{0.5}$ backbone (Setting C).}
\label{tab:appendix_hparams_pi05}
\renewcommand{\arraystretch}{1.15}
\begin{tabular}{@{}lp{11cm}@{}}
\toprule
\textbf{Hyperparameter} & \textbf{Value} \\
\midrule
Optimizer & AdamW ($\beta_1{=}0.9$, $\beta_2{=}0.95$, $\epsilon{=}10^{-8}$) \\
Weight decay & $10^{-10}$ \\
Peak learning rate & $2.5\times 10^{-5}$ \\
LR schedule & Cosine decay to $2.5\times 10^{-6}$ with 100-step linear warmup \\
Gradient clipping & $\lVert g\rVert_2 \le 1.0$ \\
Trainable parameters & Full model (SigLIP vision encoder + Gemma 2B VLM + Gemma 300M action expert) \\
Global batch size & 64 (8 per GPU $\times$ 8 GPUs) \\
Gradient accumulation & 1 \\
Precision & bfloat16 \\
Action horizon $H$ & 50 \\
Observation history & 1 frame \\
Fine-tuning steps & 4000 \\
Denoising steps at inference & 10 total (5 via frozen copy for $A_{\text{pre}}$ construction, 5 via the fine-tuned action expert) \\
\bottomrule
\end{tabular}
\end{table*}

\subsection{Fine-Tuning Hyperparameters: GR00T~N1.5 Backbone}
\label{sec:appendix_hyperparams_gr00t}

Setting C's GR00T~N1.5 runs fine-tune the public GR00T~N1.5~3B checkpoint~\cite{gr00tn1_2025} (EAGLE vision-language backbone with Qwen3-1.7B LLM and SigLIP-2 vision encoder, flow-matching DiT action head; ${\sim}3.7$B total parameters). Following the official GR00T~N1.5 release, we freeze the EAGLE backbone and update only the action-head projector and the DiT; \cref{tab:appendix_hparams_gr00t} lists the hyperparameters.

\begin{table*}[!htbp]
\centering
\caption{Fine-tuning hyperparameters for the GR00T~N1.5 backbone (Setting C).}
\label{tab:appendix_hparams_gr00t}
\renewcommand{\arraystretch}{1.15}
\begin{tabular}{@{}lp{11cm}@{}}
\toprule
\textbf{Hyperparameter} & \textbf{Value} \\
\midrule
Optimizer & AdamW ($\beta_1{=}0.95$, $\beta_2{=}0.999$, $\epsilon{=}10^{-8}$) \\
Weight decay & $10^{-5}$ \\
Peak learning rate & $3\times 10^{-5}$ \\
LR schedule & Cosine decay with 200-step linear warmup (5\% of 4000 steps) \\
Gradient clipping & $\lVert g\rVert_2 \le 1.0$ \\
Trainable parameters & Action-head projector + DiT (EAGLE backbone frozen) \\
Global batch size & 128 (64 per GPU $\times$ 2 GPUs) \\
Gradient accumulation & 1 \\
Precision & bfloat16 (with tf32 matmul accumulators) \\
Action horizon $H$ & 16 \\
Observation history & 1 frame \\
Fine-tuning steps & 4000 \\
Denoising steps at inference & 4 total (2 via frozen copy for $A_{\text{pre}}$ construction, 2 via the fine-tuned action expert) \\
\bottomrule
\end{tabular}
\end{table*}

\subsection{Tasks and Demonstration Counts}
\label{sec:appendix_tasks}

\cref{tab:appendix_tasks_sim} lists the eight LBM Eval tasks used in Settings~A and B\textsubscript{sim}, with full names as in the LBM 1.0 paper~\cite{tri2026lbm}. Setting~A uses only the five-task Kitchen-K subset. All eight tasks were unseen during LBM 1.0 pretraining: the five Kitchen-K tasks come from a new scenario, while the three Breakfast/Shelf/DryingRack tasks are unseen-variant tasks whose scenarios (but not the tasks themselves) appeared during pretraining. The three unseen-variant tasks form the ``easier'' group of \cref{fig:setting_b_seen_unseen}, and the five Kitchen-K tasks the ``harder'' group.

\begin{table*}[!htbp]
\centering
\caption{Simulation tasks used in Settings A and B\textsubscript{sim} (LBM Eval benchmark). Demonstration counts follow LBM 1.0~\cite{tri2026lbm}. Fractional data experiments in Setting B\textsubscript{sim} randomly subsample this full pool per task without replacement (e.g., 50\% of \textit{DumpVeg} uses 196 randomly selected demonstrations out of 392).}
\label{tab:appendix_tasks_sim}
\renewcommand{\arraystretch}{1.15}
\setlength{\tabcolsep}{8pt}
\begin{tabular}{@{}lllr@{}}
\toprule
\textbf{Short name} & \textbf{Full name} & \textbf{Scenario} & \textbf{\#\,demos} \\
\midrule
\multicolumn{4}{@{}l}{\textit{Easier group (three tasks, used in Setting B\textsubscript{sim} only)}} \\[2pt]
PutMug & PutMugInCenterOfTable & Breakfast (B) & 294 \\
PlaceAvocado & PlaceAvocadoFromBowlIntoBin & Shelf (S) & 196 \\
PutSpatula & PutSpatulaOnPlateFromUtensilCrock & DryingRack (D) & 196 \\
\midrule
\multicolumn{4}{@{}l}{\textit{Harder group (five tasks, used in Settings A and B)}} \\[2pt]
DumpVeg & DumpVegetablesFromSmallToLargeContainer & Kitchen (K) & 392 \\
PutContainer & PutContainersOnPlate & Kitchen (K) & 392 \\
PutFruit & PutFruitInLargeContainerAndCoverWithPlate & Kitchen (K) & 392 \\
SeparateFruit & SeparateFruitsVegetablesIntoContainers & Kitchen (K) & 392 \\
TurnContainer & TurnLargeContainerUpsideDown & Kitchen (K) & 392 \\
\bottomrule
\end{tabular}
\end{table*}

\cref{tab:appendix_tasks_hw} lists the five hardware tasks used in Setting~B\textsubscript{real}. \textit{BikeRotorInstall} and \textit{ClearKitchenCounter} are adapted from LBM 1.0~\cite{tri2026lbm}; the remaining three (\textit{FoodBank}, \textit{GatheringIngredientsForCookies}, \textit{BusBin}) are new long-horizon bimanual tasks (kitchen preparation and sorting) introduced in this work.

\begin{table*}[!htbp]
\centering
\caption{Hardware tasks used in Setting B\textsubscript{real} (bimanual Franka FR3).}
\label{tab:appendix_tasks_hw}
\renewcommand{\arraystretch}{1.15}
\setlength{\tabcolsep}{8pt}
\begin{tabular}{@{}lp{11cm}r@{}}
\toprule
\textbf{Task name} & \textbf{One-line description} & \textbf{\#\,demos} \\
\midrule
FoodBank & Pack the food box with at least one of each type of food item on the table & 354 \\
GatheringIngredientsForCookies & Gather the ingredients for making cookies & 199 \\
BusBin & Place plates and silverware into the separate bus bin, and place compost and recycling into the correct bins & 202 \\
ClearKitchenCounter~\cite{tri2026lbm} & Place tools into a tray, clean a cutting board with a sponge, and sweep the remaining waste into a trash bin & 305 \\
BikeRotorInstall~\cite{tri2026lbm} & Seat a rotor on a bicycle wheel, place a lockring, and tighten it at least one revolution with a tool & 533 \\
\bottomrule
\end{tabular}
\end{table*}

For Setting~C, we use all 34 RoboCasa~\cite{nasiriany2026robocasa} tasks (18 atomic, 16 composite) as released; demonstration counts and per-task descriptions follow the RoboCasa paper and we do not reproduce them here.

\section{Additional Experimental Results}
\label{sec:appendix_results}

\subsection{Setting B\texorpdfstring{$_{\text{sim}}$}{sim}: Easier vs.\ Harder Task Breakdown}

\begin{figure}[!htbp]
    \centering
    \includegraphics[width=\linewidth]{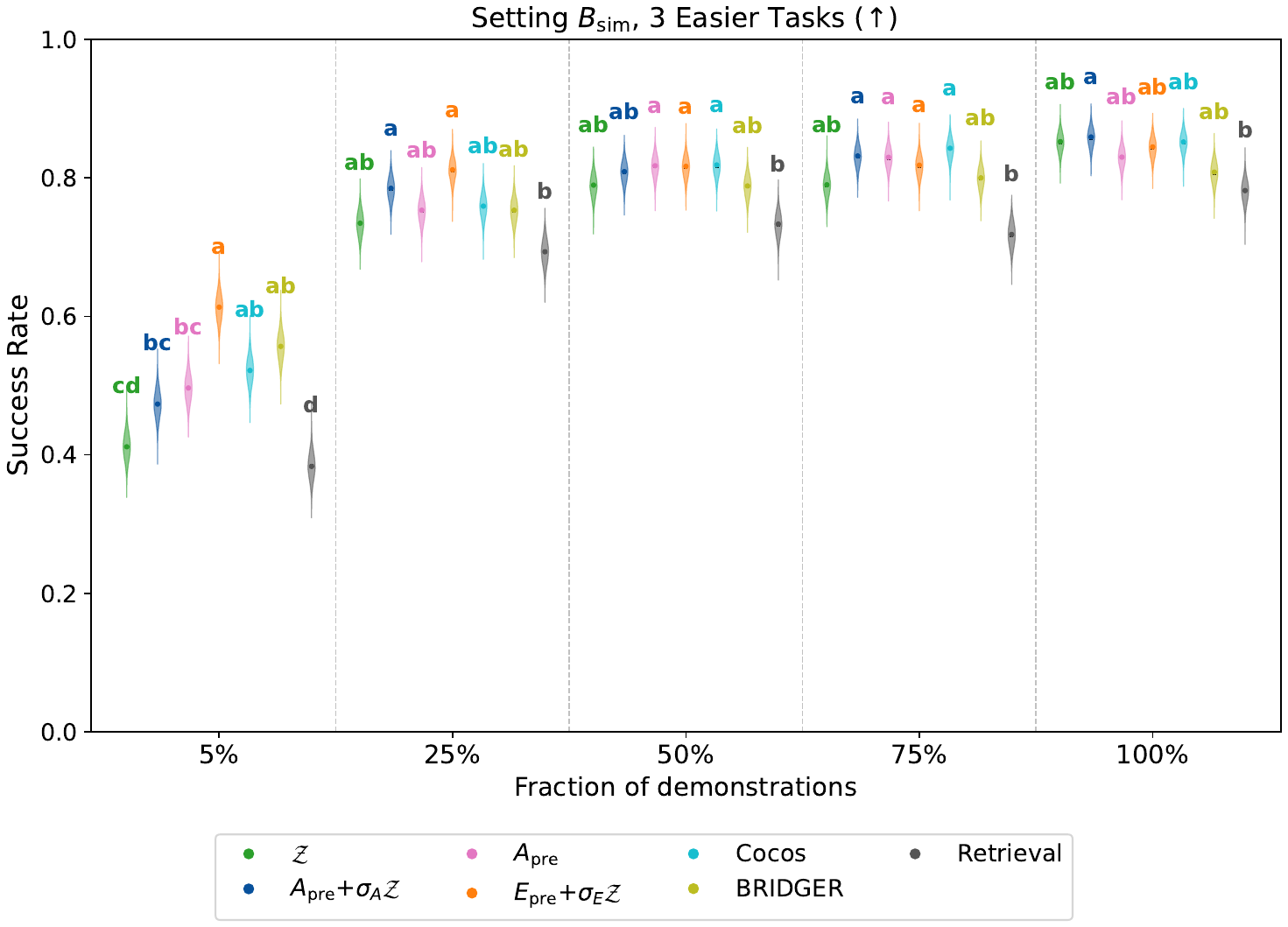}\\[4pt]
    \includegraphics[width=\linewidth]{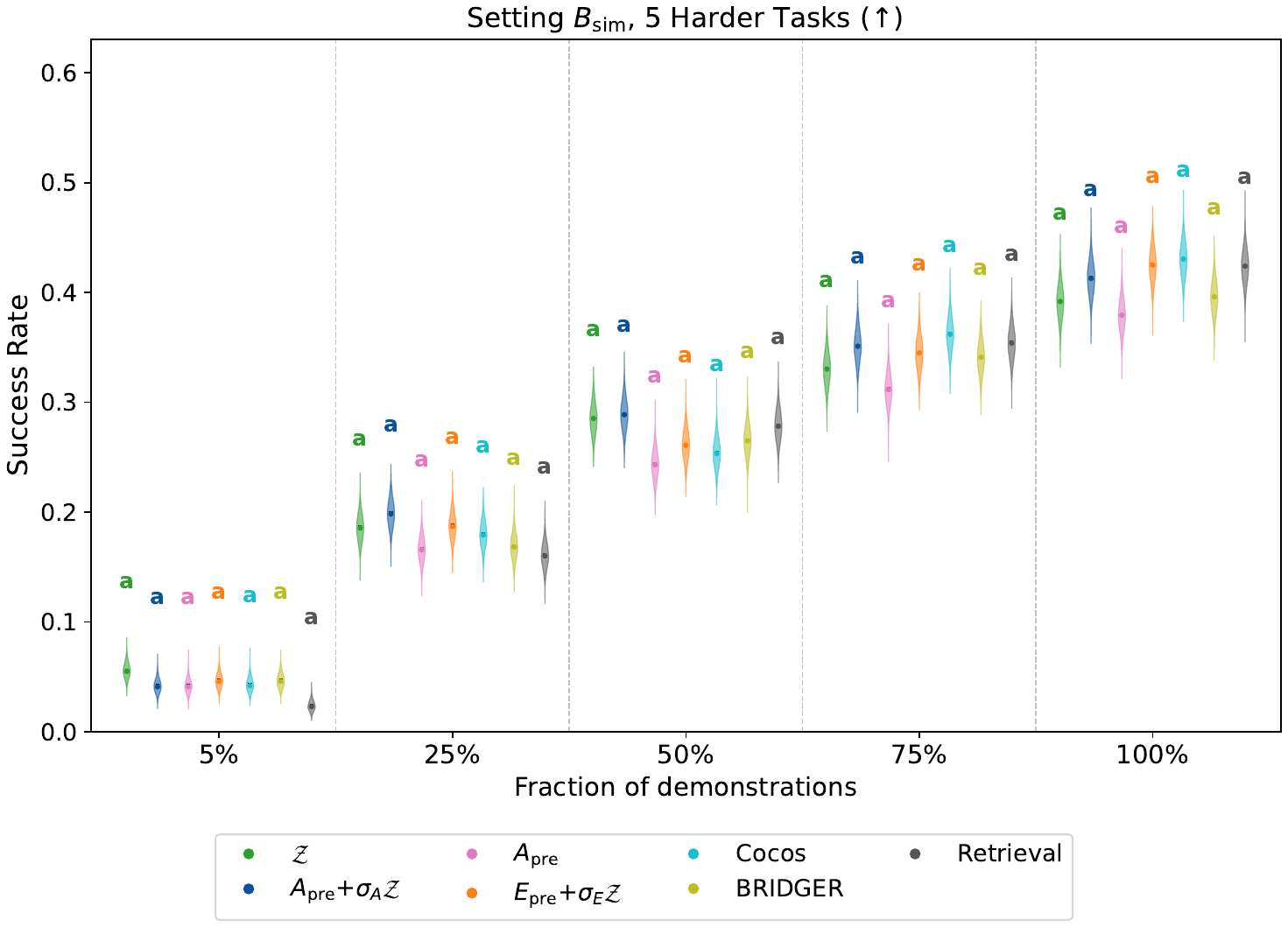}
    \caption{Setting B\textsubscript{sim}: easier (top, three tasks) vs.\ harder (bottom, five tasks) breakdown.}
    \label{fig:setting_b_seen_unseen}
\end{figure}

The eight tasks in Setting B\textsubscript{sim} were all unseen during LBM 1.0 pretraining but vary in difficulty. \cref{fig:setting_b_seen_unseen} splits them into the three easier tasks (higher baseline success rates) and the five harder Kitchen-K tasks; the prior-independence finding holds within each group.

\subsection{Raw LogME Scores}
\label{sec:appendix_logme_raw}

\begin{figure}[!htbp]
    \centering
    \includegraphics[width=\linewidth]{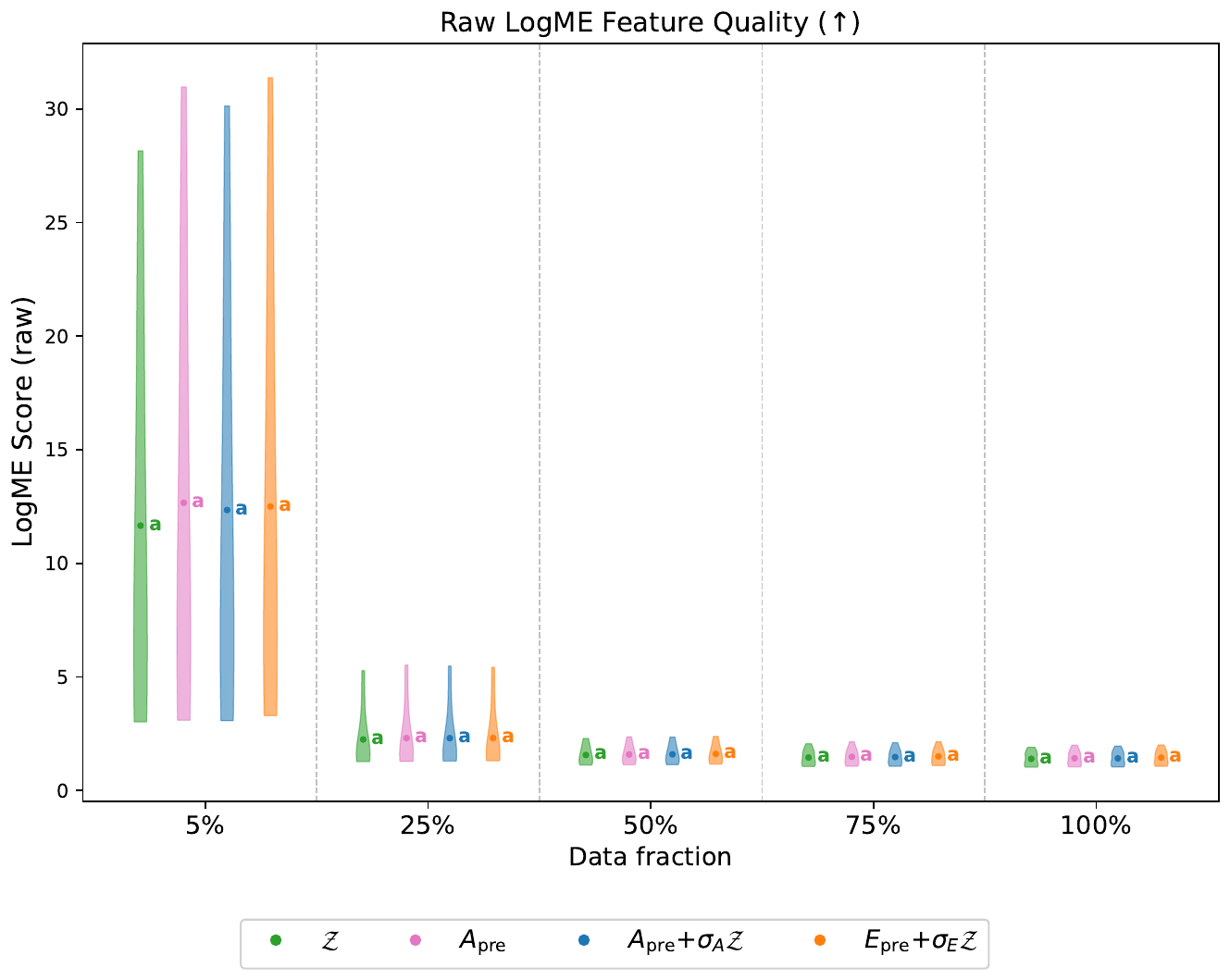}
    \caption{Raw LogME scores across data fractions (Setting B\textsubscript{sim}, eight tasks). Each violin shows the distribution of scores across tasks. The sharp drop from 5\% to 25\% is a small-sample artifact of the LogME computation, not necessarily a real change in feature quality. All priors share CLD group ``a'' at every fraction.}
    \label{fig:logme_raw}
\end{figure}

\cref{fig:logme_raw} shows the raw (unnormalized) LogME scores across data fractions. The sharp drop from 5\% to 25\% data is a small-sample artifact of the LogME computation (not necessarily a real change in feature quality). CLD letters are identical to those in \cref{fig:logme} since both are computed from the same raw values.

\subsection{Setting \texorpdfstring{B\textsubscript{FP}}{B\_FP}: Flow-Matching-Pretrained Baseline}

\begin{figure}[!htbp]
    \centering
    \includegraphics[width=\linewidth]{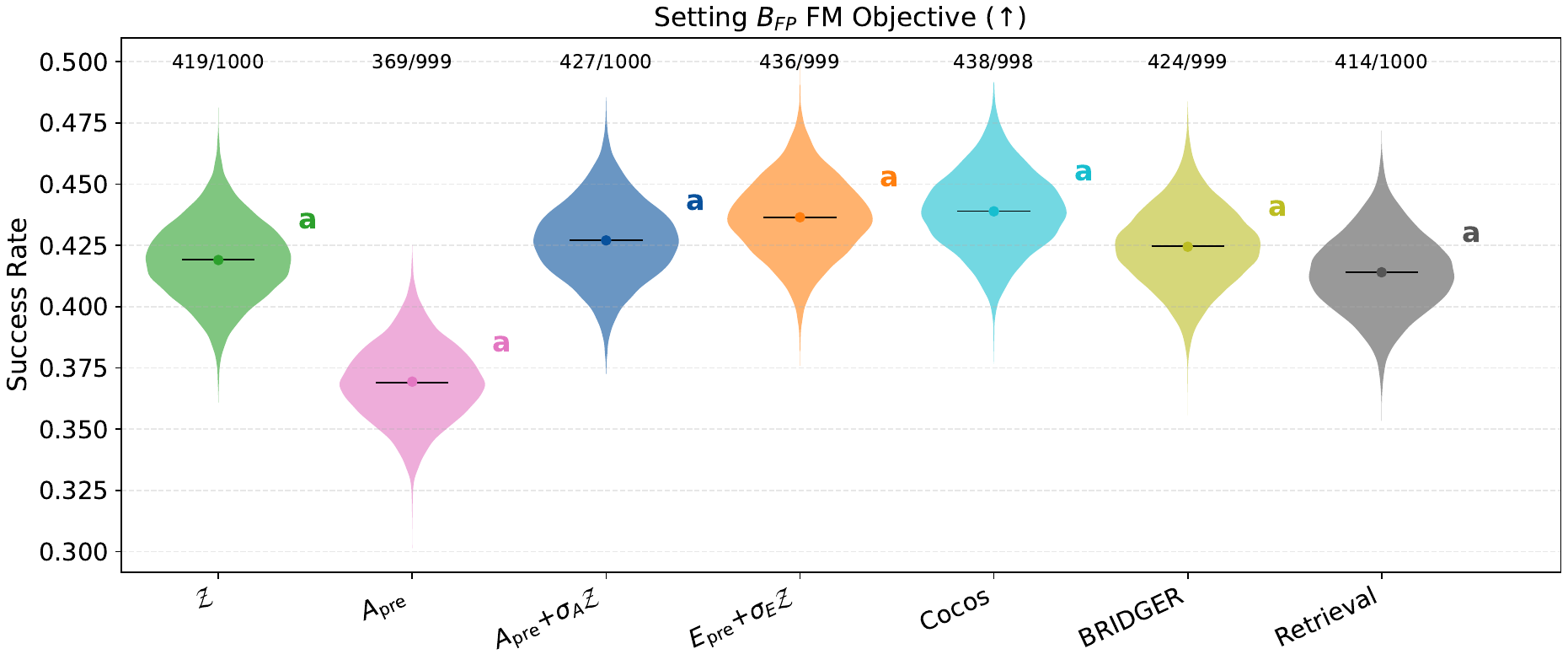}
    \caption{Setting B\textsubscript{FP}: all methods, including Gaussian, use the flow-matching objective in both pretraining and fine-tuning (five tasks).}
    \label{fig:setting_bfp}
\end{figure}

In the main experiments, the Gaussian baseline keeps LBM 1.0's DDPM pretraining objective while the non-Gaussian priors use flow matching (\cref{sec:experiments}). To rule out that this objective mismatch confounds our results, Setting~B\textsubscript{FP} retrains all seven methods---including the Gaussian---under flow matching end-to-end on the five Kitchen-K tasks at 100\% data. \cref{fig:setting_bfp} shows that the conclusion is unchanged: no prior significantly outperforms the others. The three diagnostic analyses of \cref{sec:analysis} carry over to this regime and are reported in \cref{sec:appendix_bfp_diagnostics} (\crefrange{fig:bfp_logme}{fig:bfp_repr_change_cosine}).

\subsection{Setting \texorpdfstring{B\textsubscript{FP}}{B\_FP}: Diagnostic Analyses}
\label{sec:appendix_bfp_diagnostics}

We re-run the three analyses of \cref{sec:analysis} (LogME feature quality, action-space similarity, and representation divergence) on Setting~B\textsubscript{FP} rollouts, where every method uses flow matching end-to-end, to confirm the diagnostic picture is not driven by the objective mismatch in Setting~B. Only the 100\% data fraction is available, so each prior contributes a single per-observation distribution per metric; violins are therefore laid out horizontally along the prior axis and the two representation-change panels shown side by side. Distance conventions follow \cref{sec:analysis}: un-normalized action coordinates for action-space metrics, the observation embedding $e$ for embedding-space metrics.

\begin{figure}[!htbp]
    \centering
    \includegraphics[width=0.78\linewidth]{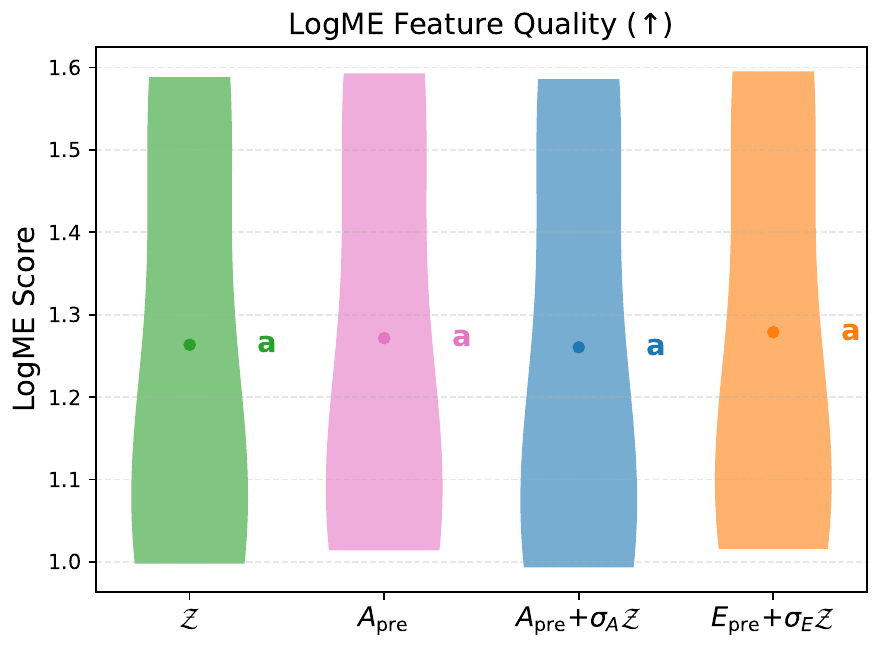}
    \caption{LogME feature quality in Setting B\textsubscript{FP} (five tasks, 100\% data). Each violin shows the distribution of scores across tasks. All four priors are indistinguishable.}
    \label{fig:bfp_logme}
\end{figure}

\textbf{Feature quality (LogME).}
\cref{fig:bfp_logme} reports LogME averaged across the five tasks. All four priors share CLD group ``a'' with means within a ${\sim}0.02$ band, matching \cref{sec:logme}: the fine-tuned encoder is equally predictive of target actions regardless of prior.

\begin{figure}[!htbp]
    \centering
    \includegraphics[width=\linewidth]{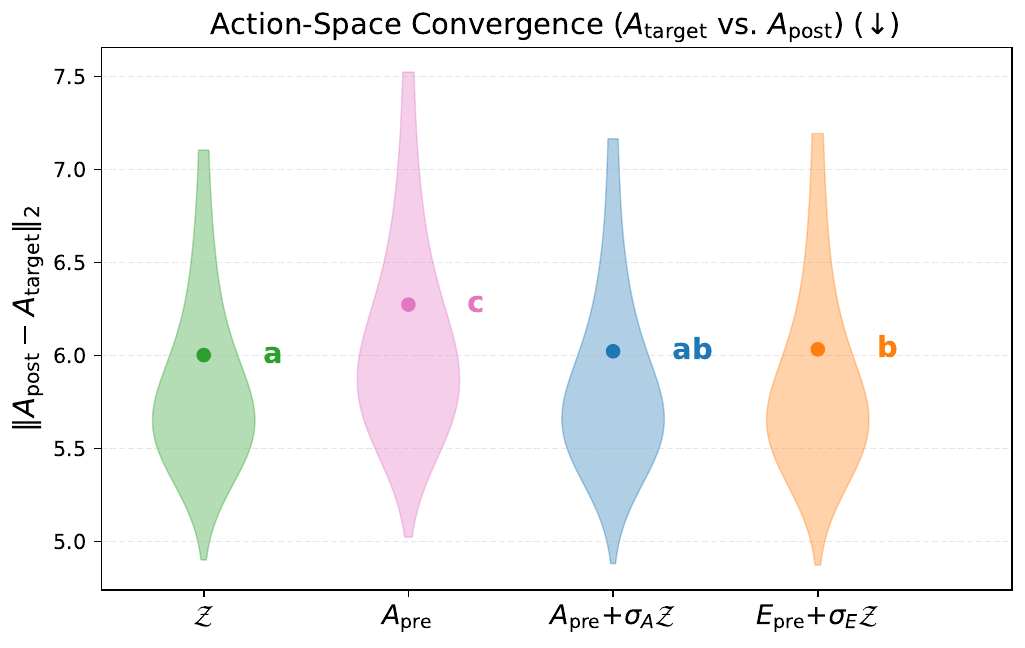}\\[4pt]
    \includegraphics[width=\linewidth]{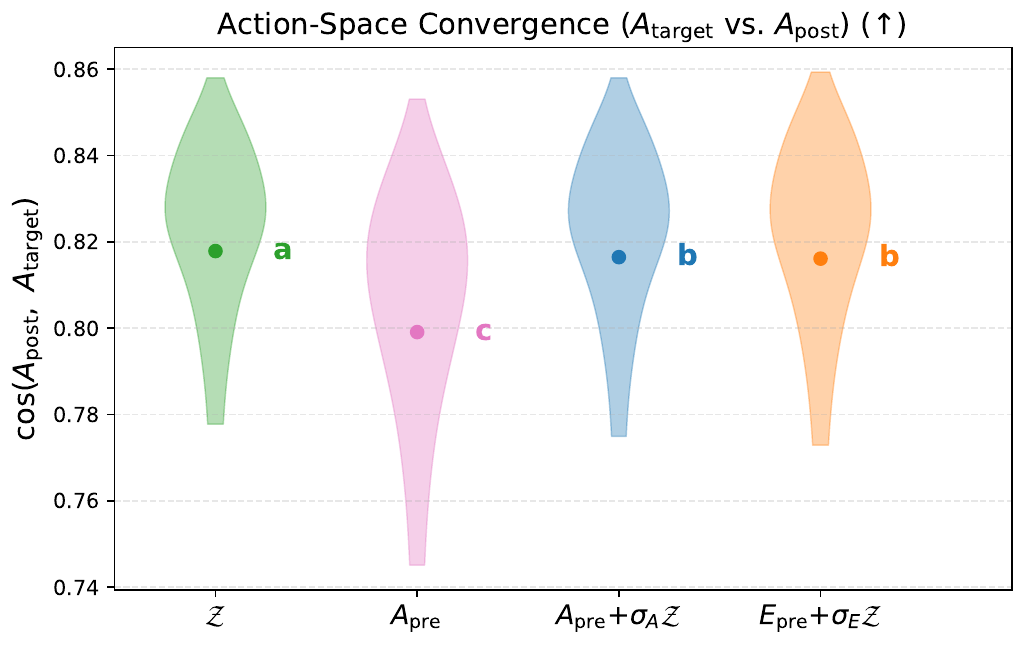}
    \caption{Action-space similarity in Setting B\textsubscript{FP} (five tasks, 100\% data). \textbf{Top:} $\|A_{\text{post}}-A_{\text{target}}\|_2$. \textbf{Bottom:} $\cos(A_{\text{post}}, A_{\text{target}})$. Violins show per-observation distributions with CLD letters from Welch's $t$-test.}
    \label{fig:bfp_action_convergence}
\end{figure}

\textbf{Action-space similarity after fine-tuning.}
\cref{fig:bfp_action_convergence} shows $\|A_{\text{post}}-A_{\text{target}}\|_2$ and the corresponding cosine similarity across priors. As in Setting~B (\cref{fig:action_convergence}), the four priors largely overlap: post-fine-tuning action predictions are comparably accurate.

\begin{figure}[!htbp]
    \centering
    \includegraphics[width=\linewidth]{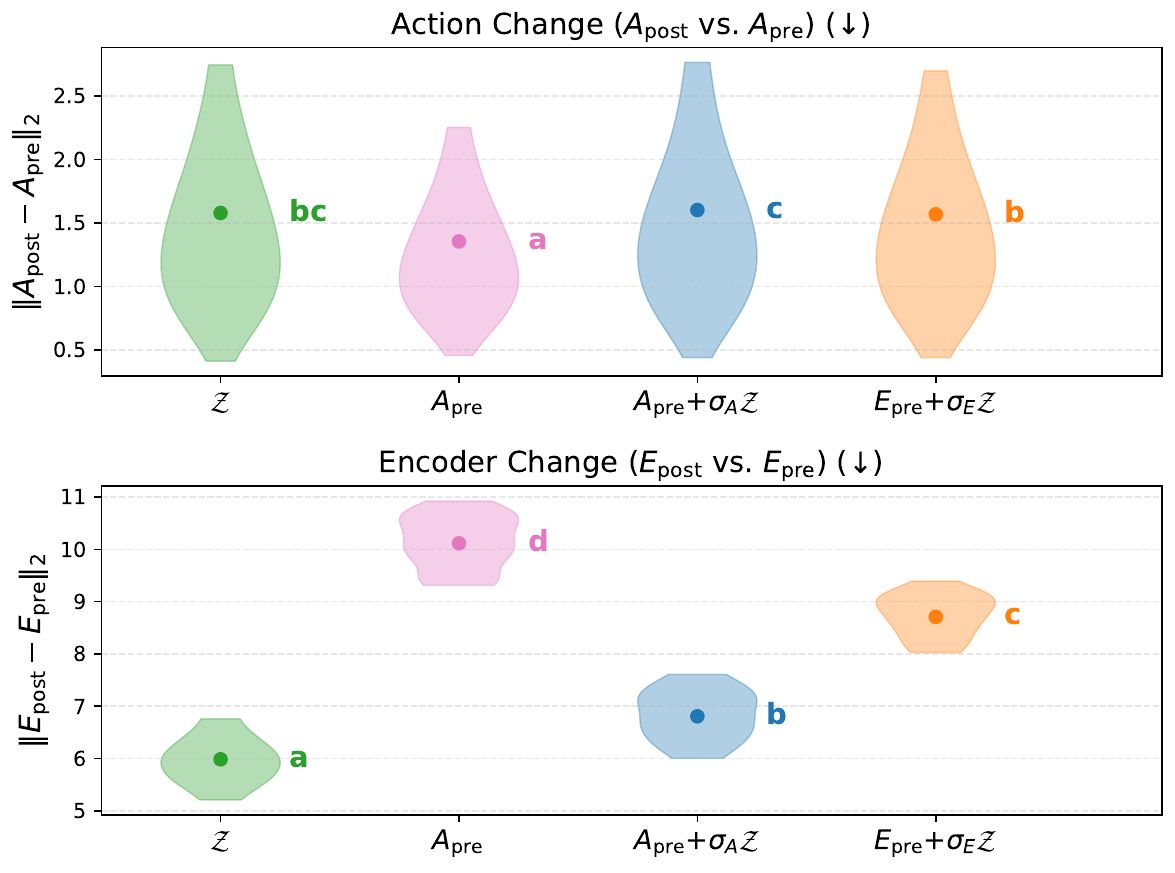}
    \caption{Representation change in Setting B\textsubscript{FP} ($\ell_2$ view, five tasks, 100\% data). \textbf{Top:} action change $\|A_{\text{post}}-A_{\text{pre}}\|_2$; all priors cluster within a narrow band. \textbf{Bottom:} encoder change $\|E_{\text{post}}-E_{\text{pre}}\|_2$; $A_{\text{pre}}$ restructures the encoder most and $\mathcal{Z}$ the least, with the mixed priors in between.}
    \label{fig:bfp_repr_change}
\end{figure}

\begin{figure}[!htbp]
    \centering
    \includegraphics[width=\linewidth]{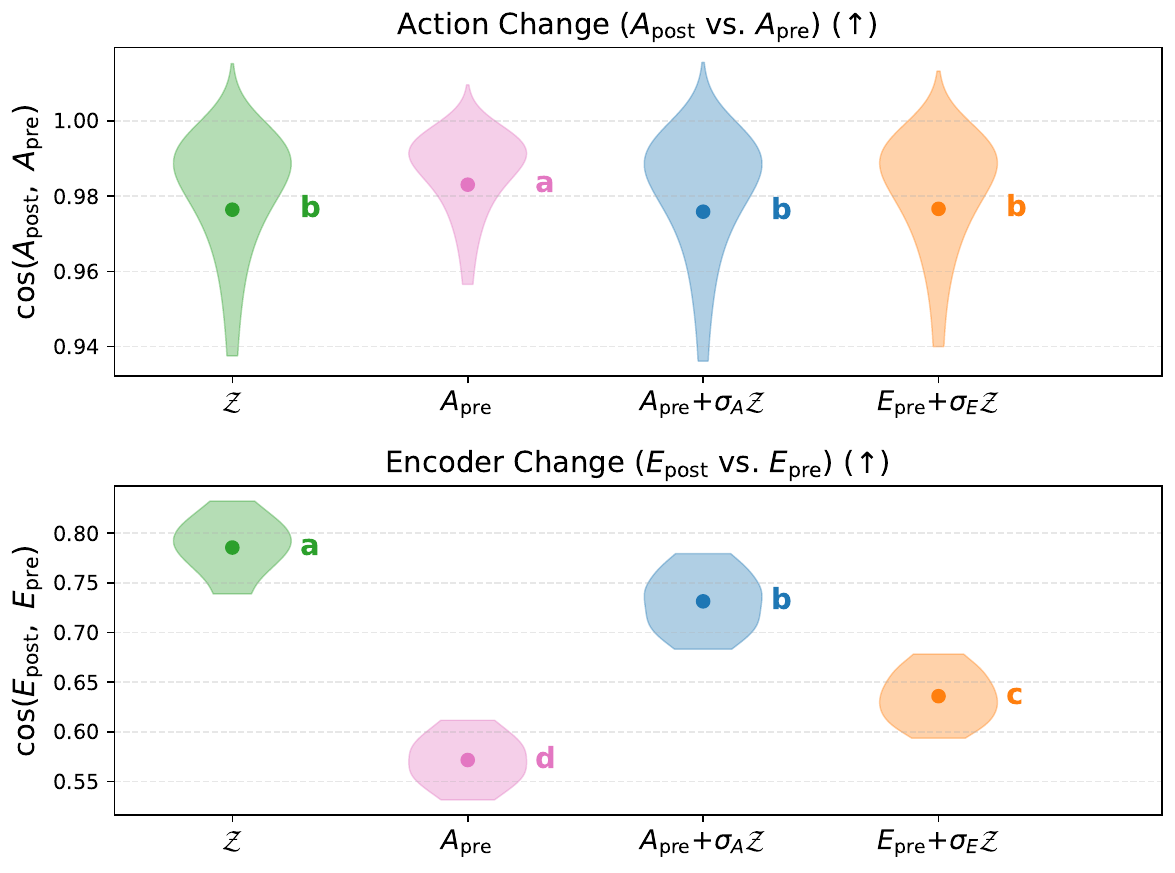}
    \caption{Representation change in Setting B\textsubscript{FP}, cosine-similarity view (companion to \cref{fig:bfp_repr_change}). Higher values indicate less directional rotation from the pretrained representation.}
    \label{fig:bfp_repr_change_cosine}
\end{figure}

\textbf{Representation divergence from the pretrained starting point.}
\cref{fig:bfp_repr_change} shows the $\ell_2$ action change $\|A_{\text{post}}-A_{\text{pre}}\|_2$ alongside the encoder change $\|E_{\text{post}}-E_{\text{pre}}\|_2$, with the cosine view in \cref{fig:bfp_repr_change_cosine}. The pattern of \cref{sec:convergence} is preserved: action-space changes largely overlap while embedding-space changes are separated more, with $\mathcal{Z}$ moving the encoder least and $A_{\text{pre}}$ the most. Since every method now trains under flow matching, the encoder-representation divergence is driven by the prior rather than by the pretraining objective.

\subsection{Setting \texorpdfstring{B\textsubscript{FP}}{B\_FP}: Noise Ablation}
\label{sec:appendix_noise_ablation}

\begin{figure}[!htbp]
    \centering
    \includegraphics[width=\linewidth]{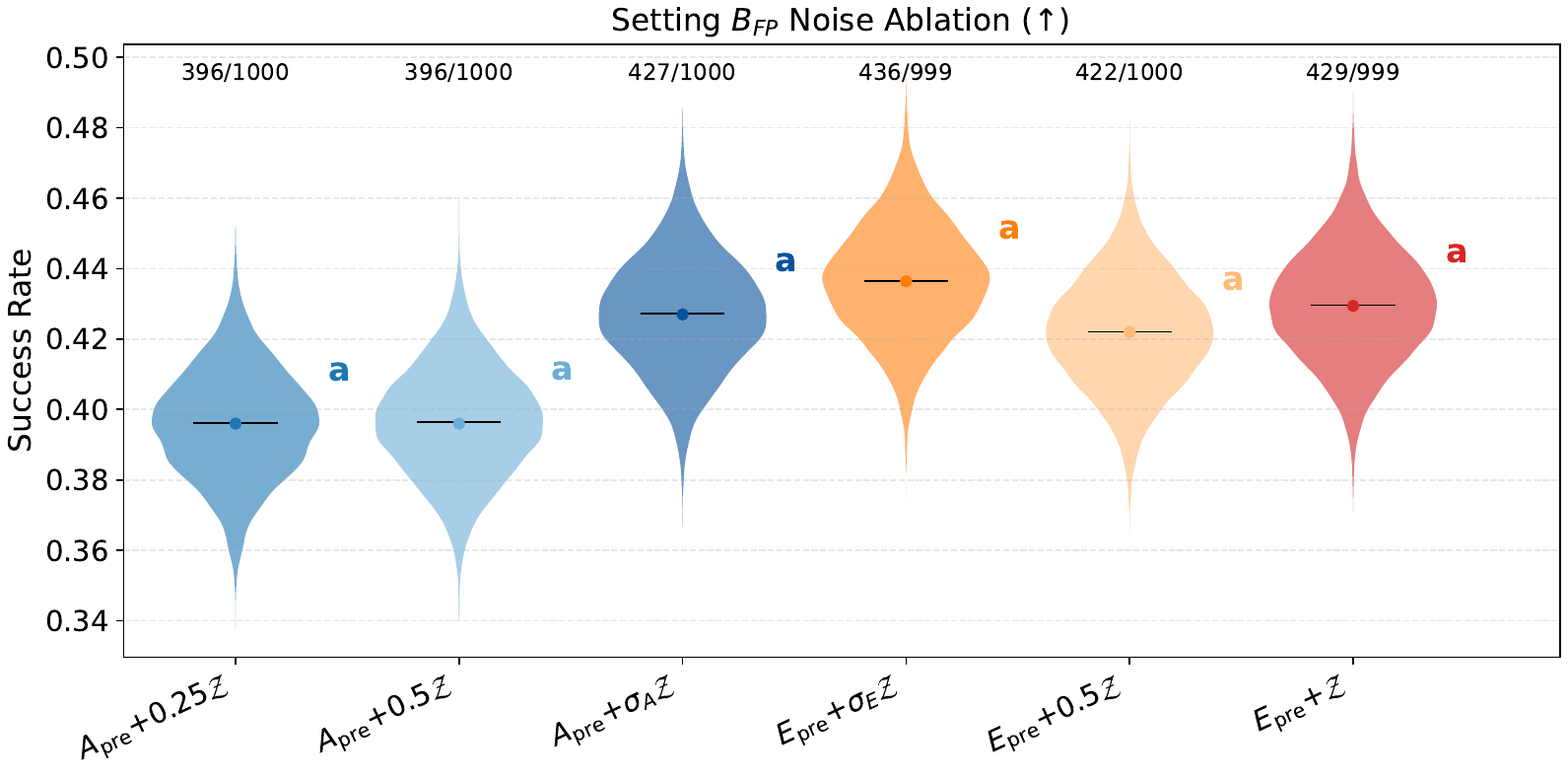}
    \caption{Noise ablation (Setting B\textsubscript{FP}, five tasks).
    Performance is robust to noise scale for $E_{\text{pre}}$ variants; $A_{\text{pre}}$ variants degrade with higher noise.}
    \label{fig:noise_ablation}
\end{figure}

The noise scales $\sigma_A$ and $\sigma_E$ in the mixed priors $A_{\text{pre}}{+}\sigma_A\mathcal{Z}$ and $E_{\text{pre}}{+}\sigma_E\mathcal{Z}$ trade off informativeness (low scale, prior dominates) against coverage (high scale, closer to Gaussian). We run this ablation under Setting~B\textsubscript{FP} so that every variant shares the flow-matching objective. \cref{fig:noise_ablation} sweeps both scales over $\{0.25, 0.5, 1.0\}$. The $E_{\text{pre}}$ variants are robust across scales; the $A_{\text{pre}}$ variants degrade at higher noise, suggesting that action-space jitter is more disruptive than embedding-space jitter. Based on this grid search we select $\sigma_A{=}1$ for $A_{\text{pre}}{+}\sigma_A\mathcal{Z}$ and $\sigma_E{=}0.25$ for $E_{\text{pre}}{+}\sigma_E\mathcal{Z}$ in the main experiments.

We also tried a learned observation-dependent noise scale $\sigma(o)$ parameterized as a small network, but this approach overfit severely: $\sigma(o)$ collapsed toward zero, driving training loss down while degrading rollout performance. We therefore use a fixed $\sigma$ throughout.

\subsection{Setting \texorpdfstring{B\textsubscript{real}}{B\_real}: Hardware Prior-Selection Pilot}
\label{sec:appendix_hw_pilot}

\begin{figure}[!htbp]
    \centering
    \includegraphics[width=\linewidth]{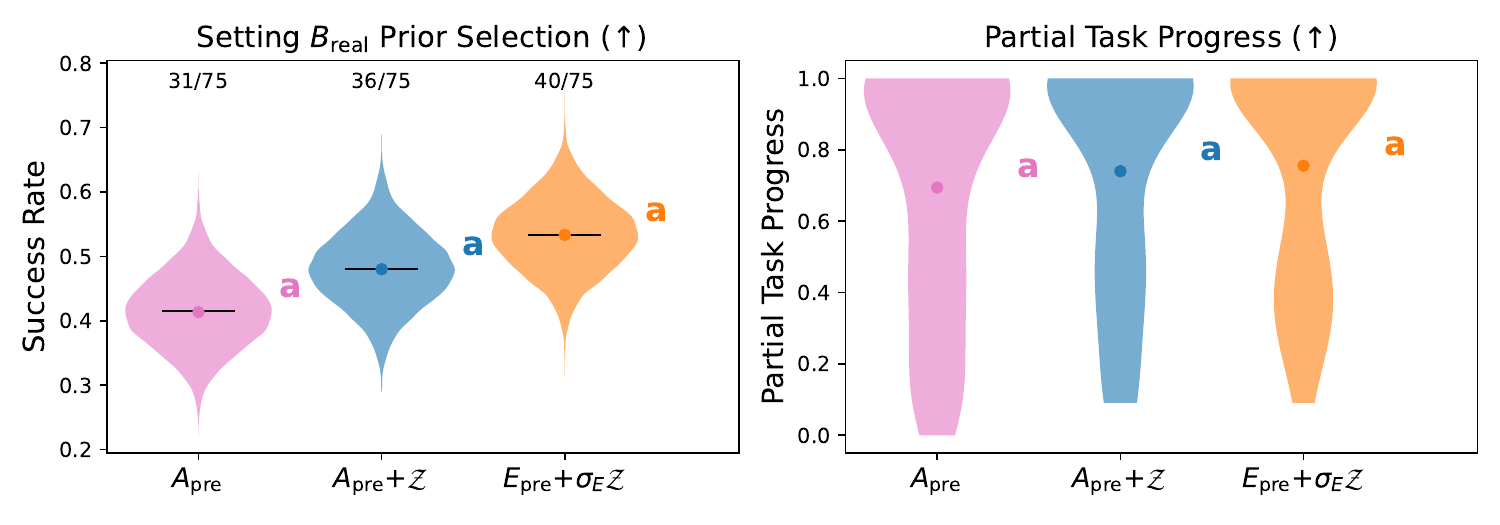}
    \caption{Setting B\textsubscript{real} prior-selection pilot (three tasks, 25 rollouts per method). \textbf{Left:} success rate. \textbf{Right:} partial task progress. All three priors are statistically tied on both metrics.}
    \label{fig:hw_pilot}
\end{figure}

Before the full hardware evaluation (\cref{sec:setting_b_real}), we ran a pilot to choose a single representative learned prior among our three proposed action- and embedding-space variants. We compared $A_{\text{pre}}$, $A_{\text{pre}}{+}\sigma_A\mathcal{Z}$, and $E_{\text{pre}}{+}\sigma_E\mathcal{Z}$ on 3 of the 5 hardware tasks with 25 rollouts per method per task (75 per method, 225 total). \cref{fig:hw_pilot} shows all three fall in CLD group ``a'' on partial-progress (69\%, 74\%, and 76\% respectively) and on binary success (31/75, 36/75, 40/75). We carry $E_{\text{pre}}{+}\sigma_E\mathcal{Z}$ forward to the full 5-method evaluation on the strength of its numerical lead, alongside $\mathcal{Z}$ and the three external baselines (Cocos, BRIDGER, Retrieval).

\subsection{Setting C: Zero-Shot Prior Quality}
\label{sec:appendix_c_prior_distance}

\begin{figure*}[!htbp]
    \centering
    \includegraphics[width=0.72\textwidth]{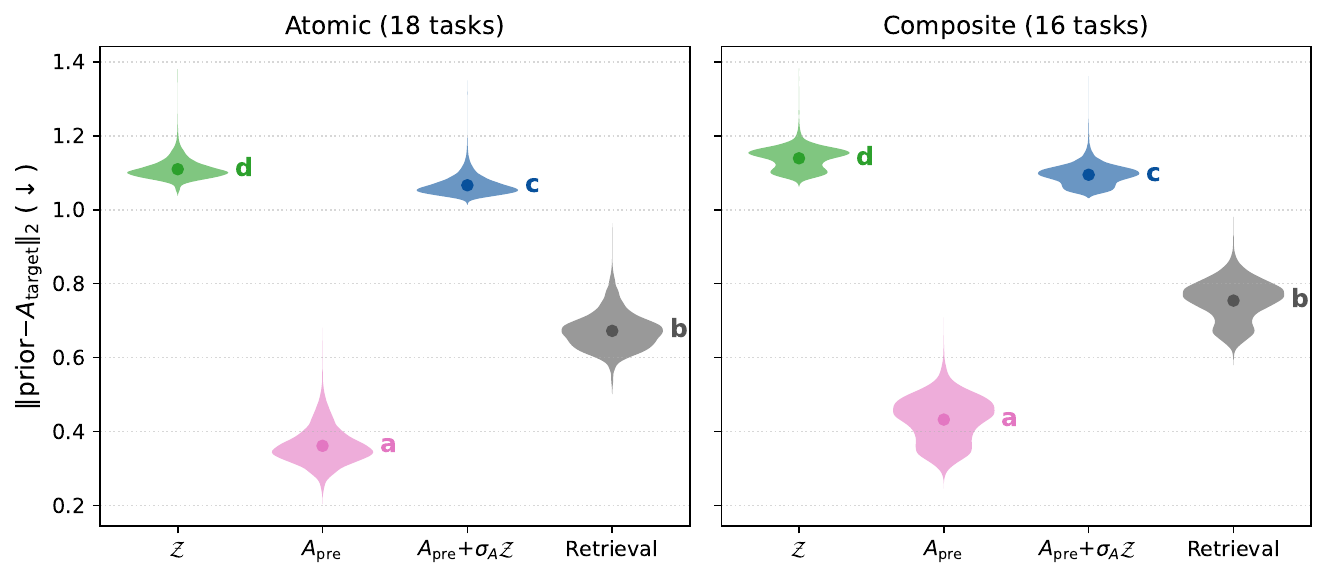}
    \caption{Zero-shot prior-to-target distance for $\pi_{0.5}$ on RoboCasa. Violins show the distribution of per-step prior-to-target $\ell_2$ distances pooled across tasks ($n{=}18$ atomic, $n{=}16$ composite); dots mark the mean and CLD letters are from Welch's $t$-test. $A_{\text{pre}}$ remains substantially closer to the target than the Gaussian on both splits.}
    \label{fig:setting_c_prior_distance}
\end{figure*}

In Setting~C (\cref{sec:setting_c}), $A_{\text{pre}}$ and Retrieval both underperform the Gaussian, which could suggest they are poor priors. We rule this out in \cref{fig:setting_c_prior_distance}: across both task splits, zero-shot $\pi_{0.5}$ actions remain measurably closer to the target than the Gaussian, by a factor of ${\sim}3\times$.

\subsection{Per-Task CLD Breakdowns}
\label{sec:appendix_per_task}

\begin{figure*}[!htbp]
    \centering
    \includegraphics[width=\textwidth]{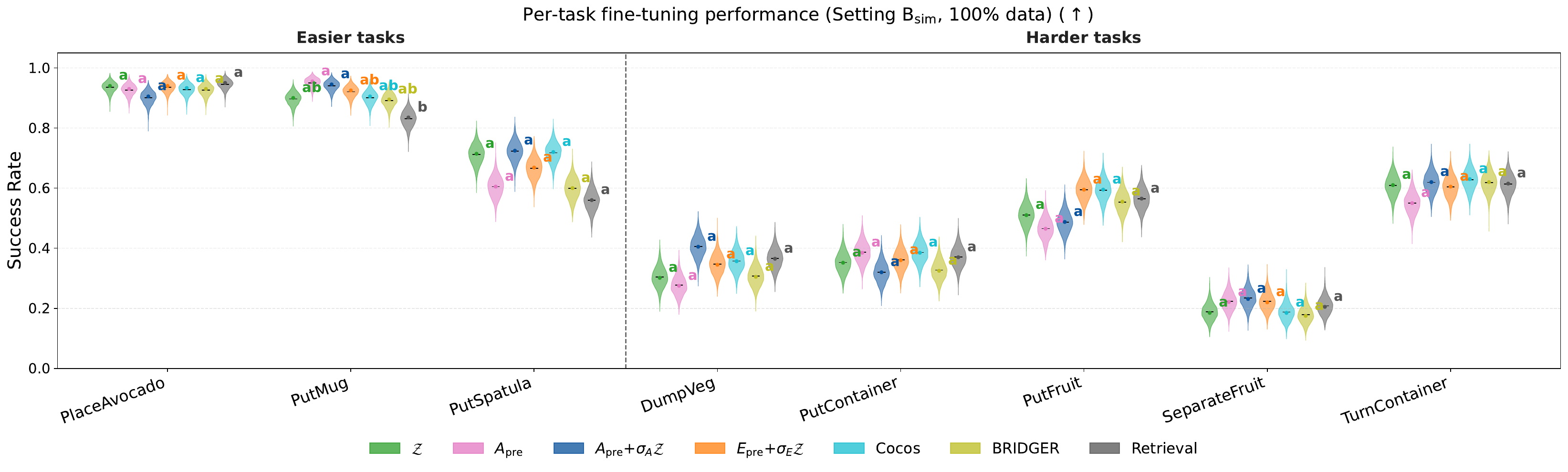}
    \caption{Setting B\textsubscript{sim} per-task success rates at 100\% data (LBM 1.0, eight tasks, 200 rollouts per method--task). CLD letters are assigned independently within each task. The vertical dotted line separates the 3 easier tasks (left) from the 5 harder Kitchen-K tasks (right).}
    \label{fig:per_task_b_sim}
\end{figure*}

\begin{figure*}[!htbp]
    \centering
    \includegraphics[width=\textwidth]{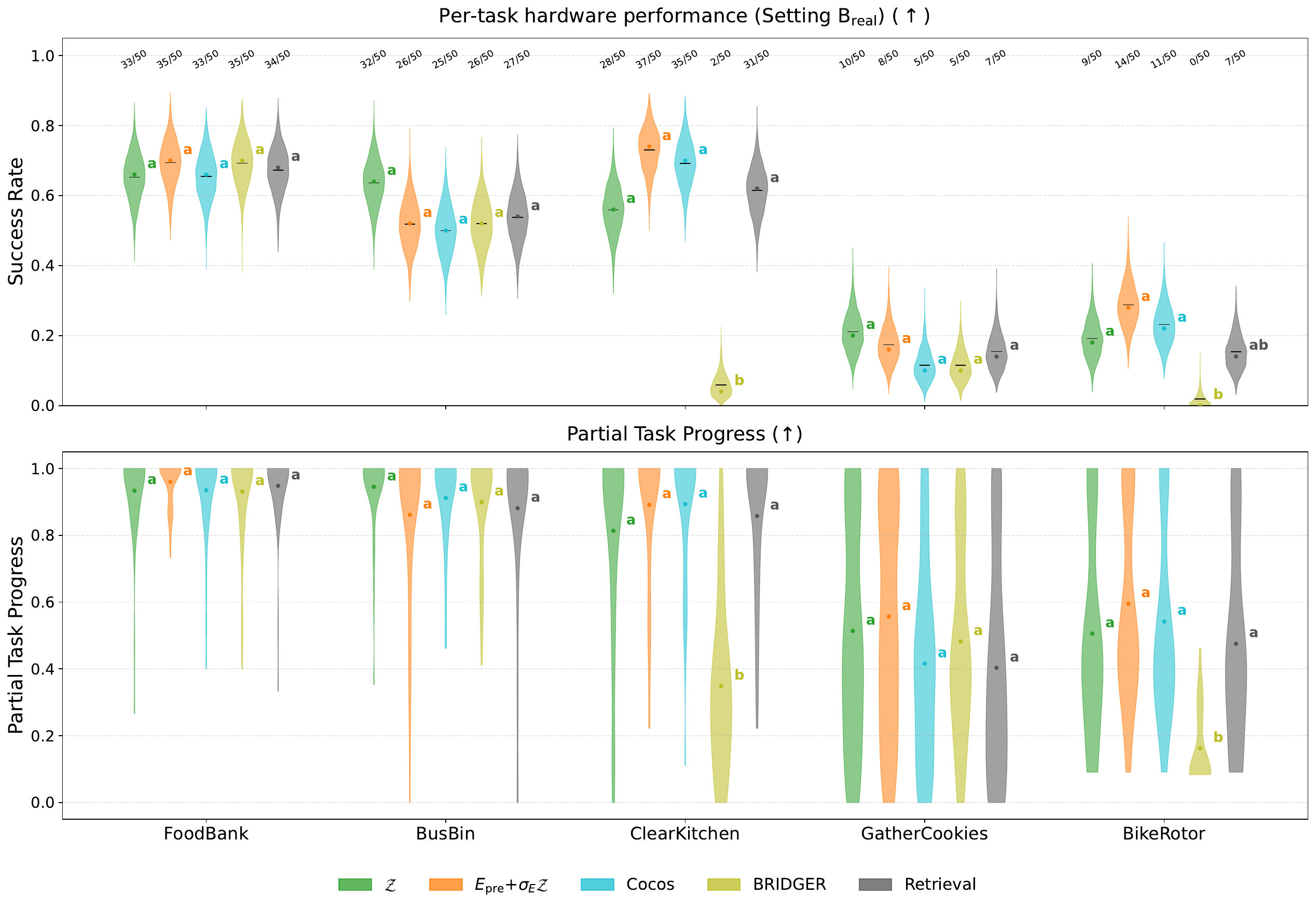}
    \caption{Setting B\textsubscript{real} per-task performance (LBM 1.0 on hardware, five tasks, 50 A/B-test rollouts per method--task). \textbf{Top:} success rate violins with CLD letters computed within each task. \textbf{Bottom:} partial-task-progress violins with Welch CLD.}
    \label{fig:per_task_b_real}
\end{figure*}

The main text reports success rates averaged across tasks within each setting. Here we break those averages down per task for the two LBM 1.0 settings, with CLD letters computed independently within each task. \cref{fig:per_task_b_sim} shows Setting~B\textsubscript{sim} at 100\% data across all eight tasks, ordered easier (three seen-scenario tasks) to harder (five Kitchen-K tasks); \cref{fig:per_task_b_real} shows Setting~B\textsubscript{real} across the five hardware tasks, with partial task progress printed under each task name. The Gaussian $\mathcal{Z}$ sits in CLD group ``a'' or ``ab'' on every task in both settings, so it remains statistically indistinguishable from every other prior at the task level.

\subsection{Zero-Shot Prior Rollouts}
\label{sec:appendix_prior_rollouts}

\begin{figure*}[t]
    \centering
    \setlength{\tabcolsep}{1.5pt}
    \renewcommand{\arraystretch}{0.0}
    \newcommand{\priorcol}[1]{\includegraphics[width=0.13\textwidth]{figures/prior_rollouts/#1}}
    \begin{tabular}{c@{\hspace{3pt}}ccccccc}
    & \small $\mathcal{Z}$ & \small $A_{\text{pre}}$ & \small $A_{\text{pre}}{+}\sigma_A\mathcal{Z}$ & \small $E_{\text{pre}}{+}\sigma_E\mathcal{Z}$ & \small Cocos & \small BRIDGER & \small Retrieval \\
    \small $\sim$1\,s  & \priorcol{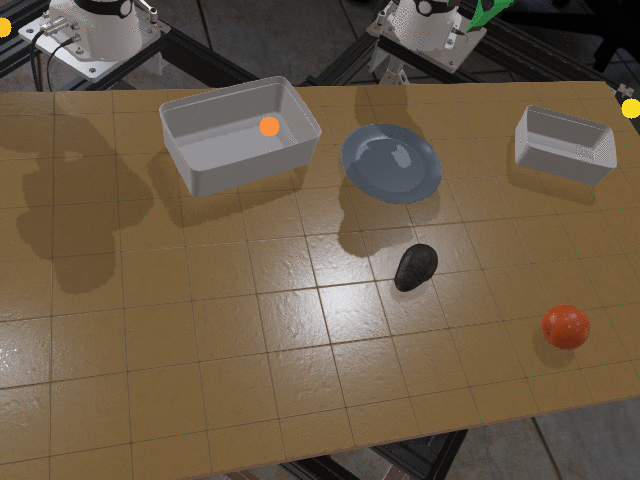} & \priorcol{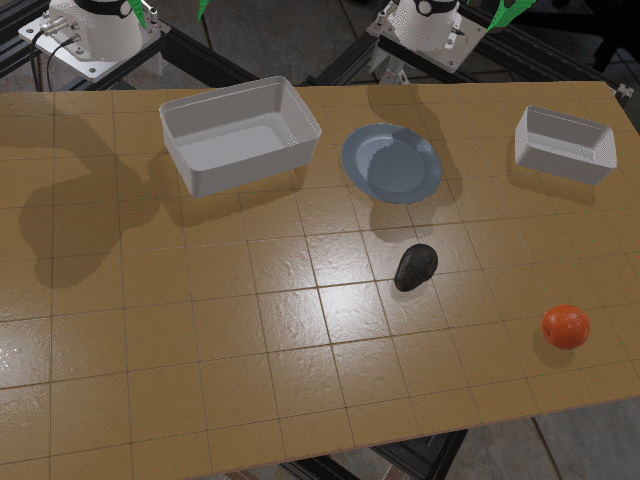} & \priorcol{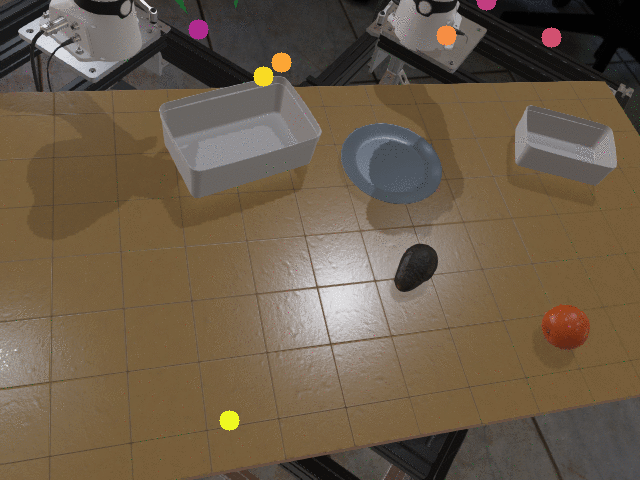} & \priorcol{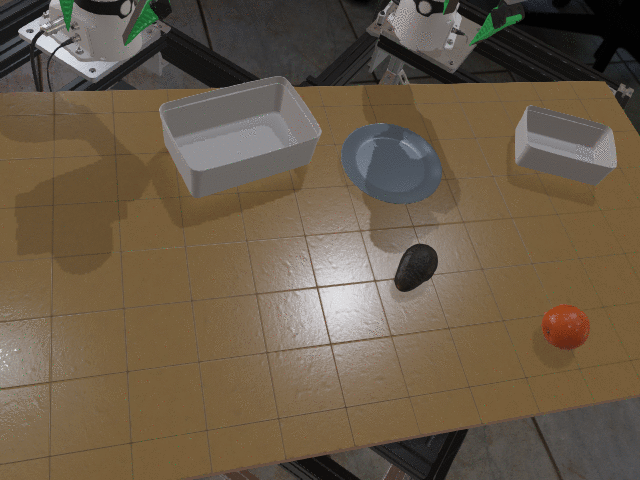} & \priorcol{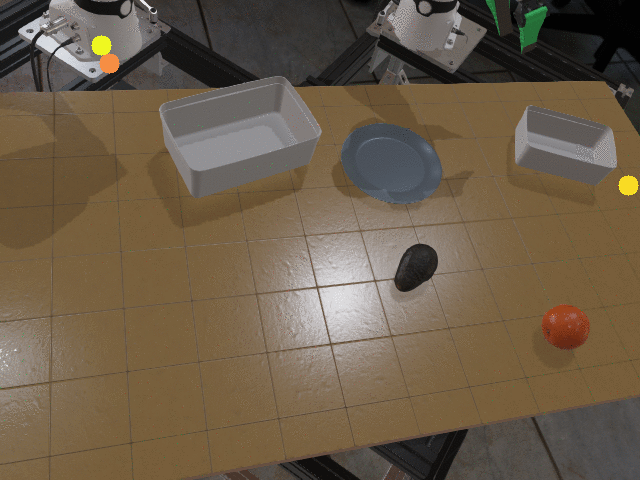} & \priorcol{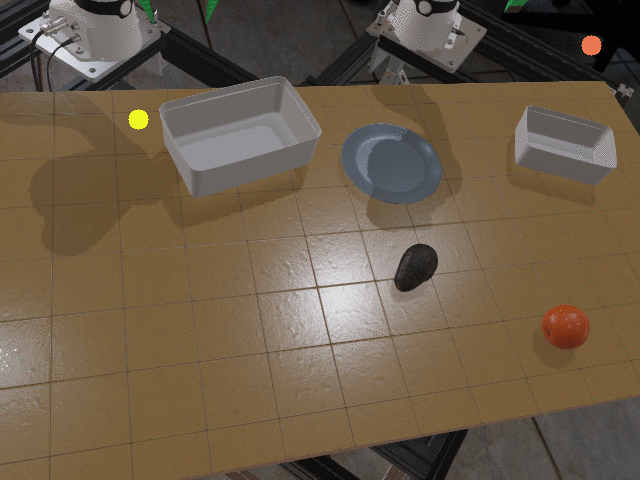} & \priorcol{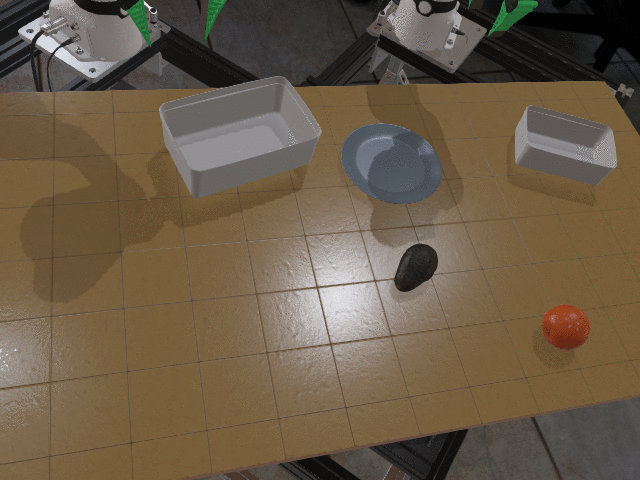} \\
    \small $\sim$5\,s  & \priorcol{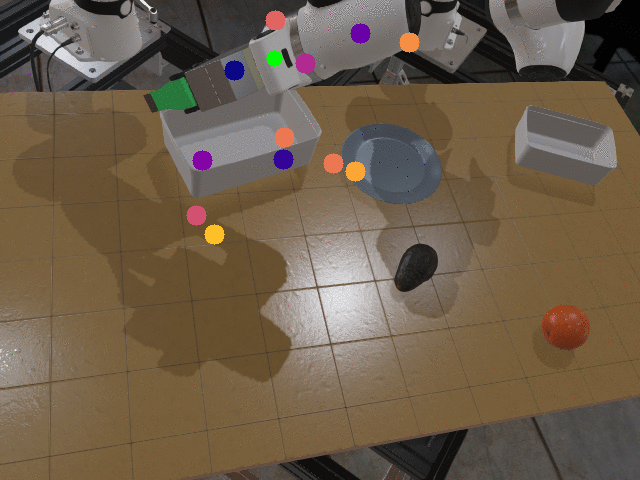} & \priorcol{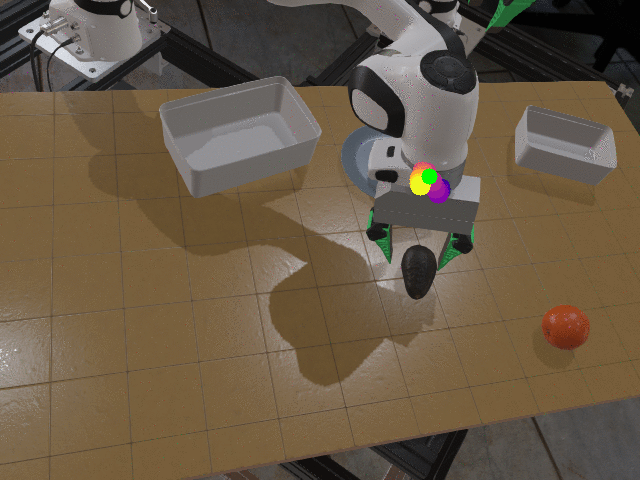} & \priorcol{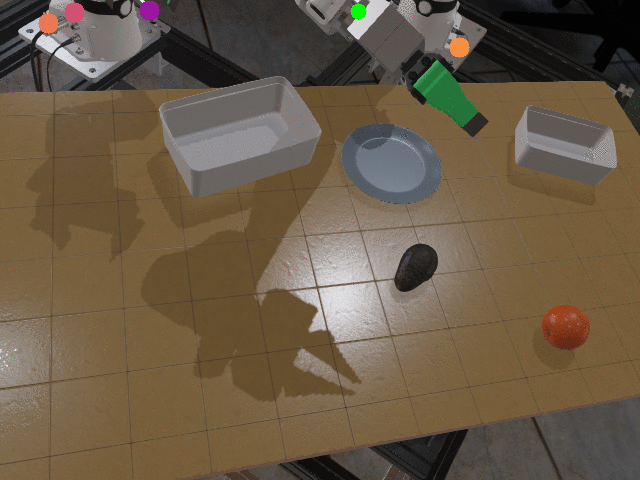} & \priorcol{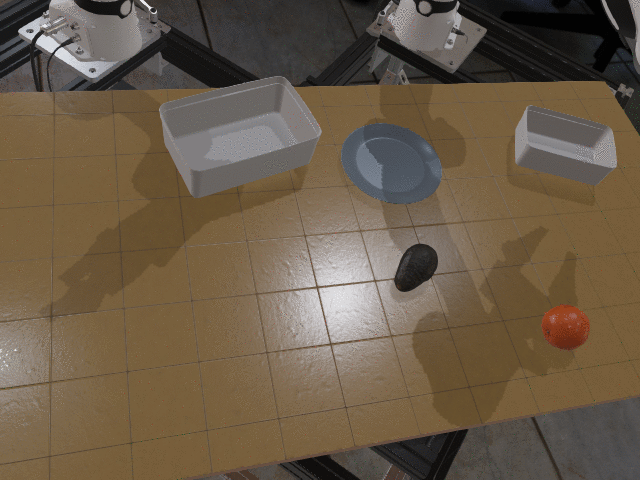} & \priorcol{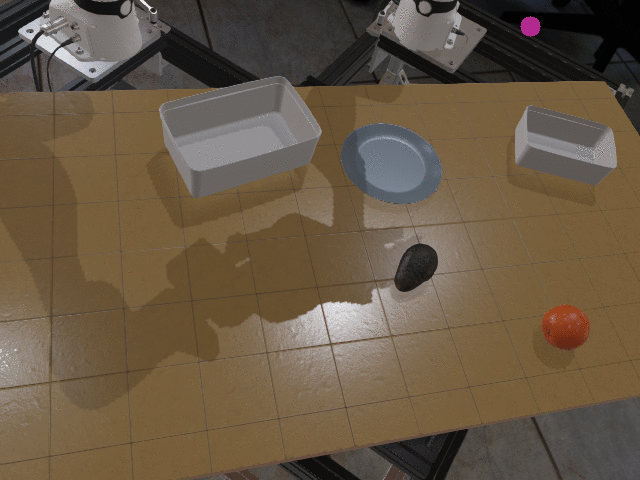} & \priorcol{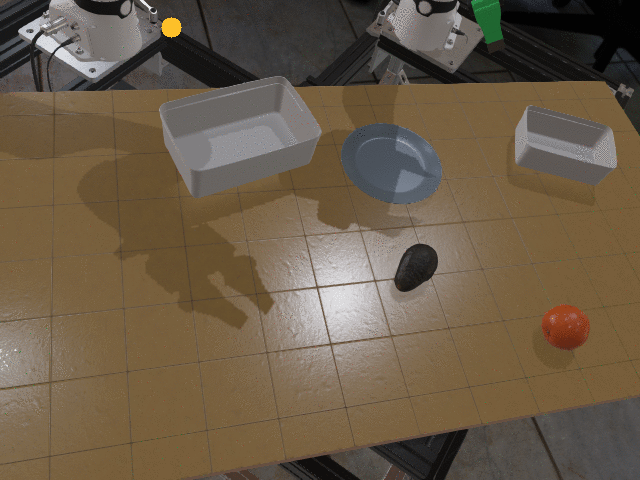} & \priorcol{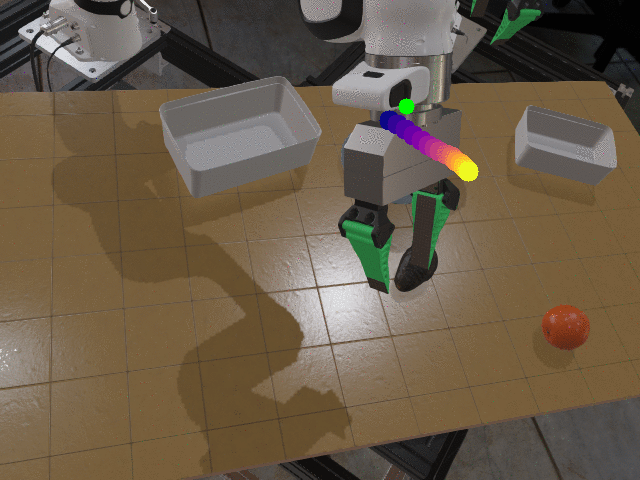} \\
    \small $\sim$9\,s  & \priorcol{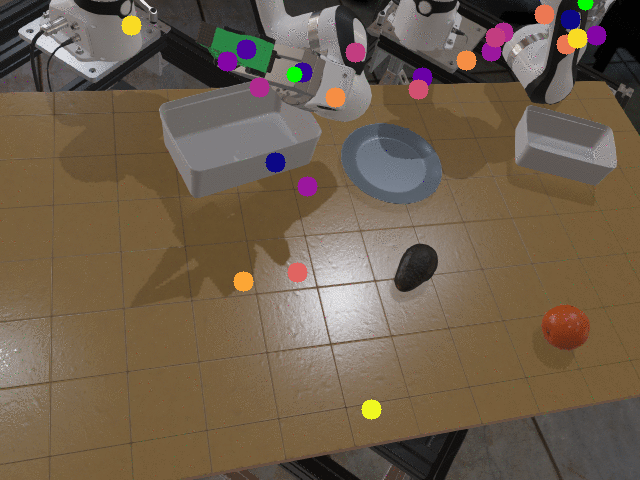} & \priorcol{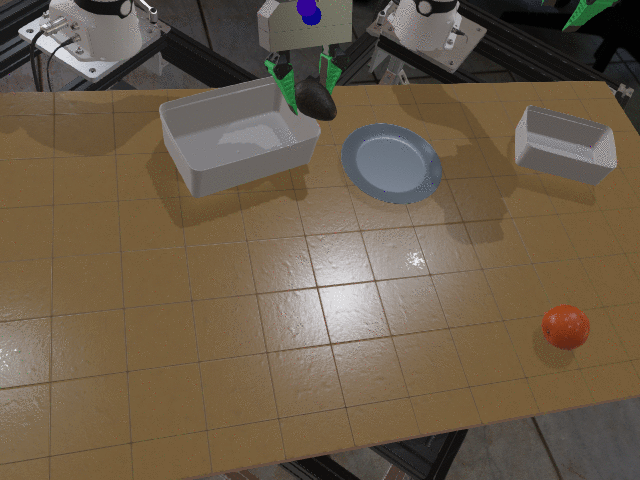} & \priorcol{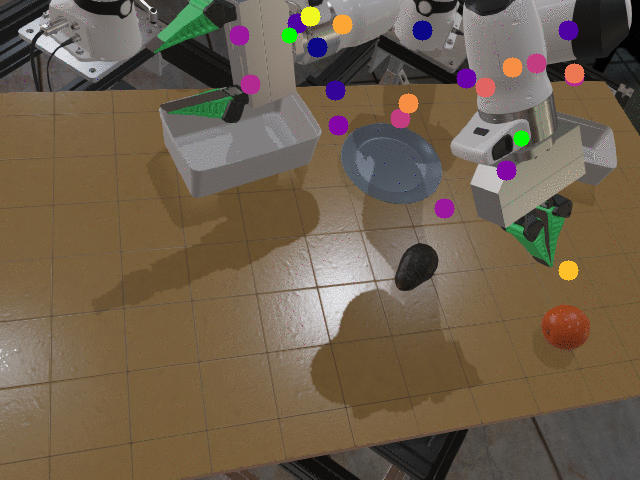} & \priorcol{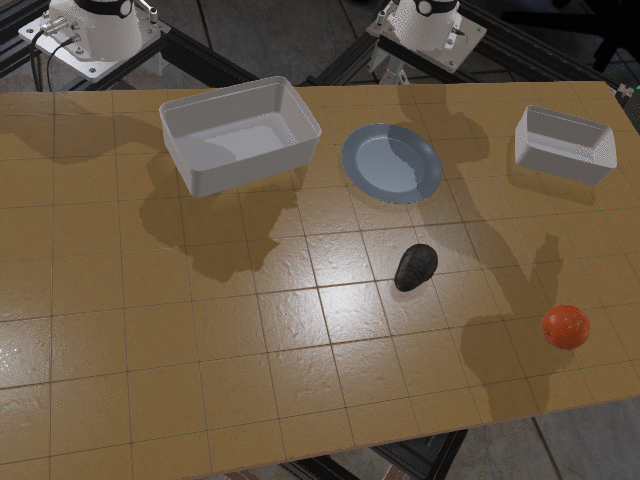} & \priorcol{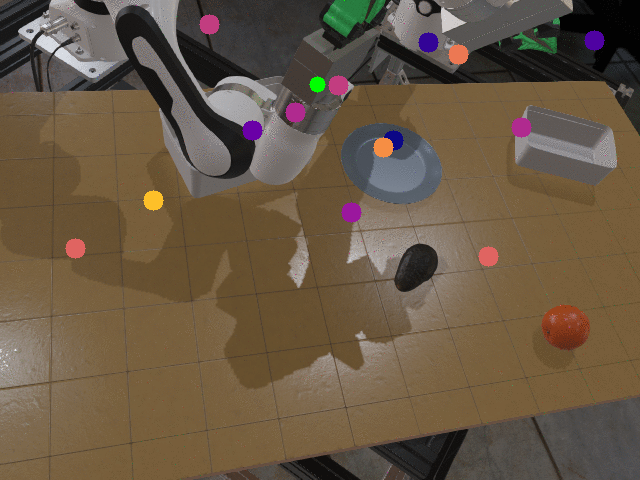} & \priorcol{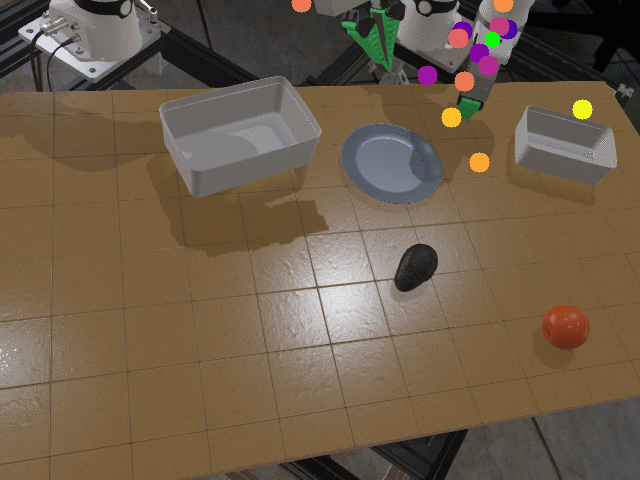} & \priorcol{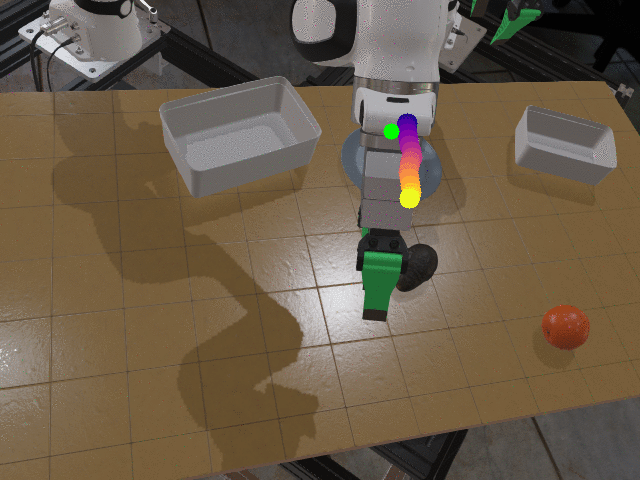} \\
    \small $\sim$13\,s & \priorcol{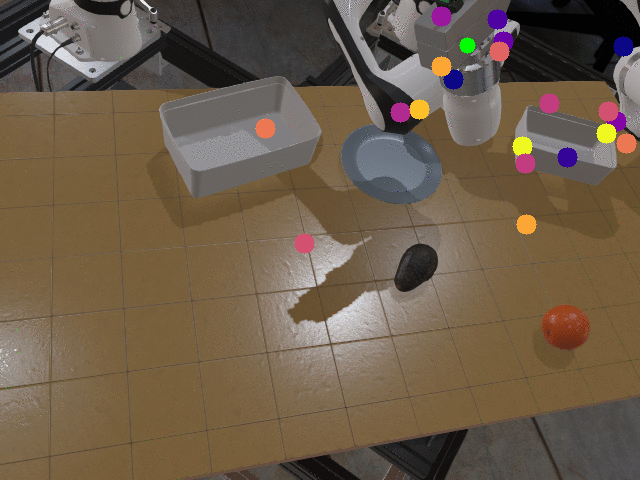} & \priorcol{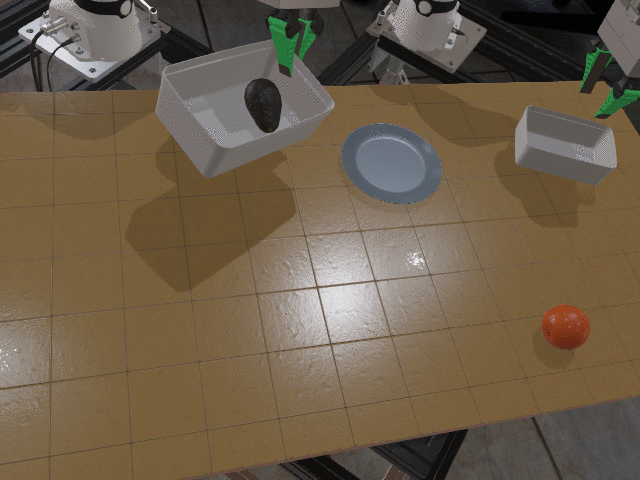} & \priorcol{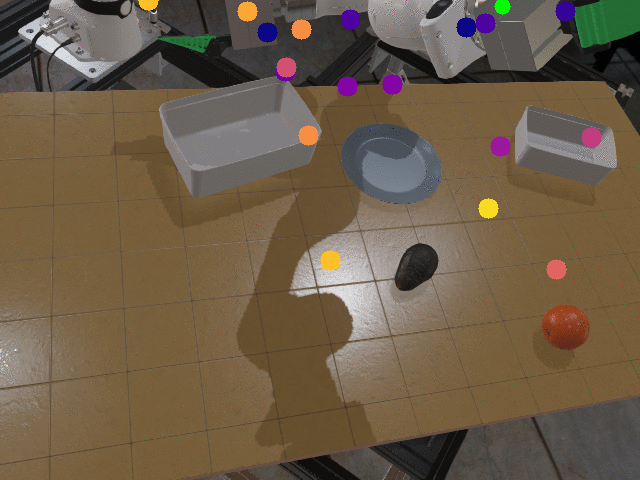} & \priorcol{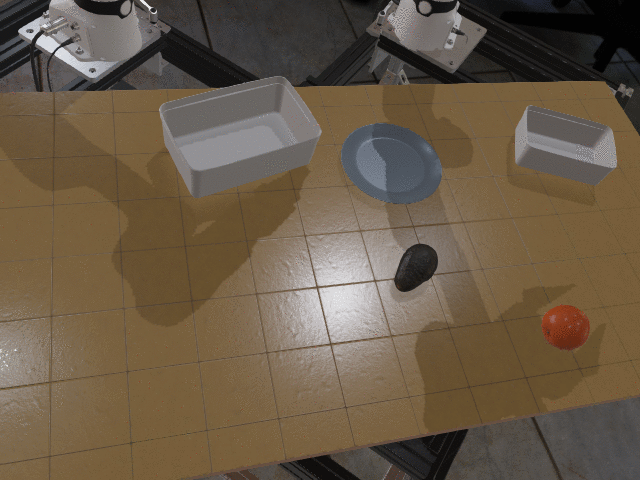} & \priorcol{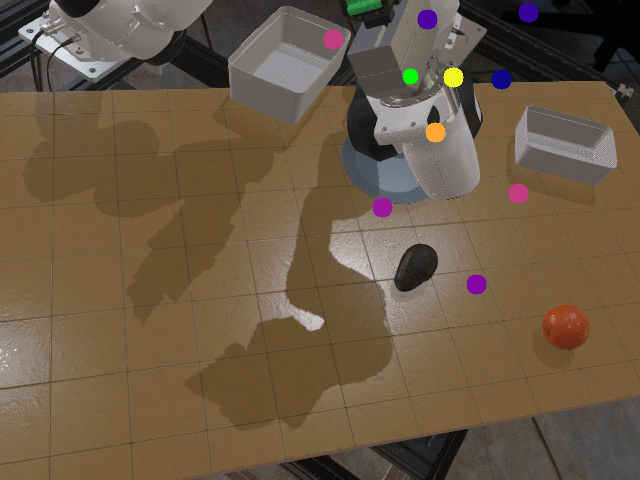} & \priorcol{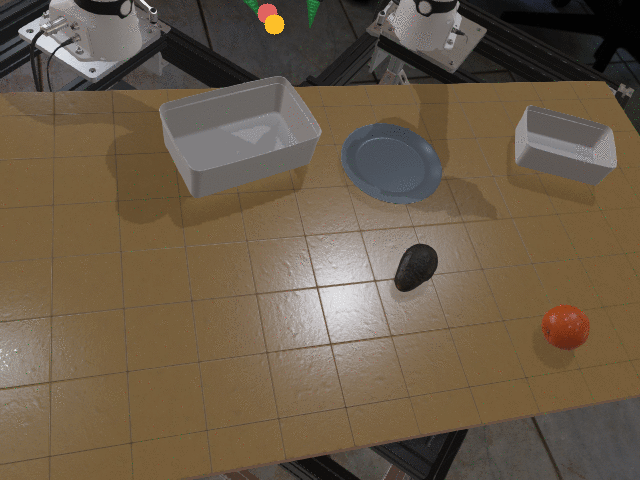} & \priorcol{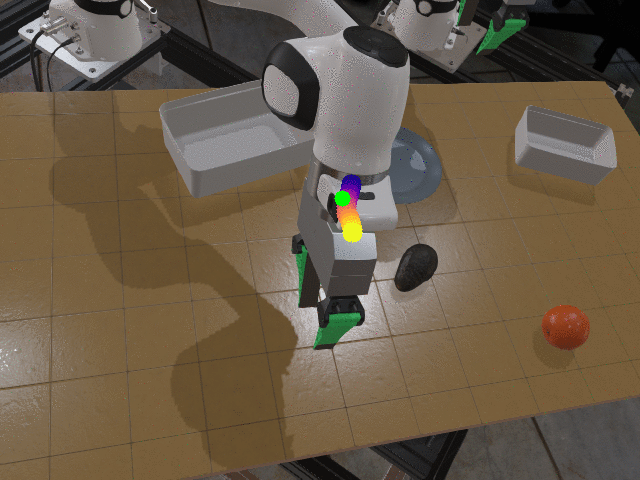} \\
    \small $\sim$15\,s & \priorcol{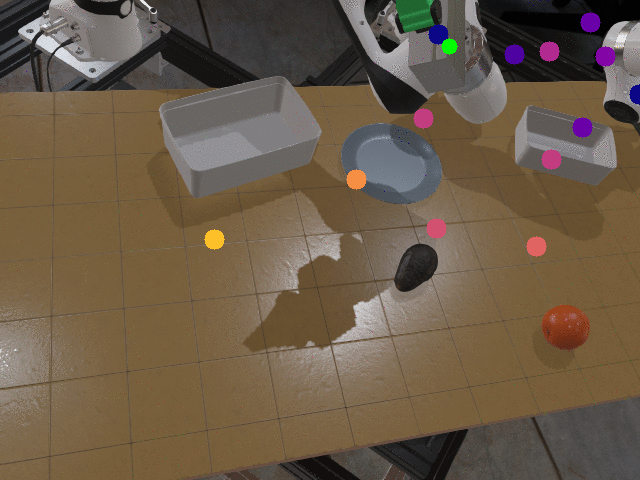} & \priorcol{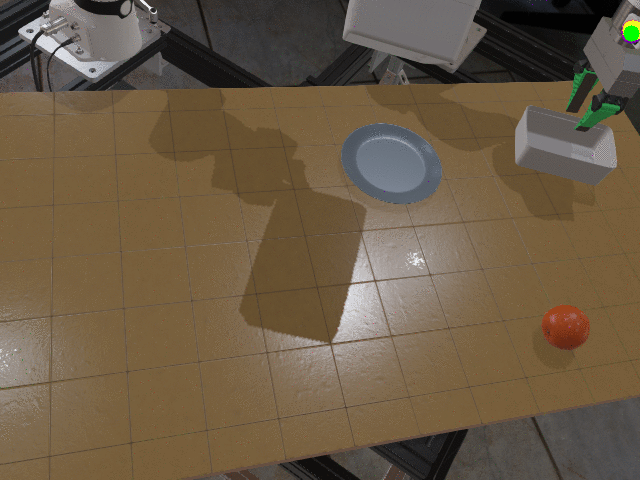} & \priorcol{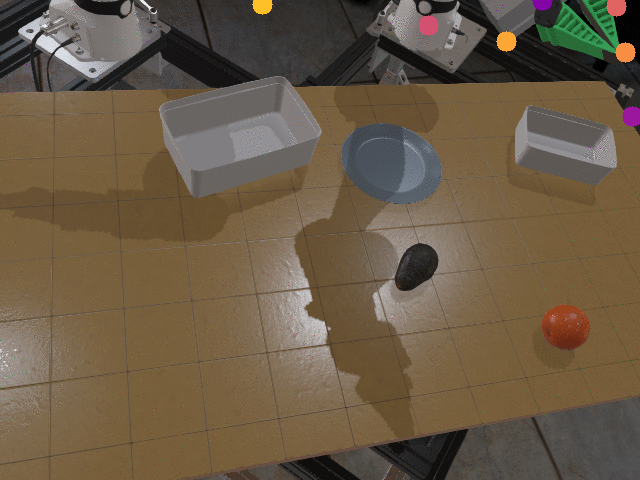} & \priorcol{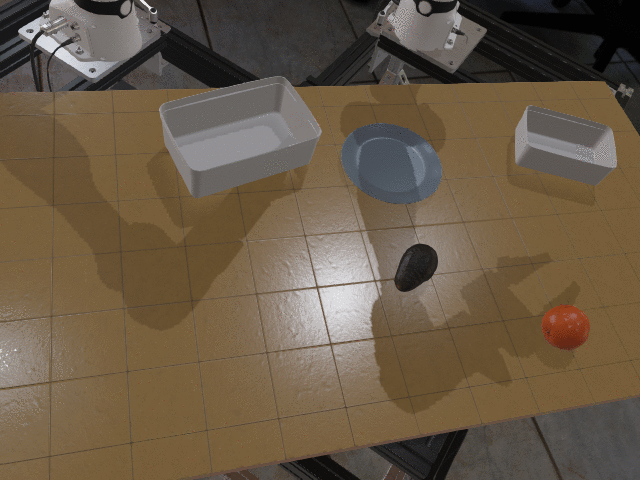} & \priorcol{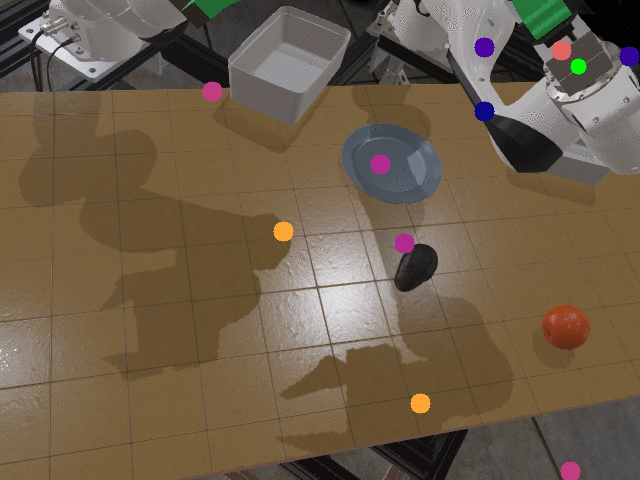} & \priorcol{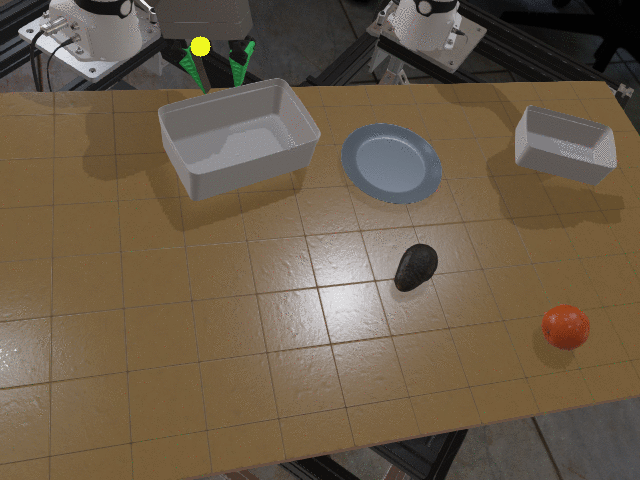} & \priorcol{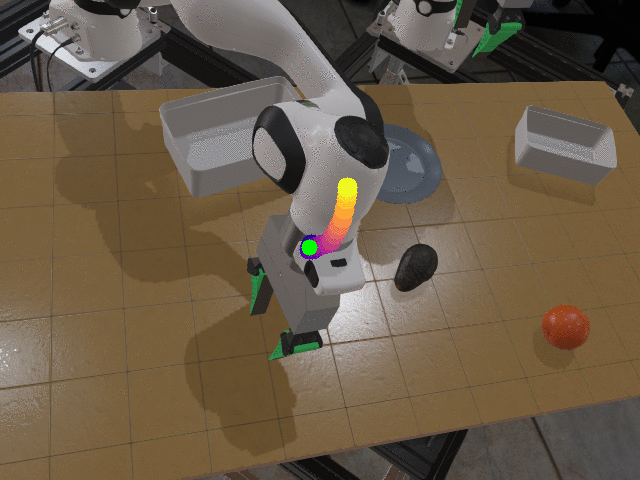} \\
    \small $\sim$19\,s & \priorcol{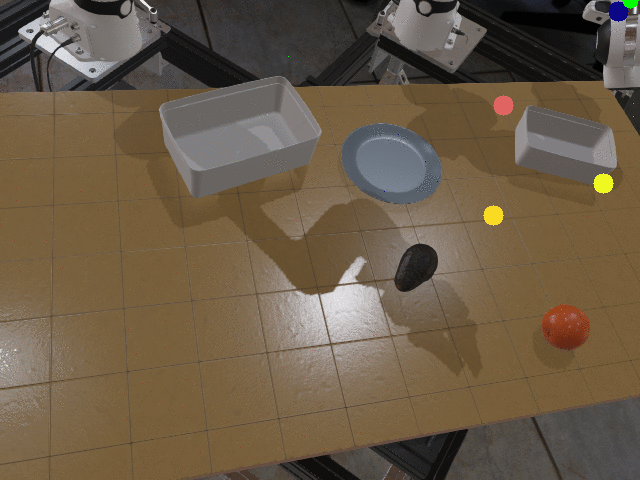} & \priorcol{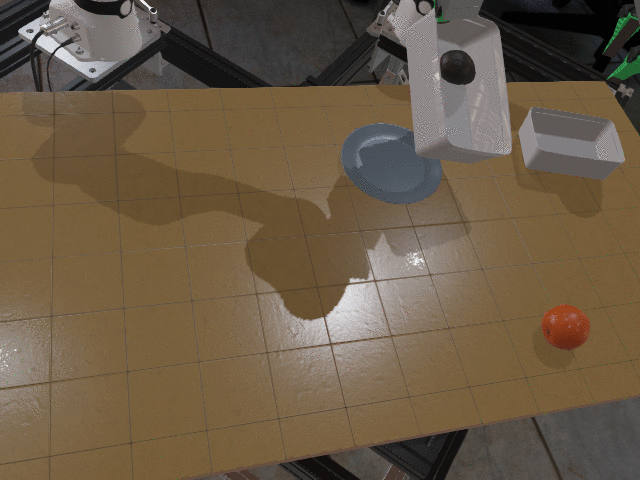} & \priorcol{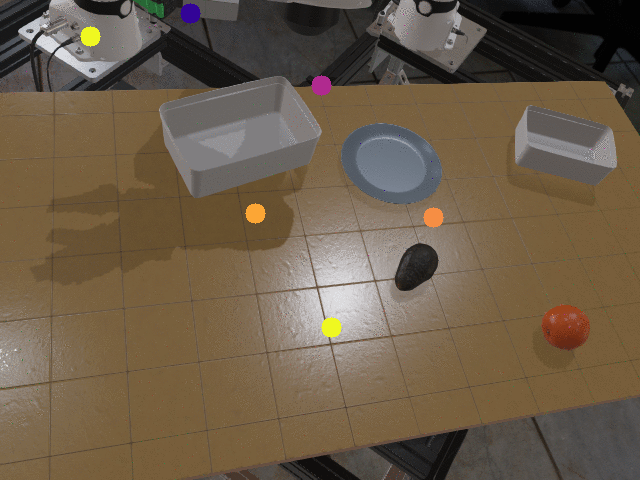} & \priorcol{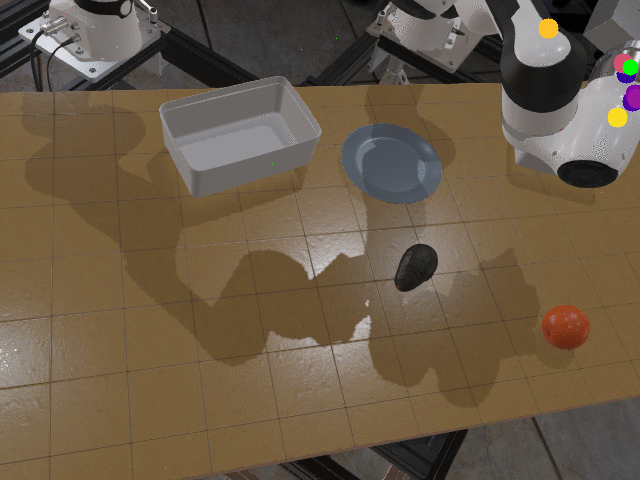} & \priorcol{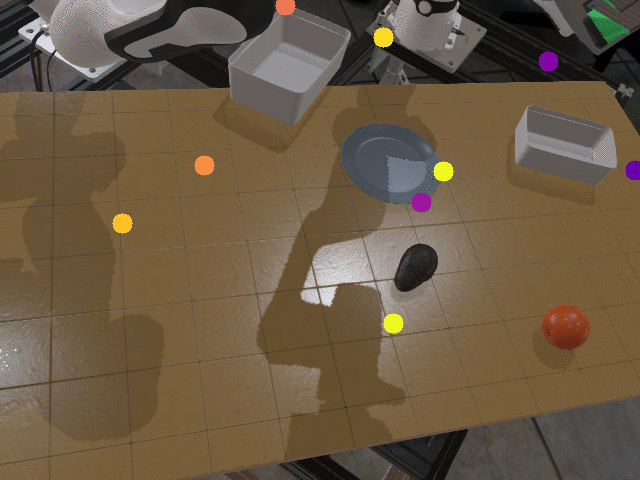} & \priorcol{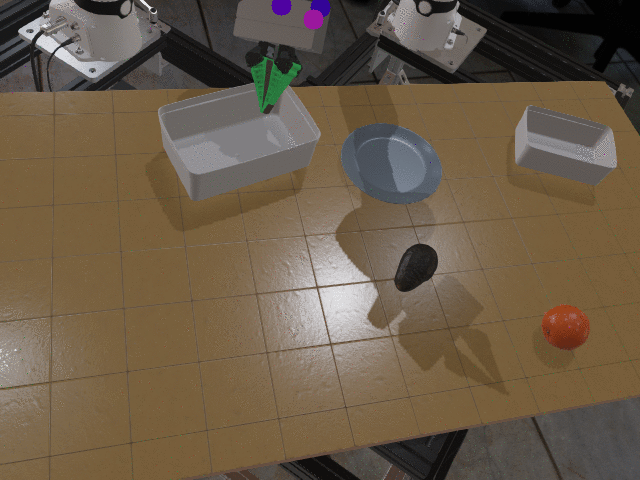} & \priorcol{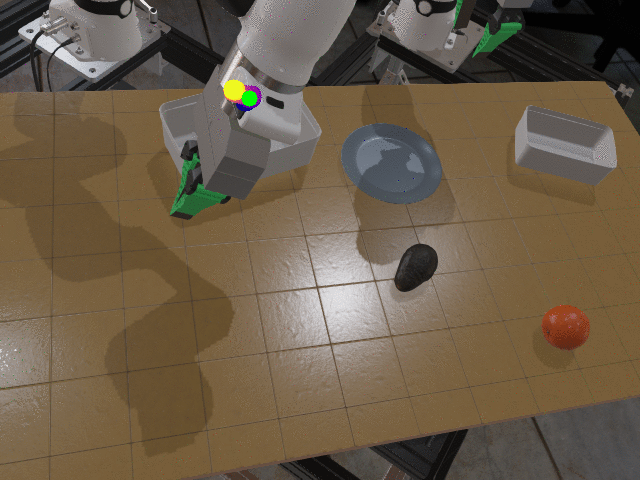} \\
    \end{tabular}
    \caption{Zero-shot rollouts of all seven priors on DumpVeg (columns: priors as in \cref{tab:priors}; rows: progression over a 20\,s rollout). Colored dots show both arms' predicted end-effector positions over the action chunk horizon (color gradient indicates temporal ordering). Frames without visible dots occur when predicted positions project outside the camera's field of view (e.g., $A_{\text{pre}}$ at ${\sim}9$--$19$\,s, where arms moved off-screen after task progress) or cluster into a single indistinguishable point due to near-stationary predictions (e.g., Cocos at ${\sim}5$\,s). The Gaussian baseline ($\mathcal{Z}$) is erratic throughout, as expected from pure noise. Several learned priors (e.g., $A_{\text{pre}}$, Cocos, Retrieval) occasionally produce rollouts that make legitimate task progress; the rest sit between these extremes.}
    \label{fig:prior_rollouts_grid}
\end{figure*}

To complement \cref{fig:prior_illustration}, \cref{fig:prior_rollouts_grid} shows rollouts of all seven prior variants side by side on the DumpVeg task: columns are priors in the order of \cref{tab:priors}, rows are fixed time steps of the 20\,s rollout (${\sim}1, 5, 9, 13, 15, 19$\,s). To probe how closely each prior already aligns with the target task before any prior-based fine-tuning, we replace the policy's predicted action at every step with a sample drawn directly from the prior.

\end{document}